\documentclass[preprint]{siamart190516}
\usepackage{hyperref}
\usepackage{url}
\usepackage{smile}
\usepackage{graphicx} 
\usepackage{algorithm}
\usepackage{algorithmic}
\usepackage{todonotes}
\usepackage{epstopdf}
\usepackage[normalem]{ulem}
\usepackage[export]{adjustbox}
\usepackage{mathtools, cuted}
\usepackage{bbm}
\usepackage{wrapfig}
\usepackage{subcaption}
\usepackage{caption}
\usepackage{enumitem}
\usepackage{smile}
\usepackage{tcolorbox}
\usepackage{wrapfig,booktabs}

\allowdisplaybreaks

\usepackage[nocompress]{cite}

\usepackage{lipsum}
\usepackage{amsfonts}
\usepackage{graphicx}
\usepackage{epstopdf}
\usepackage{algorithmic}
\ifpdf
  \DeclareGraphicsExtensions{.eps,.pdf,.png,.jpg}
\else
  \DeclareGraphicsExtensions{.eps}
\fi

\newcommand{\TheTitle}{Block Policy Mirror Descent
}

\headers{block policy mirror descent}{G. Lan, ~Y. Li, ~and T. Zhao}

\title{{\TheTitle}
\thanks{This research was partially supported by NSF grant CCF 1909298.}
}

\author{
    Guanghui Lan \thanks{H. Milton Stewart School of Industrial and Systems Engineering, Georgia Institute of Technology, Atlanta, GA, 30332. (E-mail: \email{george.lan@isye.gatech.edu}).}
      \and 
    Yan Li   \thanks{H. Milton Stewart School of Industrial and Systems Engineering, Georgia Institute of Technology, Atlanta, GA, 30332. (E-mail: \email{yli939@gatech.edu}).}
    \and
    Tuo Zhao \thanks{H. Milton Stewart School of Industrial and Systems Engineering, Georgia Institute of Technology, Atlanta, GA, 30332. (E-mail: \email{tourzhao@gatech.edu}).
}
}

\usepackage{amsmath}
\usepackage{amsfonts}
\usepackage{latexsym}
\usepackage{amssymb}
\usepackage{framed}

\usepackage{bbm}

\usepackage{multirow}
\usepackage{hhline}

\DeclarePairedDelimiter{\ceil}{\lceil}{\rceil}

\DeclarePairedDelimiter\abs{\lvert}{\rvert}

\definecolor{ao}{rgb}{0.0, 0.5, 0.0}
\definecolor{brown}{rgb}{0.65, 0.16, 0.16}

\setitemize{noitemsep,topsep=0pt,parsep=0pt,partopsep=0pt}

\allowdisplaybreaks
\numberwithin{equation}{section}
\numberwithin{theorem}{section}
\numberwithin{corollary}{section}
\numberwithin{proposition}{section}
\numberwithin{lemma}{section}
\numberwithin{remark}{section}

\allowdisplaybreaks

\newcommand\tsum{\textstyle\sum\nolimits}
\newcommand\tprod{\textstyle\prod\nolimits}

\usepackage[margin=1.0in]{geometry}

\ifpdf
\hypersetup{
  pdftitle={Block Coordinate Policy Mirror Descent},
  pdfauthor={G. Lan, ~Y. Li, ~T. Zhao}
}
\fi

\begin{document}

\setlength{\abovedisplayskip}{6pt}
\setlength{\belowdisplayskip}{6pt}
\setlength{\abovedisplayshortskip}{4pt}
\setlength{\belowdisplayshortskip}{4pt}


{
\maketitle
}

\begin{abstract}
In this paper, we present a new policy gradient (PG) methods,  namely the block policy mirror descent (BPMD) method for solving a class of regularized reinforcement learning (RL) problems with (strongly)-convex regularizers. Compared to the traditional PG methods with a batch update rule, which visits and updates the policy for every state, BPMD method has cheap  per-iteration computation via a partial update rule that performs the  policy update on a sampled state. Despite the nonconvex nature of the problem and a partial update rule, we provide a unified analysis for several sampling schemes, and show that BPMD achieves fast linear convergence to the global optimality. In particular, uniform sampling leads to comparable worst-case total computational complexity as batch PG methods. A necessary and sufficient condition for convergence with on-policy sampling is also identified. With a hybrid sampling scheme, we further show that BPMD enjoys potential instance-dependent acceleration, leading to improved dependence on the state space and consequently outperforming batch PG methods. We then extend BPMD methods to the stochastic setting, by utilizing stochastic first-order information constructed from samples. With a generative model, $\tilde{\cO}(\abs{\cS} \abs{\cA} /\epsilon)$ (resp. $\tilde{\cO}(\abs{\cS} \abs{\cA} /\epsilon^2)$) sample complexities are established for the strongly-convex  (resp. non-strongly-convex)  regularizers, where $\epsilon$ denotes the target accuracy. To the best of our knowledge, this is the first time that block coordinate descent methods have been developed and analyzed for policy optimization in reinforcement learning, which provides a new perspective on solving large-scale RL problems.
\end{abstract}

\vspace{-0.1in}

\vspace{-0.1in}
\vspace{-0.1in}


\vspace{-0.05in}
\section{Introduction}

We consider a discrete time Markov decision process (MDP)  denoted by the quintuple $\cM = (\cS, \cA, \cP, C, \gamma)$, where $\cS$ is the finite state space, $\cA$ is the finite action space, 
$\cP: \cS \times \cS \times \cA \to [0,1]$
 is the transition kernel, $C$ is the cost function, and $\gamma \in (0,1)$ is the discount factor. 
We also refer to $\cP$ as the environment parameters.

A randomized, stationary policy $\pi : \cS \to \Delta_{\cA}$ is a mapping from the state space into the probability simplex over $\cA$, and we denote the set of all such policies by $\Pi$.
At any timestep $t$,  the policy explicitly governs what action to be made given the current state $s_t$, by $a_t \sim \pi(\cdot| s_t)$. 
Then a cost $C(s_t, a_t)$ is incurred, followed by the transition to the next state $s_{t+1} \sim \cP(\cdot|s_t, a_t)$. 
The decision process is then repeated iteratively at future timesteps. 

We consider the general scenario where costs are allowed to be policy dependent. Specifically, we assume the cost  $C(s, a)$ decomposes as $C(s, a) = c(s, a) + h^{\pi}(s)$ for all state-action pair $(s,a) \in \cS \times \cA$, with some function $c: \cS \times \cA \to \RR$ and $h: \Pi \times \cS \to \RR$.
The introduction of the regularization term $h^{\pi}(s)$ allows one to induce additional desirable properties to the optimal policy beyond mere cost-minimization (e.g., policy constraint \cite{lan2021policy}, sufficient exploration \cite{mnih2016asynchronous, peters2010relative, schulman2015trust}). 
We assume $h^{\pi}(s)$  is a closed convex function w.r.t. the policy $\pi(\cdot|s)$ for any $s \in \cS$.
That is, there exists some $\mu \geq 0$ such that, for any policy $\pi, \pi' \in \Pi$, 
\begin{align*}
\textstyle
h^\pi(s) - h^{\pi'}(s) - \inner{\partial h^{\pi'}(s)}{\pi(\cdot|s) - \pi'(\cdot|s)} \geq \mu D_{\pi'}^{\pi} (s), ~~ \forall s \in \cS,
\end{align*}
where $\partial h^{\pi'}(s)$ denotes a subgradient of  $h^{\pi'}(s)$ w.r.t. $\pi'(\cdot | s)$, and $D_{\pi'}^{\pi}(s)$ denotes the Kullback-Leibler (KL) divergence between  $\pi(\cdot|s)$ and $\pi'(\cdot|s)$.
Clearly, the generalized class of MDPs  covers the standard MDP as a special case with $h \equiv 0$.
We say that $h$ is a strongly-convex regularizer if $\mu > 0$.

For a given policy $\pi$, we measure its performance by its value function $V^{\pi}: \cS \to \RR$, defined by 
\begin{align}
\textstyle
V^{\pi} (s) = \EE \sbr{\sum_{t=0}^\infty \gamma^t \rbr{c(s_t, a_t) + h^\pi(s_t) }\big| s_0 = s, a_t \sim \pi(\cdot|s_t), s_{t+1} \sim \cP(\cdot|s_t,a_t)  }.
\end{align}
We also define its state-action value function (Q-function) $Q^\pi: \cS \times \cA \to \RR$, as 
\begin{align}
\textstyle
Q^{\pi} (s, a) = \EE \sbr{\sum_{t=0}^\infty \gamma^t \rbr{ c(s_t, a_t) + h^\pi(s_t)} \big| s_0 = s, a_0 = a, a_t \sim \pi(\cdot|s_t), s_{t+1} \sim \cP(\cdot|s_t,a_t)  }.
\end{align}
One can readily verify, from the definition of $V^{\pi}$ and $Q^{\pi}$, that  
\begin{align}
\textstyle
\label{mdp:relatiion_vq}
V^{\pi}(s) = \inner{Q^{\pi} (s, \cdot)}{\pi(\cdot|s)}, ~~
Q^{\pi}(s, a) = c(s, a) + h^{\pi}(s) + \gamma \sum_{s' \in \cS} \cP(s' | s,a) V^{\pi}(s').
\end{align}
The  planning objective of the MDP $\cM$ is to find an optimal policy $\pi^*$ satisfying the following property, 
\begin{align}\label{obj:state_wise}
V^{\pi^*}(s) \leq V^{\pi} (s), ~~ \forall \pi(\cdot| s) \in \Delta_{\cA}, ~~ \forall s \in \cS.
\end{align}
Note that the existence of an optimal policy $\pi^*$  simultaneously minimizing the value for every state is well known in literature \cite{puterman2014markov}. Hence we can succinctly reformulate \eqref{obj:state_wise} into a single-objective optimization problem
\begin{align}
\textstyle
\label{eq:mdp_single_obj_raw}
\min_{\pi} \EE_{\nu} \sbr{V^{\pi}(s)}, ~~ \mathrm{s.t.} ~~ \pi(\cdot|s) \in \Delta_{\cA}, \forall s \in \cS,
\end{align}
where $\nu$ can be an arbitrary distribution defined over $\cS$. 
As pointed out by recent literature \cite{liu2019neural}, setting $\nu$ as the stationary state distribution induced by the optimal policy $\pi^*$, denoted by $\nu^*$, can simplify the analyses of various algorithms. 
In such a case, \eqref{eq:mdp_single_obj_raw} becomes 
\begin{align}
\textstyle
\label{eq:mdp_single_obj}
\min_{\pi} \cbr{f(\pi) \coloneqq \EE_{\nu^*} \sbr{V^{\pi}(s)}}, ~~ \mathrm{s.t.} ~~ \pi(\cdot|s) \in \Delta_{\cA}, \forall s \in \cS.
\end{align}
It is worth mentioning that even though the objective $f(\pi)$ explicitly depends on the unknown distribution $\nu^*$, the algorithms we develop for solving \eqref{eq:mdp_single_obj} do not necessarily require the information of $\nu^*$.

\vspace{0.1in}
{\bf Related Literature.}
Solving \eqref{obj:state_wise} or \eqref{eq:mdp_single_obj} has been long studied in the literature of dynamic programming (DP) and reinforcement learning (RL). 
When the environment parameters  are known in advance, DP based methods including the celebrated value iteration (VI) \cite{bellman1966dynamic} and (modified) policy iteration (PI) \cite{howard1960dynamic} converge to the optimal value function or policy at linear rates \cite{puterman2014markov}.
In comparison, RL methods focus on the scenario when the information of the environment parameters is only available through the form of interactions, and the improvement of policy is made upon information learned from  collected samples \cite{sutton2018reinforcement}.

There has been a surge of interests in designing efficient first order algorithms for directly searching  the optimal policy $\pi^*$ in the RL literature \cite{agarwal2020optimality, cen2021fast, lillicrap2015continuous, schulman2015trust, shani2020adaptive}, despite the objective \eqref{eq:mdp_single_obj} being  non-convex w.r.t. the policy   $\pi$ \cite{agarwal2020optimality}.
In a nutshell, all these methods share a common similarity that gradient information of objective \eqref{eq:mdp_single_obj} is used for policy improvement, and hence termed policy gradient (PG) methods.
In the simplest form, basic policy gradient method directly updates the policy by performing a  projected gradient descent step with the simplex constraint defined in \eqref{eq:mdp_single_obj},
 and converges sublinearly  with exact gradients \cite{agarwal2020optimality}.
Natural policy gradient method \cite{kakade2001natural} performs gradient update with pre-conditioning demonstrates and demonstrates better dimensional dependence \cite{agarwal2020optimality}.
Results with strong linear convergence of PG methods have been discussed in \cite{bhandari2020note,cen2021fast,khodadadian2021linear}.
The analyses therein heavily exploit the contraction properties of the Bellman optimality condition, making their extensions to the stochastic setting unclear without additional assumptions \cite{cen2021fast}.
Connections between PG methods and the classical mirror descent method \cite{beck2003mirror, nemirovski2009robust, nemirovskij1983problem} have also been established (e.g., TRPO \cite{schulman2015trust, shani2020adaptive},  REPS \cite{neu2017unified, peters2010relative}). 
Until recently, \cite{lan2021policy} proposes the policy mirror descent method and its stochastic variants, and establishes linear convergence in both deterministic and stochastic settings, as well as optimal sampling complexity bounds w.r.t. target accuracy (optimality gap).

It is worth mentioning that
all the aforementioned DP and  PG methods update the policy in a batch fashion, i.e., 
at each iteration, the update of policy needs to be made for every state in the state space. 
Accordingly, the PG methods also need to evaluate the first-order information of the policy for every state.
We will refer to PG methods with such a batch update rule as  batch PG methods.
A direct challenge rising as a consequence of  the batch update rule is that,
for problems with large state space, sweeping every state at each iteration quickly becomes a bottleneck for efficient computation. 

To address such a challenge of scale, DP methods typically resort to sequential or parallel (and asynchronous) update order \cite{bertsekas2015parallel}.
It is well known that Gauss-Seidel VI \cite{puterman2014markov} still converges linearly to the optimal value function.
By generalizing the relationship between the value iteration and policy iteration discussed in \cite{puterman2014markov}, one can  further establish  linear convergence of Gauss-Seidel PI.
In addition, while asynchronous VI converges linearly under bounded communication delay \cite{bertsekas2015parallel},   the convergence of asynchronous PI is significantly more fragile, with asymptotic convergence established in \cite{bertsekas2010distributed}.

In terms of RL methods, handling the large state space has been mostly discussed by considering function approximation, which involves paramterizing policy and value functions within a pre-specified function class \cite{agarwal2020optimality, cayci2021linear, liu2019neural, wang2019neural}. 
The introduction of function approximation mainly aims to reduce the dependence of computations associated with each policy evaluation and policy improvement step on the potentially huge state space.
Specifically, at each iteration, gradient of the policy parameters is calculated and used to update the policy, instead of updating the policy explicitly for each state.
On the other hand, the convergence of policy gradient methods involving function approximation becomes much more subtle than their tabular counterparts, due to the approximation bias of the first-order information (i.e., state-action value function) \cite{lan2021policy} and the inexact policy improvement step.
Existing convergence analyses for PG methods with function approximation often require restricted technical conditions (e.g., completeness of the value function class \cite{liu2019neural, wang2019neural}) depending on the function class adopted.
In general settings, the global convergence of PG methods with function approximation still remains unclear \cite{agarwal2020optimality}.
Our developed methods provides a new perspective on solving large-scale RL problems in addition to prior developments: 
one can perform policy updates (and consequently evaluation) on a subset of states at each iteration, and thus significantly reducing the per-iteration computation.
Moreover, an improved total computational complexity can be potentially attained in an instance-dependent fashion.


%


\vspace{0.1in}
{\bf  Main Contributions.} First, we propose the block policy mirror descent method (BPMD), which at each iteration, updates the policy of a single state sampled from some (time-dependent) distribution. 
We show that single-state policy update also implies cheap policy evaluation, and hence leads to low per-iteration computation. 
For the class of exploratory sampling schemes, 
for which the support of the sampling distribution contains that of $\nu^*$,
we establish the linear convergence of BPMD for strongly-convex regularizers, and both linear and sublinear for non-strongly-convex regularizers, with proper stepsizes. 
For the class of static and exploratory sampling schemes, we further derive simple stepsize rules with the aforementioned convergences. 
When adopting the uniform sampling, we show BPMD attains the same total computational complexity as the fastest batch PG methods in literature, while having a significant lower per-iteration cost, and hence enjoys several practical benefits, including finer granularity and low-cost early stopping.  
We also discuss on-policy sampling, where the state is sampled by following the trajectory of the current policy.
We establish the convergence with on-policy sampling provided the generated trajectory is exploratory.  An example is constructed to further demonstrate the necessity of the exploratory condition.

Second, we study the role of sampling distributions in impacting the efficiency of BPMD. 
To this end, we first establish the adaptive convergence of BPMD with any static sampling scheme,  by quantifying the interplay between sampling distribution and the (unknown) $\nu^*$.
This first allows obtaining approximate convergence of BPMD when the exploratory condition only holds partially. 
Moreover, it is particularly useful for deriving tighter convergence bounds of BPMD in the initial phase. 
Based on this observation, we then propose a hybrid sampling scheme, which first uses an approximate estimate of $\nu^*$ to sample the state and then switches to the uniform sampling scheme. 
We show using the hybrid sampling enjoys potential acceleration over the uniform sampling, provided $\nu^*$ having a light tail, and its estimation quality inversely depending on the tail mass. 
As a concrete application, we show that with a polynomial tail, the iteration complexity of the hybrid sampling  improves over the uniform sampling by a factor of $\Omega(\sqrt{\abs{\cS}})$, for finding a policy with an optimality gap $\Omega(\mathrm{poly}(\abs{\cS}^{-1}))$.
We conclude by discussing the connections and differences to the acceleration effect of the block coordinate descent (BCD) method \cite{beck2013convergence,dang2015stochastic, leventhal2010randomized, li2017faster, lu2015complexity, nesterov2012efficiency, richtarik2014iteration, tseng2001convergence, wright2015coordinate, xu2015block}.

Third, we develop the stochastic BPMD (SBPMD) method when the environment parameters are unknown, which utilizes collected trajectories to form a stochastic estimate of the first-order information.
We establish convergences of SBPMD for both strongly-convex and non-strongly convex regularizers, and the iteration complexities naturally match that of stochastic batch PG methods \cite{lan2021policy} with a multiplicative factor of $\abs{\cS}$. 
When a generative model is available, we consequently establish $\tilde{\cO}(\abs{\cS}\abs{\cA} / \epsilon)$ (resp. $\tilde{\cO}(\abs{\cS} \abs{\cA}/\epsilon^2)$) sample complexities for strongly-convex (resp. non-strongly-convex) regularizers.
These sample complexities match that of the best stochastic batch PG methods, in terms of dependencies on the state space, action space, and even the discount factor.
Adding to the competitive worst-case performance, we further highlight that each iteration requires a number of  samples $\abs{\cS}$-times smaller than that of batch PG methods.

To the best of our knowledge, the proposed BPMD methods are the first policy gradient type method that performs policy updates on a subset of states at each iteration, and thus can be computationally favorable over batch PG methods in the presence of large state space. 
The analysis focuses on the scenario where a single state is updated, and admits a straightforward extension to a multi-state update rule, which can further exploit the benefits of parallel computation.

The rest of the paper is structured as follows. 
Section \ref{sec:bpmd} proposes and analyzes the convergence of BPMD with time-dependent sampling, static sampling, and on-policy sampling.
Section \ref{sec_effect_sampling} focuses on static sampling, and discusses the role of sampling distribution in detail, resulting into a hybrid sampling scheme with instance-dependent acceleration.
Section \ref{sec:stochastic} proposes the stochastic BPMD and establishes noise conditions with which we certify convergence.
Sample complexities with a generative model are consequently developed.
Numerical studies are presented in Section \ref{sec:exp} to demonstrate the benefits of BPMD methods.

\vspace{0.1in}
{\bf Notation and Terminology.}
For any policy $\pi$ and any $s_0 \in \cS$, we define the discounted state visitation measure 
$d_{s_0}^\pi(s) = (1-\gamma) \sum_{t=0}^\infty \gamma^t \PP^{\pi}(s_t = s | s_0)$, where $\PP^{\pi}(s_t = s|s_0)$ denotes the probability of reaching state $s$ at time $t$ following policy $\pi$,  when starting at state $s_0$.
We also denote the stationary state distribution of policy $\pi$ by $\nu^{\pi}$, which satisfies $\nu^\pi(s) = \sum_{s' \in \cS} \nu^{\pi}(s') \PP^{\pi}(s|s')$ where $\PP^{\pi}(s|s') =  \sum_{a \in \cA} \pi(a|s') \PP(s|s', a)$.

For a strictly convex function with $w$ domain containing $\Delta_{\cA}$, we define 
\begin{align}
\label{def:kl_bregman}
D^{\pi}_{\pi'} (s) = w(\pi(\cdot|s)) - w(\pi'(\cdot|s)) - \inner{\nabla w(\pi'(\cdot|s))}{\pi(\cdot|s) - \pi'(\cdot|s) },
\end{align}
where $\nabla w(\pi'(\cdot|s))$ denotes the subgradient of $w(\pi'(\cdot|s))$ w.r.t. $\pi'(\cdot|s)$. 
This corresponds to the Bregman divergence applied to the policy with a distance-generating function being $w$.
We denote $\phi(\pi^*, \pi) = \sum_{s \in \cS} \nu^*(s) D^{\pi^*}_{\pi}(s) $.
Though the analyses hold for general choices of $w$, for the sake of notation, 
unless stated otherwise, we focus on $w(\pi(\cdot|s)) = \sum_{a \in \cA} \pi(a|s) \log \rbr{\pi(a|s)}$, i.e.,  the negative entropy function.

\section{Deterministic Block Policy Mirror Descent}\label{sec:bpmd}

In this section, we present the deterministic block policy mirror descent  (BPMD) method and establish its convergence properties for both strongly-convex and non-strongly-convex regularizers.
The BPMD method assumes the capability to sample states from a distribution defined over the state space,
which can change across different iterations, 
 and we establish the global convergence result under a fairly general condition that these sampling distributions have support no smaller than that of $\nu^*$. 
The necessity is further demonstrated by an example, in which policy optimization fails in the absence of this condition.
Convergence of BPMD where this condition holds approximately will be discussed in detail in Section \ref{sec_effect_sampling}.
Towards the end of this section, we will focus on two natural sampling schemes, 
and discuss their comparative advantages in terms of computational efficiency and practicality.

%
%
%
%


\subsection{BPMD and its General Convergence Properties}\label{subsec:BPMD}

We  present details of the BPMD method in Algorithm \ref{alg:BPMD}.
The key difference between BPMD and batch PG methods (e.g., \cite{agarwal2020optimality, cen2021fast, lan2021policy}) is that at each iteration $k$, BPMD samples a state $s_k$ from $\cS$ by following some pre-specified distribution $\rho_k \in \Delta_{\cS}$ defined over $\cS$, and then only updates the policy at the sampled state $s_k$.

\begin{algorithm}[htb!]
    \caption{The block  policy mirror descent (BPMD) method}
    \label{alg:BPMD}
    \begin{algorithmic}
    \STATE{\textbf{Input:} Initial uniform policy $\pi_0$,  and stepsizes $\{\eta_k\}_{k\geq 0}$, and  sampling distributions $\cbr{\rho_k}_{k \geq 0} \subset \Delta_{\cS}$.}
    \FOR{$k=0, 1, \ldots$}
	\STATE{Sample $s_k \sim \rho_k$.}
	\STATE{Update policy:
	\vspace{-0.15in}
	\begin{align}\label{eq:bpmd_update}
	\pi_{k+1} (\cdot| s) = \begin{cases}
	\argmin_{p(\cdot|s_k)  \in \Delta_{\cA}} \eta_k \sbr{ \inner{Q^{\pi_k}(s_k, \cdot)}{ p(\cdot| s_k )} + h^p(s_k) } + D^{p}_{\pi_k}(s_k), ~~~ & s = s_k, \\
	\pi_k(\cdot| s) , ~~~ & s \neq s_k.
	\end{cases}
	\end{align}}
    \ENDFOR
    \end{algorithmic}
\end{algorithm}

We next establish some generic  properties of each update in BPMD.
To proceed, we start with the following performance difference lemma adopted from \cite{lan2021policy}, which generalizes the results in \cite{kakade2002approximately}. 
Lemma \ref{lemma:performance_diff} establishes the connection between the  value functions of two policies.

\begin{lemma}\label{lemma:performance_diff}
For any pair of policies $\pi, \pi'$, 
\begin{align}
V^{\pi'}(s) - V^{\pi}(s) = \tfrac{1}{1-\gamma} \EE_{s' \sim d_s^{\pi'}} \sbr{\inner{Q^{\pi}( s', \cdot) }{ \pi'(\cdot|s') - \pi(\cdot| s') }  + h^{\pi'}(s') - h^{\pi}(s') }. \label{eq:performance_diff}
\end{align}
\end{lemma}
\begin{proof}
The claim follows from Lemma 2 in \cite{lan2021policy}, and the identity $\inner{A^{\pi}(s, \cdot)}{\pi'(\cdot|s) } = \inner{Q^{\pi}(s, \cdot)}{\pi'(\cdot|s) -\pi(\cdot|s)}$ due to \eqref{mdp:relatiion_vq} and the definition $A^{\pi}(s,a) = Q^{\pi}(s,a) - V^{\pi}(s,a)$. 
\end{proof}

The next lemma provides a characterization on the policy update of BPMD.

\begin{lemma}\label{lemma:three_point}
For any $p(\cdot|s_k) \in \Delta_{\abs{\cA}}$, the policy pair $(\pi_k, \pi_{k+1})$ in BPMD satisfies
\begin{align}\label{ineq:three_point}
\eta_k \big[ \inner{Q^{\pi_k}(s_k, \cdot)}{ \pi_{k+1} (\cdot| s_k )- p (\cdot| s_k)} & + h^{\pi_{k+1}}(s_k)  - h^p(s_k) \big]  + D^{\pi_{k+1}}_{\pi_k} (s_k) 
  \\ 
 & \leq D^p_{\pi_k} (s_k) - (1+ \eta_k \mu) D^p_{\pi_{k+1}} (s_k) . \nonumber
\end{align}
\end{lemma}

\begin{proof}
From the optimality condition of the update \eqref{eq:bpmd_update}, we have 
\begin{align*}
 \inner{\eta_k \sbr{Q^{\pi_k}(s_k, \cdot) + \partial h^{\pi_{k+1}}(s_k)} + \partial D^{\pi_{k+1}}_{\pi_k} (s_k) }{\pi_{k+1}(\cdot|s_k) - p(\cdot|s_k) } \leq 0,
\end{align*}
where $ \partial D^{\pi_{k+1}}_{\pi_k} (s_k)$ denotes the subgradient of $ D^{\pi_{k+1}}_{\pi_k}(s_k)$ w.r.t. $\pi_{k+1}(\cdot|s_k)$. 
The claim follows by noting that
\begin{align*}
\inner{\partial h^{\pi_{k+1}}(s_k)} {\pi_{k+1}(\cdot|s_k) - p(\cdot|s_k) } \geq h^{\pi_{k+1}}(s_k)  - h^p(s_k) + \mu  D_{\pi_{k+1}}^p (s_k),
\end{align*} and the fact  that
$
 \inner{\partial D^{\pi_{k+1}}_{\pi_k} (s_k) }{\pi_{k+1}(\cdot|s_k) - p(\cdot|s_k) } = -D^p_{\pi_k} (s_k) + D^p_{\pi_{k+1}} (s_k) + D^{\pi_{k+1}}_{\pi_k} (s_k) .
$
\end{proof}

Combining Lemma \ref{lemma:performance_diff} and \ref{lemma:three_point}, we can characterize the progress of value function on the updated state $s_k$.
In particular, we show that the value for every  other state that does not get updated in the current iteration never deteriorates from its current value.
Such an observation also draws an intimate connection to the classical policy improvement theorem \cite{puterman2014markov}. 

\begin{lemma}\label{lemma:step_progress}
At each iteration $k$, with $\eta_k > 0$, the progress of value at each state satisfies
\begin{align}
& V^{\pi_{k+1}}(s) \leq V^{\pi_{k}}(s), ~~\forall s \in \cS,  \label{ineq:value_decrease} \\
& V^{\pi_{k+1}}(s_k) - V^{\pi_{k}}(s_k) \leq  \inner{Q^{\pi_k}(s_k, \cdot) }{\pi_{k+1}(\cdot|s_k) - \pi_k(\cdot|s_k) } + h^{\pi_{k+1}}(s_k)- h^{\pi_{k}}(s_k) < 0 . \label{ineq:progress_lb}
\end{align}
\end{lemma}
\begin{proof}
From Lemma \ref{lemma:performance_diff}, we have 
\begin{align}\label{eq_value_progress_at_other_state_raw}
V^{\pi_{k+1}}(s) - V^{\pi_{k}}(s) = \tfrac{1}{1-\gamma} 
\EE_{s' \sim d_s^{\pi_{k+1}}} \sbr{ \inner{Q^{\pi_k}(s', \cdot)}{ \pi_{k+1}(\cdot|s')- \pi_k(\cdot|s')} + h^{\pi_{k+1}}(s') - h^{\pi_{k}}(s')}.
\end{align}
Note that  $\pi_{k+1}(\cdot| s) = \pi_k(\cdot |s)$ for any $s \neq s_k$.
In addition, by taking $p(\cdot|s_k)  = \pi_k(\cdot|s_k)$ in \eqref{ineq:three_point}, we obtain
\begin{align}
& \inner{Q^{\pi_k}(s_k, \cdot) }{ \pi_{k+1}(\cdot|s_k) - \pi_k(\cdot|s_k) } + h^{\pi_{k+1}}(s_k) - h^{\pi_{k}}(s_k) \label{ineq:bpmd_neg} \\
 \leq &  - (1+ \eta_k \mu) D^{\pi_k}_{\pi_{k+1}} (s_k) -  D^{\pi_{k+1}}_{\pi_k} (s_k)  <  0.  \nonumber
\end{align}
Hence  $V^{\pi_{k+1}}(s) - V^{\pi_{k}}(s) \leq 0$ for all $s \in \cS$ and \eqref{ineq:value_decrease} is proved. 
Finally, \eqref{ineq:progress_lb} follows from  
\begin{align*}
V^{\pi_{k+1}}(s_k)- V^{\pi_{k}}(s_k) & = \tfrac{1}{1-\gamma}  d_{s_k}^{\pi_{k+1}} (s_k)  
\sbr{ \inner{Q^{\pi_k}(s_k, \cdot) }{ \pi_{k+1}(\cdot|s_k) -\pi_k(\cdot|s_k) } + h^{\pi_{k+1}}(s_k) - h^{\pi_{k}}(s_k)} \\
& \leq  \inner{Q^{\pi_k}(s_k, \cdot) }{\pi_{k+1}(\cdot|s_k)- \pi_k(\cdot|s_k) } + h^{\pi_{k+1}}(s_k) - h^{\pi_{k}}(s_k) < 0,
\end{align*}
where the last inequality follows from \eqref{ineq:bpmd_neg}, and the fact that $d_{s_k}^{\pi_{k+1}}(s_k) \geq 1-\gamma$. 
\end{proof}

Combining Lemma \ref{lemma:performance_diff}, \ref{lemma:three_point}, and \ref{lemma:step_progress}, we can establish the generic convergence property of BPMD, presented in the following lemma.

\begin{lemma}\label{lemma:bpmd_generic_recursion}
Suppose the sampling distribution $\rho_k$ satisfies  
\begin{align}\label{state_sampling_condition} 
\rho_k^\dagger \coloneqq \min \cbr{\rho_k(s): \nu^*(s) > 0 } > 0,
\end{align} 
and let $\cS_k = \mathrm{supp}(\rho_k)$ denote the support of $\rho_k$. 
Then the policy pair $(\pi_k, \pi_{k+1})$ generated by BPMD satisfies 
\begin{align}
 \EE_{s_k} \sbr{f(\pi_{k+1}) - f(\pi^*)}   \label{eq:bpmd_generic_recursion}
 \leq &
\rbr{1 - (1-\gamma) \rho^\dagger_k} \sbr{f(\pi_k) - f(\pi^*) }
+ \rho_k^\dagger \sbr{\tfrac{1}{\eta_k} + \mu (1-\rho^\dagger_k) }    \tsum_{s \in \cS_k}  \tfrac{\nu^*(s)}{\rho_k(s)}  D^{\pi^*}_{\pi_k}(s) \\
& -  \rho_k^\dagger \sbr{\tfrac{1}{\eta_k} + \mu }    \EE_{s_k} \sbr{   \tsum_{s \in \cS_k}  \tfrac{\nu^*(s)}{\rho_k(s)} D^{\pi^*}_{\pi_{k+1}}(s)} \nonumber .
\end{align}
\end{lemma}
\begin{proof}
By taking $p(\cdot|s_k) = \pi^*(\cdot|s_k)$ in \eqref{ineq:three_point},   we have 
\begin{align*}
& \eta_k \big[ \inner{Q^{\pi_k}(s_k, \cdot)}{ \pi_{k+1}(\cdot| s_k )- \pi^* (\cdot| s_k)}
+  h^{\pi_{k+1}}(s_k)  - h^{\pi^*}(s_k) \big]  + D^{\pi_{k+1}}_{\pi_k}(s_k) \\
\leq 
& D^{\pi^*}_{\pi_k}(s_k)  - (1 + \eta_k \mu)D^{\pi^*}_{\pi_{k+1}}(s_k) .
\end{align*}
Combining the above relation with \eqref{ineq:progress_lb}, we obtain 
\begin{align}
 \eta_k \big[\inner{Q^{\pi_k}(s_k, \cdot) }{\pi_k(\cdot|s_k) - \pi^*(\cdot| s_k)  } & + h^{\pi_k}(s_k)  - h^{\pi^*}(s_k)\big]  + 
\eta_k \sbr{V^{\pi_{k+1}}(s_k) - V^{\pi_{k}}(s_k)}  \label{scBPMD:value_at_update} 
\\ & + D^{\pi_{k+1}}_{\pi_k}(s_k)  
\leq 
D^{\pi^*}_{\pi_k}(s_k)  - (1+ \eta_k \mu) D^{\pi^*}_{\pi_{k+1}}(s_k) . \nonumber 
\end{align}

There are several terms in the above inquality that take the form of $G^{\pi_{k+1}}(s_k)$, the coupling between $\pi_{k+1}$ and $s_k$ prevents directly taking expectation of these terms with respect to $s_k$. 
Our next few steps aim to decouple this dependence before taking the expectation.

First, multiplying both sides of \eqref{scBPMD:value_at_update} by $\nu^*(s_k) / \rho_k(s_k)$, we have 
\begin{align}
 \eta_k \tfrac{\nu^*(s_k)}{\rho_k(s_k)}  \big[\inner{Q^{\pi_k}(s_k, \cdot) }{\pi_k(\cdot|s_k) - \pi^*(\cdot| s_k)  }  + h^{\pi_k}(s_k)  - h^{\pi^*}(s_k)\big]  + 
\eta_k   \tfrac{\nu^*(s_k)}{\rho_k(s_k)}  \sbr{V^{\pi_{k+1}}(s_k) - V^{\pi_{k}}(s_k)}  \label{scBPMD:value_at_update_2} 
\\  +    \tfrac{\nu^*(s_k)}{\rho_k(s_k)}  D^{\pi_{k+1}}_{\pi_k}(s_k)  
\leq 
  \tfrac{\nu^*(s_k)}{\rho_k(s_k)}  D^{\pi^*}_{\pi_k}(s_k)  - (1+ \eta_k \mu)    \tfrac{\nu^*(s_k)}{\rho_k(s_k)}  D^{\pi^*}_{\pi_{k+1}}(s_k) . \nonumber
\end{align}
Next, we make the following simple yet important identity,
\begin{align}
D^{\pi^*}_{\pi_k}(s)  - (1 + \eta_k \mu) D^{\pi^*}_{\pi_{k+1}}(s) = -\eta_k \mu  D^{\pi^*}_{\pi_{k}}(s) , ~~\forall s \neq s_k. \label{scBPMD:d_at_other}
\end{align}
Let us denote $\cS_k^c = \cS_k \setminus \cbr{s_k}$.
Then multiplying both sides of above identity by $\nu^*(s)/\rho_k(s)$,  and  summing up over $\cS_k^c$ together  with \eqref{scBPMD:value_at_update_2}, we obtain 
\begin{align}
 \eta_k \tfrac{\nu^*(s_k)}{\rho_k(s_k)}  \big[\inner{Q^{\pi_k}(s_k, \cdot) }{\pi_k(\cdot|s_k) - \pi^*(\cdot| s_k)  }  + h^{\pi_k}(s_k)  - h^{\pi^*}(s_k)\big]  + 
  \eta_k \tfrac{\nu^*(s_k)}{\rho_k(s_k)}  \sbr{V^{\pi_{k+1}}(s_k) - V^{\pi_{k}}(s_k)} \label{scBPMD:value_at_update_3} 
\\  
\leq 
  \tsum_{s \in \cS_k }  \tfrac{\nu^*(s)}{\rho_k(s)}  D^{\pi^*}_{\pi_k}(s)  - (1+ \eta_k \mu)    \tsum_{s\in \cS_k}  \tfrac{\nu^*(s)}{\rho_k(s)}  D^{\pi^*}_{\pi_{k+1}}(s)
  + \eta_k \mu  \tsum_{s \in \cS_k^c }  \tfrac{\nu^*(s)}{\rho_k(s)}   D^{\pi^*}_{\pi_{k}}(s)
   .  \nonumber 
\end{align}
We then multiply both sides of \eqref{ineq:value_decrease} by $\nu^*(s)/\rho_k(s)$, and sum up over $s \in  \cS_k^c$ together with \eqref{scBPMD:value_at_update_3}, which gives
\begin{align*}
 \eta_k \tfrac{\nu^*(s_k)}{\rho_k(s_k)}  \big[\inner{Q^{\pi_k}(s_k, \cdot) }{\pi_k(\cdot|s_k) - \pi^*(\cdot| s_k)  }  + h^{\pi_k}(s_k)  - h^{\pi^*}(s_k)\big]  + 
  \eta_k \tsum_{s\in \cS_k} \tfrac{\nu^*(s)}{\rho_k(s)}  \sbr{V^{\pi_{k+1}}(s) - V^{\pi_{k}}(s)} \nonumber 
\\  
\leq 
  \tsum_{s \in \cS_k}  \tfrac{\nu^*(s)}{\rho_k(s)}  D^{\pi^*}_{\pi_k}(s)  - (1+ \eta_k \mu)    \tsum_{s\in \cS_k}  \tfrac{\nu^*(s)}{\rho_k(s)}  D^{\pi^*}_{\pi_{k+1}}(s)
    + \eta_k \mu  \tsum_{s \in \cS_k^c }  \tfrac{\nu^*(s)}{\rho_k(s)}   D^{\pi^*}_{\pi_{k}}(s) . 
\end{align*}

We can now take expectation w.r.t. $s_k \sim \rho_k$, which gives
\begin{align}
 \eta_k \underbrace{ \tsum_{s \in \cS_k} \nu^*(s)  \big[\inner{Q^{\pi_k}(s, \cdot) }{\pi_k(\cdot|s) - \pi^*(\cdot| s )  }  + h^{\pi_k}(s)  - h^{\pi^*}(s)\big] }_{(T_1)} + 
  \eta_k \tsum_{s\in \cS_k} \tfrac{\nu^*(s)}{\rho_k(s)}  \EE_{s_k} \sbr{V^{\pi_{k+1}}(s) - V^{\pi_{k}}(s)} \nonumber 
\\  
\leq 
  \tsum_{s \in \cS_k}  \tfrac{\nu^*(s)}{\rho_k(s)}  D^{\pi^*}_{\pi_k}(s)  - (1+ \eta_k \mu)    \EE_{s_k}  \big[ \tsum_{s\in \cS_k}  \tfrac{\nu^*(s)}{\rho_k(s)} D^{\pi^*}_{\pi_{k+1}}(s)\big]
    + \eta_k \mu  \underbrace{\EE_{s_k} \tsum_{s \in \cS_k^c }  \tfrac{\nu^*(s)}{\rho_k(s)}   D^{\pi^*}_{\pi_{k}}(s)}_{(T_2)}
   . 
   \label{scBPMD:value_at_update_4}
\end{align}
To simplify the above relation, we make the following observations.
First, we have 
\begin{align}
(1-\gamma)\sbr{ f(\pi_k) - f(\pi^*)} & = (1-\gamma) \tsum_{s \in \cS} \nu^*(s) \sbr{V^{\pi_k}(s) - V^*(s)}  \nonumber \\
&\overset{(a)}{=} - \tsum_{s \in \cS}  \nu^*(s) \EE_{s' \sim d_s^{\pi^*}} \sbr{\inner{Q^{\pi_k}(\cdot | s') }{ \pi^*(\cdot|s') - \pi_k(\cdot|s')} 
+ h^{\pi^*}(s') - h^{\pi_k}(s')}  \nonumber  \\
& \overset{(b)}{=}  \tsum_{s' \in \cS}  \nu^*(s')  \sbr{\inner{Q^{\pi_k}(\cdot | s') }{ \pi_k(\cdot|s') - \pi^*(\cdot|s')} 
+ h^{\pi_k}(s') - h^{\pi^*}(s')}  \label{vi_pmd}\\
& \overset{(c)}{=} 
\tsum_{s' \in \cS_k}  \nu^*(s')  \sbr{\inner{Q^{\pi_k}(\cdot | s') }{ \pi_k(\cdot|s') - \pi^*(\cdot|s')} 
+ h^{\pi_k}(s') - h^{\pi^*}(s')}
 = (T_1), \nonumber 
\end{align}
where $(a)$ follows from applying \eqref{eq:performance_diff} with $(\pi', \pi) = (\pi^*, \pi_k)$,  
$(b)$ follows from 
$\nu^{\pi}(s) = \tsum_{s' \in \cS} \nu^{\pi}(s') d_{s'}^{\pi} (s)$, and $(c)$ follows from \eqref{state_sampling_condition}.
Second, 
given the definition  of $\rho_k^\dagger$ and $\cS_k$, 
we have 
\begin{align*}
(T_2)
= \tsum_{s' \in \cS_k} \rho_k(s') \tsum_{ s \in \cS_k \setminus \cbr{s'} } \tfrac{\nu^*(s)}{\rho_k(s)} D^{\pi^*}_{\pi_k}(s) 
& = \tsum_{s \in \cS_k} \tsum_{s' \in \cS_k \setminus \cbr{s}}  \rho_k(s') \tfrac{\nu^*(s)}{\rho_k(s)} D^{\pi^*}_{\pi_k}(s) \\
& = \tsum_{s \in \cS_k} (1- \rho_k(s)) \tfrac{\nu^*(s)}{\rho_k(s)} D^{\pi^*}_{\pi_k}(s) \\
& \leq  (1 - \rho_k^\dagger) \tsum_{s \in \cS_k}  \tfrac{\nu^*(s)}{\rho_k(s)} D^{\pi^*}_{\pi_k}(s) .
\end{align*}
Combining the above two observations with \eqref{scBPMD:value_at_update_4}, 
and using the definition of $\rho_k^\dagger$ again, 
we arrive at 
\begin{align*}
& \eta_k  (1-\gamma) \sbr{f(\pi_k) - f(\pi^*) } + 
  \tfrac{\eta_k }{\rho^\dagger_k}  \EE_{s_k} \sbr{f(\pi_{k+1}) - f(\pi_k)} \nonumber 
\\  
\leq &
  \tsum_{s \in \cS_k}  \tfrac{\nu^*(s)}{\rho_k(s)}  D^{\pi^*}_{\pi_k}(s)  - (1+ \eta_k \mu)    \EE_{s_k}  \big[ \tsum_{s\in \cS_k}  \tfrac{\nu^*(s)}{\rho_k(s)} D^{\pi^*}_{\pi_{k+1}}(s)\big]
    + \eta_k \mu   (1 - \rho_k^\dagger) \tsum_{s \in \cS_k}  \tfrac{\nu^*(s)}{\rho_k(s)} D^{\pi^*}_{\pi_k}(s)
   . 
\end{align*}
 The desired result \eqref{eq:bpmd_generic_recursion} follows immediately after simple rearrangement.
\end{proof}

Note that as stated in \eqref{state_sampling_condition}, Lemma \ref{lemma:bpmd_generic_recursion} crucially relies on the condition of the sampling distribution $\rho_k$ to have a support as least as large as that of $\nu^*$, the stationary state distribution induced by the optimal policy. 
We will term such a type of sampling distributions as {\it exploratory distributions}.

\begin{definition}[Exploratory Distribution]\label{def_explore_sampling}
A distribution $\rho \in \Delta_{\cS}$ is called an exploratory distribution if $\rho^\dagger \coloneqq \min \cbr{\rho(s): \nu^*(s) > 0 } > 0$, and we say $\rho^\dagger$ is its minimal effective state frequency.
\end{definition}

Notably, the class of exploratory distributions trivially includes the uniform distribution. 
More importantly, we will construct a MDP instance in Section \ref{subsec_sampling_comparison}, and demonstrate that exploratory sampling distributions are indeed necessary for successful policy optimization (see Remark \ref{remark_motivate_explore_sampling}).
We will also discuss the approximate convergence of BPMD when this condition holds partially in Section \ref{sec_effect_sampling}.


In the following sections, we proceed to 
establish concrete conditions on the stepsizes $\cbr{\eta_k}$ and sampling distributions $\cbr{\rho_k}$, and
develop non-asymptotic convergence results of BPMD for both strongly-convex and non-strongly-convex regularizers.
We will pay special attention to the class of static sampling distributions, for which simple stepsize rules can be derived. 

\begin{definition}[Static Sampling Distribution]
The sampling distributions 
$\cbr{\rho_t}$ are called static if $\rho_t = \rho$ for all $t \geq 0$ for some fixed distribution $\rho$.
\end{definition}

\subsection{Convergence with Strongly-convex Regularizers}\label{subsec_sc}
For strongly-convex regularizers, we first establish that with proper conditions on the choice of stepsizes and sampling distribution, 
BPMD achieves global linear convergence.

\begin{theorem}[$\mu>0$, linear convergence]\label{thrm:bpmd_sc}
Suppose $h$ satisfies $\mu > 0$, the sampling distributions $\cbr{\rho_k}$ satisfy $\cS_{k+1} = \cS_k$,  and the stepsizes $\cbr{\eta_k}$ satisfy
\begin{align}\label{sc_step_sampling_condition_general}
\tfrac{1}{\eta_{k-1}} + \mu \geq 
\rbr{\tfrac{1}{\eta_k} + \mu (1- \rho_k^\dagger)} \tfrac{\rho^\dagger_k}{\rho^\dagger_{k-1}} \norm{\tfrac{\rho_{k-1}}{\rho_k}}_\infty \sbr{ 1 - (1-\gamma) \rho_k^\dagger}^{-1},
\end{align}
then BPMD satisfies 
\begin{align}
& \EE\sbr{f(\pi_{k}) - f(\pi^*) \label{eq_sc_linear_general} 
+ 
  \rho_{k-1}^\dagger \rbr{\tfrac{1}{\eta_{k-1}} + \mu }      \tsum_{s \in \cS_{k}}  \tfrac{\nu^*(s)}{\rho_{k-1}(s)} D^{\pi^*}_{\pi_{k}}(s)}
\\
\leq &   \rbr{1 - (1-\gamma) \vartheta_k}^k \sbr{f(\pi_0) - f(\pi^*) 
+ \rbr{\tfrac{1}{\eta_0} + \mu (1-\rho^\dagger_0) }  \log \abs{\cA} }, ~\forall k \geq 1, \nonumber
\end{align}
 where $\vartheta_k = \min_{0 \leq t \leq k-1} \rho_t^\dagger$ denotes the minimal effective state frequency visited by the sampling distributions.
\end{theorem}
\begin{proof}
Given the choices of $\cbr{\eta_k}$ and $\cbr{\rho_k}$ satisfying \eqref{sc_step_sampling_condition_general} and $\cS_{k+1} = \cS_k$, 
we can recursively apply \eqref{eq:bpmd_generic_recursion} and obtain
\begin{align*}
& \EE\sbr{f(\pi_{k}) - f(\pi^*) + 
  \rho_{k-1}^\dagger \rbr{\tfrac{1}{\eta_{k-1}} + \mu }      \tsum_{s \in \cS_{k-1}}  \tfrac{\nu^*(s)}{\rho_{k-1}(s)} D^{\pi^*}_{\pi_{k}}(s)}  \\
  \leq &
\tprod_{t=0}^{k-1} \rbr{1 - (1-\gamma) \rho^\dagger_t} \sbr{f(\pi_0) - f(\pi^*) 
+ \rho_0^\dagger \rbr{\tfrac{1}{\eta_0} + \mu (1-\rho^\dagger_0) }    \tsum_{s \in \cS_0}  \tfrac{\nu^*(s)}{\rho_0(s)}  D^{\pi^*}_{\pi_0}(s) } \\
 \overset{(a)}{\leq} &
\tprod_{t=0}^{k-1} \rbr{1 - (1-\gamma) \rho^\dagger_t} \sbr{f(\pi_0) - f(\pi^*) 
+ \rbr{\tfrac{1}{\eta_0} + \mu (1-\rho^\dagger_0) }  \log \abs{\cA} } \\
 \overset{(b)}{\leq} &
 \rbr{1 - (1-\gamma) \vartheta_k}^k \sbr{f(\pi_0) - f(\pi^*) 
+ \rbr{\tfrac{1}{\eta_0} + \mu (1-\rho^\dagger_0) }  \log \abs{\cA} },
\end{align*}
where $(a)$ uses $\rho_0^\dagger \leq \rho_0(s)$, and the fact that
$\phi(\pi^*, \pi_0) \leq \log \abs{\cA}$ since
$\pi_0$ is the uniform policy;
$(b)$ uses the definition of $\vartheta_k$.
\end{proof}

In view of \eqref{eq_sc_linear_general}, Theorem \ref{thrm:bpmd_sc} states that BPMD attains linear convergence for strongly-convex regularizers provided the minimal effective state frequency $\vartheta_k$ is independent of the total iterations. 
This condition appears to be fairly general as it can be readily satisfied by any exploratory and static sampling distribution.

Moreover, although \eqref{sc_step_sampling_condition_general} appears to be abstract, we make two remarks worthy of attention.
First, \eqref{sc_step_sampling_condition_general} can be pre-verified before execution of BPMD, since $\cbr{\rho_k}$ do not depend on the generated polices $\cbr{\pi_k}$.
We will turn our attention to the case where $\cbr{\rho_k}$ indeed depend on the generated policies $\cbr{\pi_k}$ in Section \ref{subsec_sampling_comparison}.
Second, condition \eqref{sc_step_sampling_condition_general} becomes easier to satisfy when $\cbr{\rho_k}$ become closer to static sampling distributions.  
 Indeed, when $\cbr{\rho_k}$ are static sampling distributions, we show in the following corollary that constant stepsizes suffice to yield linear convergence.


\begin{corollary}[$\mu > 0$, static sampling]\label{cor_sc_linear_static}
Suppose $h$ satisfies $\mu > 0$.
Let $\cbr{\rho_k}$ be static and exploratory sampling distributions. That is, $\rho_k = \rho$ for all $k \geq  0$ and $\rho$ is exploratory.
 Let $\eta_t = \eta$ for all $t \geq 0$, where $\eta > 0$ satisfies $
1 + \eta \mu \geq \tfrac{1}{\gamma},
$
then for all $k \geq 1$, BPMD satisfies 
\begin{align*}
& \EE\sbr{f(\pi_{k}) - f(\pi^*) + \rho^\dagger (\tfrac{1}{\eta} + \mu) \phi(\pi^*, \pi_k) }   
\leq   \rbr{1 - (1-\gamma) \rho^\dagger}^k \sbr{f(\pi_0) - f(\pi^*) 
+ \rbr{\tfrac{1}{\eta} + \mu (1- \rho^\dagger) }  \log \abs{\cA} }.
\end{align*}
\end{corollary}

\begin{proof}
The claim follows directly from verifying condition \eqref{sc_step_sampling_condition_general} holds if $\eta_k = \eta$ and $1 + \eta \mu \geq 1/\gamma$, 
and the simplification of \eqref{eq_sc_linear_general} with a constant stepsize.
\end{proof}

It should be noted that  the choice of stepsize for static sampling distributions does not depend on the concrete choice of the sampling choice itself. 
Moreover,  the established convergence for static sampling distributions is stronger than the one obtained for time-dependent sampling,  in the sense that we can certify  linear convergence for both the optimality gap and the distance to the (unique) optimal policy.

\subsection{Convergence with Non-strongly-convex Regularizers}\label{subsec_non_sc}
We next discuss the convergence of BPMD when the regularizer lacks strong convexity.
Specifically, we will propose two stepsize schemes and establish their sublinear, and linear convergence, respectively. 
The first scheme adopts a mildly increasing stepsize rule, and reduces to the constant stepsize for static sampling distributions,
while the second one will adopt a much more aggressive, in fact exponentially increasing stepsize rule.

\subsubsection{Sublinear Convergence}
We first consider the case where sublinear convergence is attained by adopting a mildly increasing stepsize rule.

\begin{theorem}[$\mu=0$, sublinear convergence]\label{thrm:bpmd_nsc}
Suppose $h$ satisfies $\mu = 0$. Suppose  the sampling distribution $\cbr{\rho_k}$ satisfies $\cS_{k+1} = \cS_k$,    and the stepsizes $\cbr{\eta_k}$ satisfy 
\begin{align}\label{step_nonsc_sublinear}
\eta_{k+1} \geq \eta_k \norm{ \tfrac{\rho_k}{\rho_{k+1}}}_\infty \tfrac{\rho_{k+1}^\dagger}{\rho_k^\dagger}.
\end{align}
Then BPMD satisfies
\begin{align*}
 \EE \sbr{  f(\pi_k) - f(\pi^*)}
 \leq  \tfrac{ \log \abs{\cA} /\eta_0 +  f(\pi_0) - f(\pi^*)}{k (1-\gamma) \vartheta_k }, ~\forall k \geq 1,
 \end{align*}
 where $\vartheta_k = \min_{1 \leq t \leq k} \rho_t^\dagger$ denotes the minimal effective state frequency visited by the sampling distributions.
\end{theorem}

\begin{proof}
By taking $\mu = 0$ in  \eqref{eq:bpmd_generic_recursion}, we have  
\begin{align}
\EE_{s_k} \sbr{f(\pi_{k+1}) - f(\pi^*)}   
 \leq &
\rbr{1 - (1-\gamma) \rho^\dagger_k} \sbr{f(\pi_k) - f(\pi^*) }
+ \tfrac{\rho_k^\dagger}{\eta_k}   \tsum_{s \in \cS_k}  \tfrac{\nu^*(s)}{\rho_k(s)}  D^{\pi^*}_{\pi_k}(s) \nonumber  \\
&  -  \tfrac{\rho_k^\dagger}{\eta_k}     \EE_{s_k}  \big[ \tsum_{s\in \cS_k}  \tfrac{\nu^*(s)}{\rho_k(s)} D^{\pi^*}_{\pi_{k+1}}(s)\big]\label{BPMD:recursion}.
\end{align}

We now sum up \eqref{BPMD:recursion} from $t=0$ to $k-1$.
Further taking total expectation, 
we obtain 
\begin{align*}
\tsum_{t=1}^k   (1-\gamma)   \rho_t^\dagger  \EE \sbr{ f(\pi_t) - f(\pi^*)}
& \overset{(a)}{=} \tsum_{t=1}^k   (1-\gamma) \EE \sbr{  \rho_t^\dagger  \rbr{f(\pi_t) - f(\pi^*)}}   \\
& 
\overset{(b)}{\leq} \tfrac{\rho^\dagger_0}{\eta_0} \tsum_{s \in \cS_0} \tfrac{\nu^*(s)}{\rho_0(s)} D^{\pi^*}_{\pi_0}(s)  
+ f(\pi_0) - f(\pi^*) \\
& \overset{(c)}{\leq} 
\tfrac{1}{\eta_0} \phi(\pi^*, \pi_0) +  f(\pi_0) - f(\pi^*) ,
\end{align*}
where $(a)$ follows from the fact that $\cbr{\rho_t}$ does not depends on $\cbr{\pi_t}$,
 $(b)$ follows from the choice of stepsizes in \eqref{step_nonsc_sublinear}, $\cS_{k+1} = \cS_k$, and a telescoping argument, 
and  $(c)$ uses the definition of $\rho_0^\dagger$, and the definition of $\phi(\pi^*, \pi)$.
In addition, given \eqref{ineq:value_decrease}, \eqref{ineq:progress_lb}, and the definition of $f$ in \eqref{eq:mdp_single_obj}, we also know that 
$f(\pi_{t+1}) \leq f(\pi_t)$. 
Hence
\begin{align*}
k (1-\gamma)  \min_{1 \leq t \leq k} \rho_t^\dagger  \EE \sbr{  f(\pi_k) - f(\pi^*)}
\leq \tfrac{1}{\eta_0} \phi(\pi^*, \pi_0) +  f(\pi_0) - f(\pi^*).
\end{align*}
The desired claim follows immediately after rearrangement and the fact $\phi(\pi^*, \pi_0) \leq \log \abs{\cA}$.
\end{proof}

Note that \eqref{step_nonsc_sublinear} can be pre-verified before the execution of the BPMD method. 
As an even simpler alternative to this pre-verification, we next show that
similar to Corollary \ref{cor_sc_linear_static}, the stepsize scheme \eqref{step_nonsc_sublinear} simplifies significantly when we consider static sampling distributions, 
which is also agnostic to the concrete choice of static sampling distribution.

\begin{corollary}[$\mu = 0$, static sampling]\label{cor_non_sc_sublinear}
Suppose $h$ satisfies $\mu = 0$.
Let $\cbr{\rho_k}$ be static and exploratory sampling distributions. That is, $\rho_k = \rho$ for all $k \geq  0$ and $\rho$ is exploratory.
Then for any constant stepsize $\eta_t = \eta > 0$, BPMD satisfies
\begin{align*}
 \EE \sbr{  f(\pi_k) - f(\pi^*)}
 \leq  \tfrac{ \log \abs{\cA} /\eta +  f(\pi_0) - f(\pi^*)}{k (1-\gamma) \rho^\dagger }, ~\forall k \geq 1.
 \end{align*}
\end{corollary}

\begin{proof}
The claim follows immediately from observing that condition \eqref{step_nonsc_sublinear} holds with any $\eta_t = \eta > 0$ and if $\cbr{\rho_t}$ is static.
\end{proof}

\vspace{-0.05in}
\subsubsection{Linear Convergence}
We proceed to show that with an exponentially increasing stepsize rule, BPMD attains global linear convergence.
Though linear convergence with exponentially increasing stepsizes has been extensively studied for batch PG methods (see \cite{lan2021policy, khodadadian2021linear, xiao2022convergence, li2022homotopic}), the analysis presented here is considerably simpler, by directly exploiting the general convergence properties established in Lemma \ref{lemma:bpmd_generic_recursion}.

\begin{theorem}[$\mu=0$, linear convergence]\label{thrm:bpmd_non_sc}
Suppose $h$ satisfies $\mu = 0$, the  sampling distributions $\cbr{\rho_k}$ satisfy $\cS_{k+1} = \cS_k$,  and stepsizes $\cbr{\eta_k}$ satisfy
\begin{align}\label{non_sc_step_sampling_condition_general}
\eta_k  \geq 
\eta_{k-1} \cdot \tfrac{\rho^\dagger_k}{\rho^\dagger_{k-1}} \norm{\tfrac{\rho_{k-1}}{\rho_k}}_\infty \sbr{ 1 - (1-\gamma) \rho_k^\dagger}^{-1},
\end{align}
then BPMD satisfies 
\begin{align*}
 \EE\sbr{f(\pi_{k}) - f(\pi^*) 
+ 
  \tfrac{\rho_{k-1}^\dagger}{\eta_{k-1}}   \phi(\pi^*, \pi_k) }
\leq    \rbr{1 - (1-\gamma) \vartheta_k}^k \sbr{f(\pi_0) - f(\pi^*) 
+ \tfrac{\log \abs{\cA}}{\eta_0} }, ~\forall k \geq 1,
\end{align*}
 where $\vartheta_k = \min_{0 \leq t \leq k-1} \rho_t^\dagger$ denotes the minimal effective state frequency visited by the sampling distributions.
\end{theorem}

\begin{proof}
The claim follows by taking $\mu = 0$ in the proof of Theorem \ref{thrm:bpmd_sc}, and follows the same lines therein. 
\end{proof}

It is immediate to observe that the stepsize scheme \eqref{non_sc_step_sampling_condition_general} reduces to a simple exponential scaling rule when considering static sampling distributions. 
Note that unlike prior scenarios in Corollary \ref{cor_sc_linear_static} and \ref{cor_non_sc_sublinear}, the stepsize scheme here indeed will depend on the concrete sampling distribution via its minimal effective state frequency.

\begin{corollary}[$\mu=0$, static sampling]\label{cor_non_sc_static_linear}
Suppose $h$ satisfies $\mu = 0$.
Let $\cbr{\rho_k}$ be static and exploratory sampling distributions. That is, $\rho_k = \rho$ for all $k \geq  0$ and $\rho$ is exploratory.
Let stepsizes $\cbr{\eta_k}$ staisfy
\begin{align*}
\eta_k  \geq 
\eta_{k-1} \cdot  \sbr{ 1 - (1-\gamma) \rho^\dagger}^{-1}.
\end{align*}
Then BPMD satisfies 
\begin{align}
 \EE\sbr{f(\pi_{k}) - f(\pi^*) 
+ 
  \tfrac{\rho^\dagger}{\eta_{k-1}}   \phi(\pi^*, \pi_k) } 
&   \leq    \rbr{1 - (1-\gamma) \rho^\dagger}^k \sbr{f(\pi_0) - f(\pi^*) 
+ \tfrac{\phi(\pi^*, \pi_0)}{\eta_0} } \label{eq_nonsc_linear_static_use_for_unif}
  \\
& \leq    \rbr{1 - (1-\gamma) \rho^\dagger}^k \sbr{f(\pi_0) - f(\pi^*) 
+ \tfrac{\log \abs{\cA}}{\eta_0} }
, ~ \forall k \geq 1.  \nonumber
\end{align}
\end{corollary}

In view of Corollary \ref{cor_sc_linear_static} and \ref{cor_non_sc_static_linear}, BPMD converges linearly in optimality gap for both strongly-convex regularizers and non-strongly-convex regularizers, with their differences being two-fold: (1) strongly-convex regularizers only require constant stepsizes to ensure linear convergence, while non-strong-convex regularizers require exponentially increasing stepsizes; (2) with strongly-convex regularizers, one can also ensure linear convergence of the distance to the (unique) optimal policy, while one can not make similar claims for the non-strongly-convex regularizers due to the vanishing coefficient of $\phi(\pi^*, \pi_k)$ in \eqref{eq_nonsc_linear_static_use_for_unif}.

In addition, by comparing Corollary \ref{cor_non_sc_sublinear} and \ref{cor_non_sc_static_linear}, it is clear that for non-strongly-convex regularizers and static sampling distributions, the exponential scaling stepsize rule should be always preferred over the constant stepsize rule.
We summarize the convergence results in Section \ref{subsec_sc}, \ref{subsec_non_sc} in Table \ref{tab_iteration_comp_compare_static}.

\begin{table}[tb!]
\centering
\begin{tabular}{ ||c|c|c|| } 
\hline
Regularizers & Stepsize Scheme & Iteration Complexity   \\
\hhline{||=|=|=||}
\multirow{1}{*}{$\mu > 0$} & $\eta_k = \eta$ & $\tfrac{1}{(1-\gamma)\rho^\dagger} \log (\tfrac{1}{\epsilon})$   \\
\hline
\multirow{2}{*}{$\mu = 0$} & $\eta_k = \eta$ & $\tfrac{1}{(1-\gamma)\rho^\dagger} \cdot \tfrac{1}{\epsilon}$   \\ 
\cline{2-3}
& $\eta_k  \geq 
\eta_{k-1} \cdot  [ 1 - (1-\gamma) \rho^\dagger]^{-1}$  & $\tfrac{1}{(1-\gamma)\rho^\dagger} \log (\tfrac{1}{\epsilon})$   \\ 
\hline
\end{tabular}
\caption{Iteration complexity of BPMD with a static sampling distribution $\rho$,  for different stepsize schemes and strong convexity modulus of regularizers. }
\label{tab_iteration_comp_compare_static}
\vspace{-0.2in}
\end{table}

\subsection{Uniform Sampling versus On-policy Sampling}\label{subsec_sampling_comparison}
Having focused on general sampling distributions in our prior discussions, we now turn our attention to two seemingly natural sampling schemes, namely the uniform and on-policy sampling, and discuss their comparative advantage in computational efficiency and implementation practicality. 

\subsubsection{Uniform Sampling Scheme}\label{subsubsec_uniform_sampling_bpmd}
For the uniform sampling scheme, the sampling distributions $\cbr{\rho_k}$ are given as $\rho_k =  \mathrm{Unif}(\cS)$ for all $k \geq 0$. 
Since the uniform distribution is exploratory, we can specialize all previously obtained results of BPMD for this sampling scheme. 

\begin{corollary}[Uniform Sampling Scheme]\label{thrm_uniform_deterministic}
Let $\rho_k = \mathrm{Unif}(\cS)$ for all $k \geq 0$ in the BPMD method.
Then the convergence of BPMD can be characterized as follows.

 \underline{I}. Strongly-convex regularizers: Suppose $h$ satisfies $\mu > 0$.
Let  $\eta_t = \eta$ for all $t \geq 0$, with
$
1 + \eta \mu \geq \tfrac{1}{\gamma},
$
then BPMD satisfies 
\begin{align*}
& \EE\sbr{f(\pi_{k}) - f(\pi^*) +  \tfrac{1 + \eta \mu }{\eta \abs{\cS}}  \phi(\pi^*, \pi_k) }   
\leq   \rbr{1 - \tfrac{1-\gamma}{\abs{\cS}}}^k \sbr{f(\pi_0) - f(\pi^*) 
+ \rbr{\tfrac{1}{\eta} + \mu  }  \log \abs{\cA} }, ~ \forall k \geq 1.
\end{align*}

 \underline{II}. Non-strongly-convex regularizers: Suppose $h$ satisfies $\mu = 0$.
Let stepsizes $\cbr{\eta_k}$ staisfy
$
\eta_k  \geq 
\eta_{k-1} \cdot  \rbr{ 1 - \rbr{1-\gamma}/ \abs{\cS} }^{-1}.
$
Then BPMD satisfies 
\begin{align}
 \EE\sbr{f(\pi_{k}) - f(\pi^*) 
 }
 \leq
 \rbr{1 - \tfrac{1-\gamma}{\abs{\cS}}}^k \sbr{f(\pi_0) - f(\pi^*) 
+ \tfrac{\log \abs{\cA}}{\eta_0} }
, ~ \forall k \geq 1. \nonumber 
\end{align}
\end{corollary}

\begin{proof}
The desired claims follow immediately by applying Corollary \ref{cor_sc_linear_static} and \ref{cor_non_sc_static_linear} with $\rho = \mathrm{Unif}(\cS)$.
\end{proof}

To better understand the implication of Corollary \ref{thrm_uniform_deterministic}, we will focus on the iteration complexity, per-iteration computational cost, and total computational cost of BPMD with the uniform sampling scheme, for finding an $\epsilon$-optimal policy.
We will compare these key metrics of BPMD to that of the best batch PG methods, studied in \cite{lan2021policy} (APMD), which is later simplified in   \cite{li2022homotopic} (HPMD) and concurrently in \cite{xiao2022convergence} (exponential-stepsize PMD), and illustrate the practical benefits of BPMD over these batch PG methods.
We first focus on establishing the per-iteration computation of BPMD and aforementioned batch PG methods.




\underline{\it Batch PG Methods.} For policy evaluation, we recall the following Bellman equations,
\begin{align}\label{complexity_q_v_bellman}
Q^{\pi}(s,a) = r(s,a) + \gamma \tsum_{s'} \cP(s'|s,a) V^{\pi}(s') , ~\forall (s,a), ~ \text{and} ~ 
V^\pi = r^\pi + \gamma \PP^\pi V^\pi,
\end{align}
where $r^\pi (s) = \tsum_{a} r(s,a) \pi(a|s)$, and $\PP^\pi (s, s') = \tsum_{a} \cP(s'|s,a) \pi(a|s)$. The steps to obtain $Q^{\pi}$ and its associated computational cost are:
\begin{itemize}[noitemsep,topsep=0pt]
\item Step 1 - Forming $r^\pi$: $\cO(\abs{\cS} \abs{\cA})$. 
\item Step 2 - Forming $\PP^\pi$: $\cO(\abs{\cS}^2 \abs{\cA})$.
\item Step 3 - Computing $V^\pi$ by solving $V^{\pi} = (I - \gamma \PP^\pi)^{-1} r^\pi$ with Gauss-Jordan elimination: $\cO(\abs{\cS}^3)$. 
\item Step 4 - Forming $Q^\pi$ using \eqref{complexity_q_v_bellman} and the obtained $V^\pi$: $\cO(\abs{\cS}^2 \abs{\cA})$. 
\end{itemize}

For policy update, recall that batch PG method simply performs update \eqref{eq:bpmd_update} for every state instead of the sampled state, the computation is thus $\cO(\abs{\cS} \abs{\cA})$. 
Hence the per-iteration computation for batch PG method is given by  $\cO( \abs{\cS}^3 + \abs{\cS}^2 \abs{\cA} )$.

\underline{\it BPMD.} At any given iteration, let $\pi^{-}$ be the policy before the last policy update, and let  $r^{\pi^-}$, $\PP^{\pi^{-}}$ be the quantities defined as before,
and suppose we store $(I-\gamma \PP^{\pi^-})^{-1}$. 
Let $s^-$ denote the sampled state of the last iteration,  $s$ denote the sampled state to be updated, and $\pi$ denote the current policy (to be updated).
By observing that $\pi$ and $\pi^{-}$ differ only in one state (i.e., state $s^-$), we have the following.
\begin{itemize}[noitemsep,topsep=0pt]
\item Step 1 - Forming $r^\pi$ by computing $\Delta_r = r^\pi - r^{\pi^-}$, which has only one non-zero entry corresponding to $s^-$. This takes $\cO(\abs{\cA})$ computation. 
\item Step 2 - Computing $\Delta_{s^-}^\top$. This is the only non-zero row corresponding to state $s^{-}$ in matrix  $\Delta = \PP^{\pi} - \PP^{\pi^-}$. This takes $\cO(\abs{\cS} \abs{\cA})$ computation.  
\item Step 3 - Computing $V^\pi$ by solving $V^{\pi} = (I - \gamma \PP^\pi)^{-1} r^\pi$.
Equivalently, by denoting $e_{s^-} \in \RR^{\abs{\cS}}$ as the one-hot vector with non-zero entry corresponding to $s^-$, 
and $\Delta_{s^-}^\top$ denoting the $s^{-}$-row of $\Delta$, 
then 
\begin{align*}
&   \rbr{I - \gamma \PP^\pi}^{-1}
 = \rbr{ I -\gamma \PP^{\pi^-} - \gamma e_{s^-} \Delta_{s^-}^\top}^{-1} 
\\
   \overset{(a)}{=} &
\rbr{ I - \gamma \PP^{\pi^-}}^{-1} + 
\gamma \rbr{ I - \gamma \PP^{\pi^-}}^{-1} e_{s^-}  \Delta_{s^-}^\top \rbr{ I - \gamma \PP^{\pi^-}}^{-1}
/ \sbr{
1 - \gamma \Delta_{s^-}^\top  \rbr{ I - \gamma \PP^{\pi^-}}^{-1} e_{s^-} 
}
,
\end{align*}
where equality $(a)$ follows from applying Woodbury matrix identity.
Thus forming $\rbr{I - \gamma \PP^\pi}^{-1}$ requires $\cO(\abs{\cS}^2)$ computation, and hence forming $V^\pi$ costs $\cO(\abs{\cS}^2)$ computation.
\item Step 4 - Forming $Q^\pi(s, \cdot)$ using \eqref{complexity_q_v_bellman} and the obtained $V^\pi$, takes $\cO(\abs{\cS} \abs{\cA})$ computation. 
\end{itemize}

Hence each policy evaluation step in BPMD takes $\cO(\abs{\cS}^2 + \abs{\cS} \abs{\cA})$ computation. Combined with the $\cO(\abs{\cA})$ computation for policy update \eqref{eq:bpmd_update}, the per-iteration computation of BPMD is $\cO(\abs{\cS}^2 + \abs{\cS} \abs{\cA})$.

We now turn to the total iteration complexity. For any $\epsilon > 0$, 
the iteration complexity for finding an $\epsilon$-optimal policy by BPMD is readily implied by Corollary \ref{thrm_uniform_deterministic}.
The iteration complexity of exponential-stepsize PMD/HPMD/APMD is  $\cO(\log(1/\epsilon)/(1-\gamma))$ \cite{lan2021policy, li2022homotopic, xiao2022convergence}. 

Summarizing our above discussions, for any $\epsilon > 0$, the total iterations and computational complexity required by BPMD and the best batch PG methods to find an $\epsilon$-optimal policy are given by Table \ref{tab_comparison_uniform_bpmd_batch}.

\begin{table}[htb!]
\centering
\begin{tabular}{ ||c|c|c|c|| } 
\hline
Method & Iter. Complexity & Per-iter. Computation & Total Computation \\
\hhline{||=|=|=|=||}
\multirow{1}{*}{BPMD} & $\tfrac{\abs{\cS}}{1-\gamma} \log \rbr{\tfrac{1}{\epsilon}}$ & $\abs{\cS}^2 + \abs{\cS} \abs{\cA} $ &  $\tfrac{\abs{\cS}^2  ( \abs{\cS} + \abs{\cA}) }{1-\gamma} \log \rbr{\tfrac{1}{\epsilon}}$  \\
\hline
\multirow{1}{*}{Batch PG Methods \cite{lan2021policy, xiao2022convergence, li2022homotopic}} & $\tfrac{1}{1-\gamma} \log \rbr{\tfrac{1}{\epsilon}}$ & $\abs{\cS}^3 + \abs{\cS}^2 \abs{\cA} $  & $\tfrac{\abs{\cS}^2 (\abs{\cS} + \abs{\cA}) }{1-\gamma} \log \rbr{\tfrac{1}{\epsilon}}$   \\ 
\hline
\end{tabular}
\caption{{Comparison on the  per-iteration and total computational cost of BPMD and best batch PG methods for finding an $\epsilon$-optimal policy.
BPMD and batch PG methods  have the same total computational cost. }}
\label{tab_comparison_uniform_bpmd_batch}
\end{table}



\begin{remark}[{\bf Benefits of Uniform BPMD over Batch PG}]\label{remark_benefits_uniform}
The bottom line of Table \ref{tab_comparison_uniform_bpmd_batch} is that 
BPMD with the uniform sampling shares the same total computation as that of the batch PG methods proposed in prior literature, but does so with a much cheaper per-iteration cost.
This brings at least two practical benefits of BPMD worthy of one's attention when considering solving large state-space MDPs.
\vspace{0.05in}
\begin{enumerate}[noitemsep,topsep=0pt]
\item[{\bf (A)}] The policy optimization process has a finer granularity, with each cheap iteration bringing steady improvement of the current policy. The mild changes of policies throughout this process can also be beneficial for risk-sensitive applications (e.g., medical treatment planning \cite{steimle2017markov}) where drastic changes in decision making should be preferably avoided;
\item[{\bf (B)}] The fine granularity of BPMD also provides more information when monitoring the optimization process, and more importantly, allows early stopping whenever necessary (e.g., when the performance is satisfactory enough), without worrying about loss of progress.
In contrast, batch PG methods have to wait for a complete cycle of policy evaluation and policy improvement over all states. 
Waiting for the completion of each cycle is time consuming for large-state MDPs, and applying early stopping during this cycle would completely waste the computation spent within this cycle. 
\end{enumerate}
\end{remark}

\vspace{0.05in}
\begin{remark}[{\bf Extensions to multi-state update}]
All our previously discussion on BPMD was based on the prior that only one state is sampled and updated at each iteration.  
This is rather a simplification, instead of a restriction of BPMD, and we now discuss the feasibility of a multi-state BPMD variant, that extends BPMD to allow updating multiple states at each iteration.
With a uniform sampling scheme, the multi-state BPMD variant samples a block of states  $\cB$ at each iteration, which contains $\abs{\cB}$ distinct states sampled uniformly from $\cS$ without replacement.
The multi-state BPMD then performs policy update for every state within the block $\cB$.
Following similar analysis, one can readily obtain,
\begin{align*}
\EE \sbr{ f(\pi_{k}) - f(\pi^*) } = \cO \rbr{
\rbr{1- \tfrac{\abs{\cB} (1-\gamma)}{\abs{\cS}}}^k 
\sbr{ \rbr{ f(\pi_0) - f(\pi^*) } +  \tfrac{\log \abs{\cA}}{\eta_0}}} , ~\forall k \geq 1,
\end{align*}
for strongly-convex regularizers with constant stepsizes and 
for non-strongly-convex regularizers with exponentially increasing stepsizes.
Clearly, when $\cB = \cS$, then this multi-state BPMD variant reduces to the batch PG method PMD \cite{lan2021policy} with a matching convergence rate.
The flexibility of choosing  $\abs{\cB}$ allows us to strike a balance between the convergence rate and the per-iteration computation in a problem-dependent~way.
\end{remark}

\subsubsection{On-policy Sampling}
We now turn our attention to the on-policy sampling scheme.
In contrast to the sampling schemes considered before, the on-policy sampling scheme does not require  pre-specified $\cbr{\rho_k}$ as the input to BPMD.
Instead, at each iteration, $\rho_k$ is the steady state distribution, generated by following the latest policy $\pi_k$ starting from the last updated state $s_{k-1}$, and we let the state before sampling for the first policy update be denoted as $s_{-1}$.

Clearly, on-policy sampling offers a unique practical benefit over uniform sampling, as it can be easily implemented by simply following the current policy, without doing any explicit exploration over the state space.
On the other hand, it should be clear that $\rho_k$ belongs to the $\sigma$-algebra  generated by the past sampled states and hence is also random. 
Our ensuing global convergence guarantees for BPMD with on-policy sampling requires the following condition on this sequence of random measures $\cbr{\rho_k}$.

\begin{assumption}\label{assump_exploratory_on_policy}
The generated sampling distributions $\cbr{\rho_k}$ in BPMD satisfy $\rho_k^\dagger \geq \varrho$ for all $k \geq 0$ with probability 1, for some $\varrho > 0$.
\end{assumption}

 Assumption \ref{assump_exploratory_on_policy} can be satisfied, for instance, 
 when the transition kernel is stochastic enough so that every state can transit to any other state regardless of the action.
 For general MDPs, this assumption appears to be strong given its algorithmic-dependent nature.
  Yet it turns out we can construct a simple MDP instance,
 which demonstrates the necessity of this condition for successful policy optimization with on-policy sampling. 

\begin{proposition}[Hard MDP Instance for On-policy Sampling]\label{prop_hard_example_on_policy}
There exists an MDP instance, where  Assumption \ref{assump_exploratory_on_policy} fails, and more importantly BPMD with on-policy sampling also fails to find the optimal policy.
\end{proposition}

\begin{figure}[htb!]
\centering 
    \includegraphics[width=0.45\textwidth]{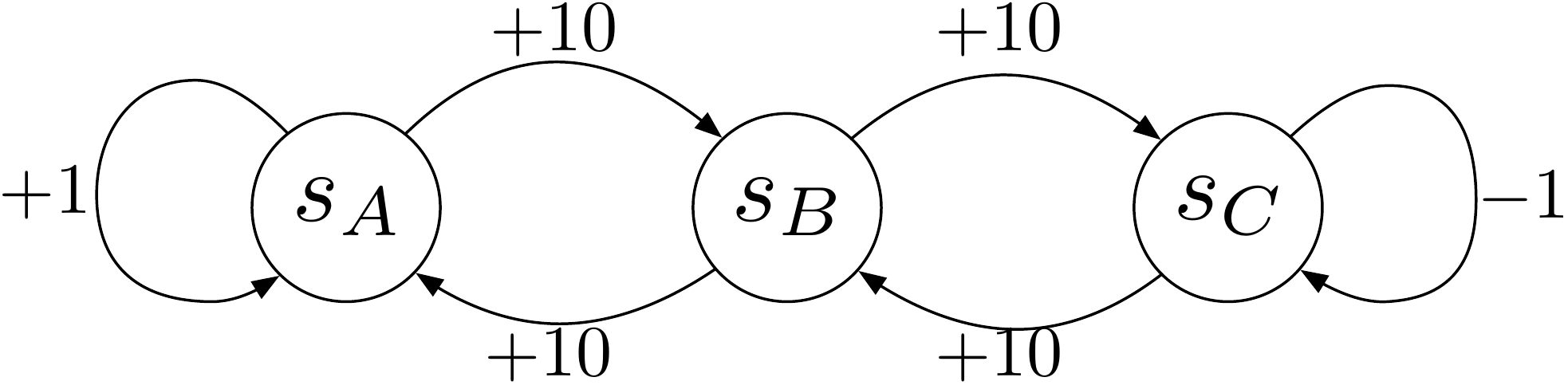}
  \caption{ 
 An MDP instance for which on-policy sampling fails to find the optimal policy. Each arc $(s,a, s')$ specifies the transition of starting from state $s$, making action $a$ and then transits to state $s'$. The value of the arc denotes the corresponding cost $c(s,a)$.
  }
  \label{fig_on_policy}
\end{figure}

\begin{proof}

Consider the following MDP with three states, with every state associated with two actions $\cbr{L, R}$, denoting going left and right. 
The transition diagram and the associated costs are given in Figure \ref{fig_on_policy}.
Suppose the initial policy is given as 
\begin{align*}
\pi_0(L | s_A) = 1, ~ \pi_0(R| s_A) = 0; ~ \pi_0(\cdot | s) = \mathrm{Unif}(\cA), ~ \forall s \neq s_A.
\end{align*}
Since KL-divergence is not suited for policies assigning zero probability to actions, we consider the squared euclidean distance ($D^{\pi}_{\pi'}(s) = \norm{\pi(\cdot|s) - \pi'(\cdot|s)}_2^2$) as the Bregman divergence.
It should be clear all prior analyses of BPMD carry through without any essential change. 
Due to the symmetry of the transition, the optimal policy $\pi^*$ should satisfy $\pi^*(R|s) = 1$ for all $s \neq s_A$, regardless of the discount factor $\gamma$.

Let us consider starting BPMD from state $s_{-1} = s_A$.
By an induction argument, it should be clear that at any given iteration $k$, the only state that can be visited by the policy $\pi_k$ is the state $s_A$. 
Consequently, the policy never changes from the initial policy, and hence $\pi_k(\cdot|s) = \mathrm{Unif}(\cA)$ for all $s \neq s_A$, and can not be optimal.
\end{proof}

\begin{remark}\label{remark_motivate_explore_sampling}
Note that the prior example also warrants our restriction to the class of exploratory sampling distribution (Definition \ref{def_explore_sampling}) in Section \ref{subsec:BPMD}.
Specifically,  the distribution $\rho$ that places probability $1$ on state $s_A$ is not an exploratory distribution. The proof of Proposition \ref{prop_hard_example_on_policy} then shows that BPMD with this static sampling distribution does not find the optimal policy. 
\end{remark}

The construction of the hard instance in Figure \ref{fig_on_policy} relies on the observation that without any explicit exploration over states (e.g., traveling to $s_B$ so that information on $s_C$ can be obtained through $Q^\pi(s_B, R)$), the policy optimization process becomes myopic. 
Having discussed the necessity of Assumption \ref{assump_exploratory_on_policy} for the on-policy sampling scheme, 
we can now obtain the following linear convergence for non-strongly-convex regularizers. 

\begin{theorem}[$\mu=0$, on-policy sampling]\label{thrm_bpmd_nonsc_on_policy}
Suppose $h$ satisfies $\mu = 0$,
the on-policy sampling distributions $\cbr{\rho_k}$ satisfy Assumption \ref{assump_exploratory_on_policy},
and  the stepsizes $\cbr{\eta_k}$  satisfy
\begin{align}\label{stepsize_nonsc_on_policy}
\eta_k  \geq 
\eta_{k-1} \cdot \tfrac{\rho^\dagger_k}{\rho^\dagger_{k-1}} \norm{\tfrac{\rho_{k-1}}{\rho_k}}_\infty \sbr{ 1 - (1-\gamma) \rho_k^\dagger}^{-1},
\end{align}
Then it follows 
\begin{align*}
 \EE\sbr{f(\pi_{k}) - f(\pi^*) 
+ 
  \tfrac{\varrho}{\eta_{k-1}}   \phi(\pi^*, \pi_k) }
\leq    \rbr{1 - (1-\gamma) \varrho}^k \sbr{f(\pi_0) - f(\pi^*) 
+ \tfrac{\log \abs{\cA}}{\eta_0} }, ~\forall k \geq 1.
\end{align*}
\end{theorem}

\begin{proof}
Note that with KL-divergence, update \eqref{eq:bpmd_update} does not change the support of the policy, and hence the structure of the Markov chains induced by the policies are time-invariant,  which implies $\cS_k = \cS_{k+1}$ holds for all $k$.
The desired claim then follows by taking $\mu = 0$ in the proof of Theorem \ref{thrm:bpmd_sc},  then following exactly the same lines, and using Assumption \ref{assump_exploratory_on_policy} in inequality $(b)$ therein. 
\end{proof}

It should be noted that since $\rho_{k} \in \sigma( \cbr{s_t}_{t < k})$, i.e., the $\sigma$-algebra  generated by the past sampled states, the right hand side of \eqref{stepsize_nonsc_on_policy} also belongs to $\sigma( \cbr{s_t}_{t < k})$ and is completely determined once the $(k-1)$-th iteration of BPMD is completed. 
Hence the condition \eqref{stepsize_nonsc_on_policy} is well defined. 
One can then use an upper-bounding estimate of the right hand side of \eqref{stepsize_nonsc_on_policy} to specify $\eta_k$.
Sublinear convergence for non-strongly-convex regularizers and convergence for strongly-convex regularizers follow similar discussions as before, and we omit their details here.

To summarize our discussions in Section \ref{subsec_sampling_comparison}, both the uniform sampling and the on-policy sampling have their unique strength and limitations. 
For the uniform sampling, it is guaranteed that BPMD achieves the same worst-case computational efficiency as that of batch PG methods. 
On the other hand, it requires explicit exploration over the state space. 
In comparison, on-policy sampling admits easy and natural implementation, without any explicit exploration. 
The price of this convenience is the global and non-asymptotic convergence being much more fragile, as suggested by Proposition \ref{prop_hard_example_on_policy}.

\section{Accelerating BPMD with Instance-dependent Sampling}\label{sec_effect_sampling}
We have so far established that with the uniform sampling scheme, BPMD achieves the same  worst-case computational complexity as the fast batch PG methods \cite{xiao2022convergence, lan2021policy, li2022homotopic}, while having a cheaper per-iteration computation.
 Though we believe this already makes BPMD favorable for solving large-state-space MDPs, the worst-case nature of the above claim naturally raises the following important question.

 
{\it Can we accelerate  BPMD in an instance-dependent fashion, so that it outperform the fast batch PG methods?} 
Or equivalently, {\it can we choose better sampling distributions that outperform the uniform sampling scheme?}

To answer above questions, this section is dedicated to discuss the role of sampling distributions in affecting the computational efficiency of the BPMD method.
For simplicity, we will focus on BPMD with a static sampling scheme, denoted by $\rho$. 
Let us briefly focus on three distinct, and conceptually simple scenarios, which we point out some important observations that prepare us for technical developments.
Each of this scenario contains more information about the optimal policy compared to its predecessor. 


{\it $\bullet$ Scenario I - when no information is available}. 
In view of Table \ref{tab_iteration_comp_compare_static}, it should be clear that to minimize the iteration complexity for finding an $\epsilon$-optimal policy, one should choose a sampling distribution with large enough $\rho^\dagger$.
Note that in the worst-case $\mathrm{supp}(\nu^*) = \cS$, and $\max_{\rho \in \Delta_\cS} \rho^\dagger = 1/ \abs{\cS}$, and the maximum is attained only when taking $\rho$ to be the uniform distribution over $\cS$, denoted by $\mathrm{Unif}(\cS)$. 
Thus when no information is available on the underlying MDP, the uniform distribution seems a reasonable choice.

{\it $\bullet$ Scenario II - when $\mathrm{supp}(\nu^*)$ is available}.
Suppose $\cS^* = \mathrm{supp}(\nu^*)$ is known, then by letting the sampling distribution $\rho$ be the uniform distribution over $\cS^*$, denoted by $\mathrm{Unif}(\cS^*)$.
It is immediate to see that this sampling distribution is exploratory, with $\rho^\dagger = 1/ \abs{\cS^*}$, thus
 improving over the worst-case iteration complexity of using $\mathrm{Unif}(\cS)$ (and consequently the fast batch PG methods), by a factor of $\abs{\cS} / \abs{\cS^*}$. 
The improvement can be substantial if $\abs{\cS}^* \ll \abs{\cS}$, i.e., when the optimal policy visits a vanishing subset of states.

{\it $\bullet$ Scenario III - when $ \nu^*$ is available}.
Compared to the support, 
the distribution $\nu^*$ itself further prioritizes learning the policy for the state that is frequently visited by the optimal policy.
As an extreme case, one can even avoid learning policy for state $s$ if $\nu^*(s) = 0$.
Given this observation, it could be expected that using $\nu^*$ as the sampling distribution would easily beat the instance-agnostic uniform distribution $\mathrm{Unif}(\cS)$, or $\mathrm{Unif}(\cS^*)$ that only exploits the support information of $\nu^*$. 
However, we will show that the situation is in fact more nuance. 
Specifically, we consider the following simple example.

\begin{figure}[htb!]
\centering 
    \includegraphics[width=0.45\textwidth]{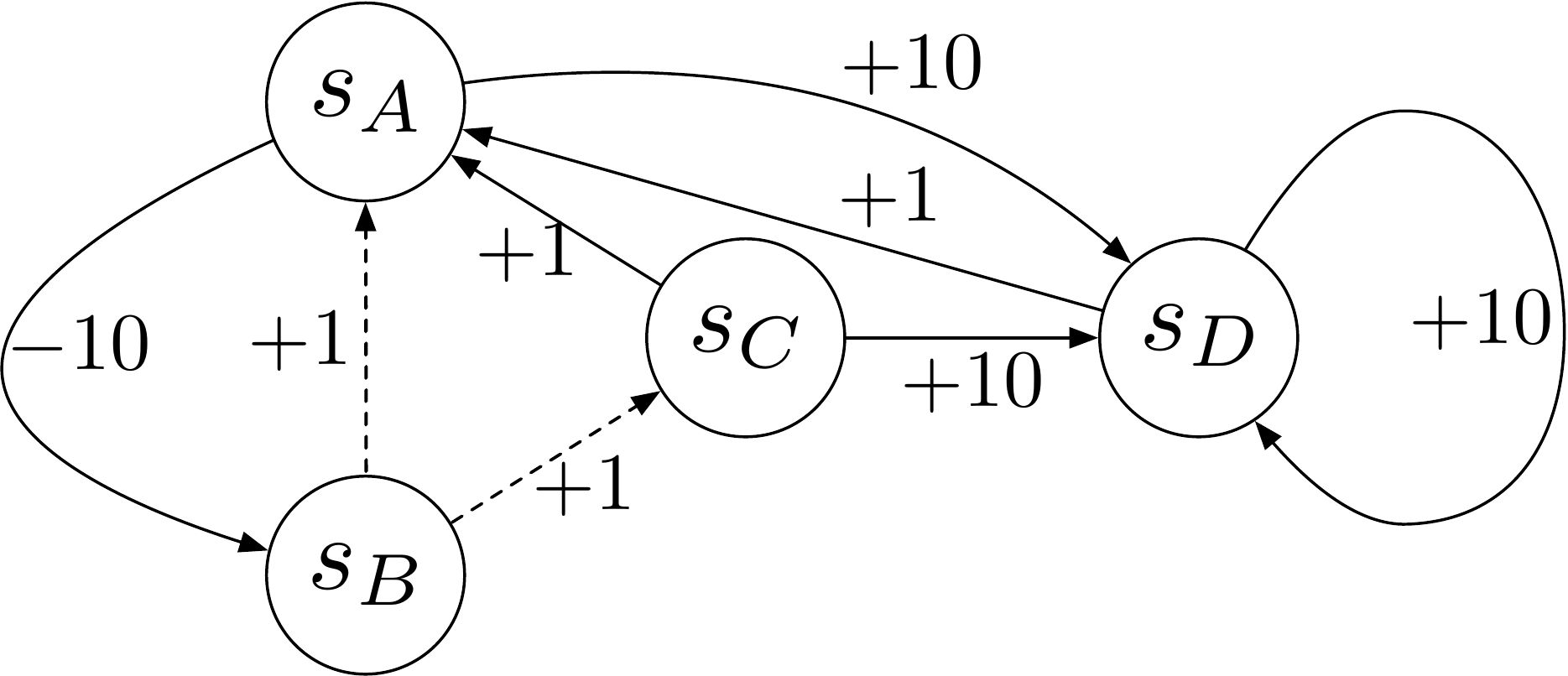}
  \caption{ 
 An MDP instance for which BPMD with static sampling $\nu^*$ is slower than BPMD with $\mathrm{Unif}(\cS)$. 
 Each state is associated with actions $\cA= \cbr{L, R}$, denoting going left and right. 
 Except $(s_B, L, s_A)$ and $(s_B, R, s_C)$, each
  arc $(s,a, s')$ specifies the transition of starting from state $s$, making action $a$ and then transits to state $s'$. The value of the arc denotes the cost $c(s,a)$.
  In addition, $\PP(s_C|s_B, a) = p$ for any $a \in \cA$.
  }
  \label{fig_optimal_visitation_sampling}
\end{figure}

\begin{proposition}\label{prop_optimal_visitation_worse_than_uniform}
Consider the MDP with its transition diagram and cost illustrated in Figure \ref{fig_optimal_visitation_sampling}.
The only stochastic transition appears at state $s_B$, for which the state transits to $s_C$ with probability $p \in (0,1)$, regardless of the action made at $s_B$.
Let the initial policy be given by 
\begin{align*}
\pi_0(L|s_A) = \pi_0(L | s_B) = \pi_0(L |s_D) = 1,~  \pi(\cdot|s_C) = \mathrm{Unif}(\cA).
\end{align*}
Let $\cbr{\pi_k^*}$ denote the policy generated by BPMD with the static sampling distribution $\nu^*$.
Let $\cbr{\pi_k^{\mathrm{unif}}}$ denote the policy generated by BPMD with the static sampling distribution $\mathrm{Unif}(\cS)$. 
Suppose both variants of BPMD use the same stepsize scheme, 
and $p < 1/2$, 
then we have 
\begin{align}\label{optimal_visitation_worse_than_uniform}
\EE \big[V^{\pi_k^*}(s) \big] > \EE \big[V^{\pi_k^{\mathrm{unif}}}(s) \big], ~\forall s \in \cS, ~ \forall k \geq 0.
\end{align}
\end{proposition}

\begin{proof}
It should be clear that regardless the values of the discount factor $\gamma$ and the probability $p$, the optimal policy $\pi^*$ is to always go left: $\pi^*(L |s) = 1$ for any $s \in \cS$.
Thus, the initial policy is only sub-optimal in state $s_C$.
It is worth mentioning indeed $V^{\pi_0}(s) > V^*(s)$ for every $s \in \cS$, given the sub-optimality of $\pi_0$ at state $s_C$.
Moreover,  one can easily calculate the $\nu^*$ as 
\begin{align}\label{example_mdp_optimal_visitation}
\nu^*(s_A) = \nu^*(s_B) = \tfrac{1}{2+p}, ~ \nu^*(s_C) = \tfrac{p}{2+ p} < \tfrac{p}{2}, ~ \nu^*(s_D) = 0.
\end{align}

Given the update rule \eqref{eq:bpmd_update} of BPMD and the optimality of $\pi_0$ at state $\cbr{s_A, s_B, s_D}$, it is clear that the policy $\pi_k$ will not change unless $s_k = s_C$. 
%
Given this  observation, at any iteration $k$ and for any $s\in \cS$, the value $V^{\pi_k}(s)$ is a deterministic and strictly decreasing function of the random variable $I_k = \tsum_{t = 0}^{k-1} \mathbbm{1}_{\cbr{s_t = s_C}}$.
Hence to show \eqref{optimal_visitation_worse_than_uniform}, it suffices to show $\PP^{*}(I_k = m) < \PP^{\mathrm{unif}}(I_k = m)$ for any $m < k$,
where $\PP^{*}$ denotes taking probability with respect to $\cbr{\pi_k^*}$, and $\PP^{\mathrm{unif}}$ denotes taking probability with respect to $\cbr{\pi_k^{\mathrm{unif}}}$.
Note that in view of \eqref{example_mdp_optimal_visitation},  this is trivial to verify provided $p/2 < 1/\abs{\cS}$, hence the proof is completed.
\end{proof}

There are several points regarding Proposition \ref{prop_optimal_visitation_worse_than_uniform}  worthy of attentions.
\begin{enumerate}[noitemsep,topsep=0pt]
\item[{\bf (A)}] Static sampling with $\nu^*$ gives strictly worse global convergence guarantee than the instance-agnostic uniform sampling.
\item[{\bf (B)}] More importantly,
in light of \eqref{optimal_visitation_worse_than_uniform}, which is a state-wise value comparison, 
 the worse convergence of sampling with $\nu^*$ can not be mitigated 
 by switching $\nu^*$ with other measure $\nu \in \Delta_\cS$  in the weighted policy optimization objective \eqref{eq:mdp_single_obj}.
This shows that the potential pitfall of sampling with $\nu^*$ is independent with the weights in optimization objective.
\item[{\bf (C)}] The construction heavily exploits the following fact: the initial policy is carefully constructed in a way that frequently sampled states by $\nu^*$ are already optimal, and hence the initial value is already close-to-optimal.
 It does not rule out the possibility that for common initializations (e.g., uniform policy), BPMD can benefit from sampling with $\nu^*$, at least in the early stage.
\end{enumerate}

%

%

Up to now, our discussions in Scenario II and III are purely idealistic, as neither $\cS^*$, or the distribution $\nu^*$ can be known exactly before learning.
Nevertheless, within these ideal cases we have identified some important observations that hint upon what can and cannot be expected regarding the potential acceleration of BPMD with instance-dependent sampling schemes. 

Our next main result is to develop {\it adaptive convergence of BPMD} that aligns with above discussions.
Moreover, the adaptive convergence applies to the setting where
the sampling scheme is constructed with 
only approximate information on the particular instance.
This nice feature thus allows the result to be applicable to  scenarios where this approximate information can be inferred through domain expertise.

\begin{lemma}\label{lemma_adaptive_convergence_general}
Let $\rho$ be the static sampling policy, not necessarily exploratory in the sense of Definition \ref{def_explore_sampling},
and let $\cS_\rho$ denote the support of $\rho$. Let $\cS^*$ denote the support of $\nu^*$.
Suppose $0 \leq c(s,a) \leq \overline{c}$ and $0 \leq h^{\pi}(s) \leq \overline{h}$, for any $(s,a)$ and $\pi \in \Pi$.

For any $\cH \subseteq \cS_\rho$, define $\nu^{\#}_\cH = \tsum_{s \in \cS^* / \cH} \nu^*(s)$, and $\rho^\dagger_{\cH} = \min_{s \in \cH \cap \cS^*} \rho(s)$.
Then the policy pair $(\pi_k, \pi_{k+1})$ generated by BPMD satisfies 
\begin{align}
 \EE_{s_k} \sbr{f(\pi_{k+1}) - f(\pi^*)}  
 \leq &
\rbr{1 - (1-\gamma) \rho^\dagger_{\cH}} \sbr{f(\pi_k) - f(\pi^*) }
+ \rho^\dagger_{\cH} \sbr{\tfrac{1}{\eta_k} + \mu (1- \rho^\dagger_{\cH}) }    \tsum_{s \in \cH}  \tfrac{\nu^*(s)}{\rho(s)}  D^{\pi^*}_{\pi_k}(s) \label{eq_adaptive_convergence_general}  \\
& - \rho^\dagger_{\cH} \sbr{\tfrac{1}{\eta_k} + \mu }    \EE_{s_k} \sbr{   \tsum_{s \in \cH}  \tfrac{\nu^*(s)}{\rho(s)} D^{\pi^*}_{\pi_{k+1}}(s)} 
+  \tfrac{ 2 \nu^{\#}_{\cH}  \rho^\dagger_{\cH} ( \overline{c} + \overline{h})}{1-\gamma}
. \nonumber 
\end{align}
\end{lemma}

\begin{proof}
Recall that \eqref{scBPMD:value_at_update_2} still holds with $\rho = \rho$, that is, 
\begin{align}
 \eta_k \tfrac{\nu^*(s_k)}{\rho(s_k)}  \big[\inner{Q^{\pi_k}(s_k, \cdot) }{\pi_k(\cdot|s_k) - \pi^*(\cdot| s_k)  }  + h^{\pi_k}(s_k)  - h^{\pi^*}(s_k)\big]  + 
\eta_k   \tfrac{\nu^*(s_k)}{\rho(s_k)}  \sbr{V^{\pi_{k+1}}(s_k) - V^{\pi_{k}}(s_k)} \label{adaptive_starting_eq} 
\\  +    \tfrac{\nu^*(s_k)}{\rho(s_k)}  D^{\pi_{k+1}}_{\pi_k}(s_k)  
\leq 
  \tfrac{\nu^*(s_k)}{\rho(s_k)}  D^{\pi^*}_{\pi_k}(s_k)  - (1+ \eta_k \mu)    \tfrac{\nu^*(s_k)}{\rho(s_k)}  D^{\pi^*}_{\pi_{k+1}}(s_k) .  \nonumber
\end{align}
Let us denote $\cS_\rho = \mathrm{supp}(\rho)$, and let $\cS_k^c = \cS_\rho \setminus \cbr{s_k}$.
We next multiply both sides of the above inequality by the indicator function $\mathbbm{1}_{\cbr{s_k \in \cH }}$,
where $\cH \subseteq \cS_\rho$, which gives
\begin{align*}
&  \eta_k \tfrac{\nu^*(s_k)}{\rho(s_k)} \mathbbm{1}_{\cbr{s_k \in \cH }}  \big[\inner{Q^{\pi_k}(s_k, \cdot) }{\pi_k(\cdot|s_k) - \pi^*(\cdot| s_k)  }  + h^{\pi_k}(s_k)  - h^{\pi^*}(s_k)\big]    \\
& ~~~ +  \eta_k   \tfrac{\nu^*(s_k)}{\rho(s_k)} \mathbbm{1}_{\cbr{s_k \in \cH }} \sbr{V^{\pi_{k+1}}(s_k) - V^{\pi_{k}}(s_k)}    \\ 
 \leq &
  \tfrac{\nu^*(s_k)}{\rho(s_k)} \mathbbm{1}_{\cbr{s_k \in \cH }}  D^{\pi^*}_{\pi_k}(s_k)  - (1+ \eta_k \mu)    \tfrac{\nu^*(s_k)}{\rho(s_k)} \mathbbm{1}_{\cbr{s_k \in \cH }}  D^{\pi^*}_{\pi_{k+1}}(s_k) . 
\end{align*}
Observing that \eqref{scBPMD:d_at_other} still holds, we then multiply both sides of \eqref{scBPMD:d_at_other} by $ \nu^*(s) \mathbbm{1}_{\cbr{s \in \cH}}/\rho(s)$,  and  sum up over $\cS_k^c$ together with the above inequality, which gives 
\begin{align}
&  \eta_k \tfrac{\nu^*(s_k)}{\rho(s_k)}  \mathbbm{1}_{\cbr{s_k \in \cH }}    \big[\inner{Q^{\pi_k}(s_k, \cdot) }{\pi_k(\cdot|s_k) - \pi^*(\cdot| s_k)  }  + h^{\pi_k}(s_k)  - h^{\pi^*}(s_k)\big] \nonumber  \\ 
& ~~~~~ + 
  \eta_k  \mathbbm{1}_{\cbr{s_k \in \cH }}   \tfrac{\nu^*(s_k)}{\rho(s_k)}  \sbr{V^{\pi_{k+1}}(s_k) - V^{\pi_{k}}(s_k)}   \nonumber
\\  
 \leq &  
  \tsum_{s \in \cS_\rho }  \tfrac{\nu^*(s)}{\rho(s)}   \mathbbm{1}_{\cbr{s \in \cH }}   D^{\pi^*}_{\pi_k}(s)  - (1+ \eta_k \mu)    \tsum_{s\in \cS_\rho}  \tfrac{\nu^*(s)}{\rho(s)}   \mathbbm{1}_{\cbr{s \in \cH }}  D^{\pi^*}_{\pi_{k+1}}(s) \nonumber
\\ & ~~~~~  + \eta_k \mu  \tsum_{s \in \cS_k^c }  \tfrac{\nu^*(s)}{\rho(s)}    \mathbbm{1}_{\cbr{s \in \cH }}  D^{\pi^*}_{\pi_{k}}(s)  \nonumber.
\end{align}
Further multiplying both sides of \eqref{ineq:value_decrease} by $\nu^*(s) \mathbbm{1}_{\cbr{s \in \cH}}/\rho(s)$, and summing up over $s \in  \cS_k^c$ together with the above relation, we arrive at 
\begin{align}
&  \eta_k \tfrac{\nu^*(s_k)}{\rho(s_k)}  \mathbbm{1}_{\cbr{s_k \in \cH }}    \big[\inner{Q^{\pi_k}(s_k, \cdot) }{\pi_k(\cdot|s_k) - \pi^*(\cdot| s_k)  }  + h^{\pi_k}(s_k)  - h^{\pi^*}(s_k)\big] \nonumber  \\ 
& ~~~~~ + 
  \eta_k  \tsum_{s \in \cS_\rho}  \mathbbm{1}_{\cbr{s \in \cH }}   \tfrac{\nu^*(s)}{\rho(s)}  \sbr{V^{\pi_{k+1}}(s) - V^{\pi_{k}}(s)}   \nonumber
\\  
 \leq &  
  \tsum_{s \in \cS_\rho }  \tfrac{\nu^*(s)}{\rho(s)}   \mathbbm{1}_{\cbr{s \in \cH }}   D^{\pi^*}_{\pi_k}(s)  - (1+ \eta_k \mu)    \tsum_{s\in \cS_\rho}  \tfrac{\nu^*(s)}{\rho(s)}   \mathbbm{1}_{\cbr{s \in \cH }}  D^{\pi^*}_{\pi_{k+1}}(s) \nonumber
\\ & ~~~~~  + \eta_k \mu  \tsum_{s \in \cS_k^c }  \tfrac{\nu^*(s)}{\rho(s)}    \mathbbm{1}_{\cbr{s \in \cH }}  D^{\pi^*}_{\pi_{k}}(s)  \nonumber.
\end{align}

We can now take expectation w.r.t. $s_k \sim \rho$, which gives
\begin{align}]\label{adaptive_expectation_raw}
&  \eta_k \underbrace{\tsum_{s \in \cS_\rho } \nu^*(s)  \mathbbm{1}_{\cbr{s \in \cH }}    \big[\inner{Q^{\pi_k}(s, \cdot) }{\pi_k(\cdot|s) - \pi^*(\cdot| s)  }  + h^{\pi_k}(s)  - h^{\pi^*}(s)\big]}_{(T_1)} \\
& ~~~~~ + 
  \eta_k \underbrace{ \tsum_{s \in \cH}   \tfrac{\nu^*(s)}{\rho(s)} \EE_{s_k} \sbr{V^{\pi_{k+1}}(s) - V^{\pi_{k}}(s)} }_{(T_2)}  \nonumber
\\  
&  \leq   
  \tsum_{s \in  \cH }  \tfrac{\nu^*(s)}{\rho(s)}    D^{\pi^*}_{\pi_k}(s)  - (1+ \eta_k \mu)    \tsum_{s\in \cH}  \tfrac{\nu^*(s)}{\rho(s)}    \EE_{s_k} \big[ D^{\pi^*}_{\pi_{k+1}}(s) \big]+ \eta_k \mu  \underbrace{\EE_{s_k} \big[ \tsum_{s \in \cS_k^c \cap \cH }  \tfrac{\nu^*(s)}{\rho(s)}     D^{\pi^*}_{\pi_{k}}(s) \big]}_{(T_3)}  \nonumber.
\end{align}
To simplify the above relation, we first observe that 
\begin{align*}
(T_1) &= \tsum_{s \in \cH } \nu^*(s)   \big[\inner{Q^{\pi_k}(s, \cdot) }{\pi_k(\cdot|s) - \pi^*(\cdot| s)  }  + h^{\pi_k}(s)  - h^{\pi^*}(s)\big] \\
& \overset{(a)}{\geq} \tsum_{s \in \cS^* } \nu^*(s)   \big[\inner{Q^{\pi_k}(s, \cdot) }{\pi_k(\cdot|s) - \pi^*(\cdot| s)  }  + h^{\pi_k}(s)  - h^{\pi^*}(s)\big] 
- \tsum_{s \in \cS^* \setminus \cH } \nu^*(s)  \big[ \tfrac{\overline{c} + \overline{h}}{1-\gamma} + \overline{h} \big] \\
& \overset{(b)}{\geq} (1-\gamma)\sbr{ f(\pi_k) - f(\pi^*)} - \tfrac{2 \nu^{\#}_{\cH}  ( \overline{c} + \overline{h})}{1-\gamma},
\end{align*}
where $(a)$ follows from the $ \abs{ \inner{Q^{\pi_k}(s, \cdot) }{\pi_k(\cdot|s) - \pi^*(\cdot| s) } } \leq \norm{Q^{\pi_k}(s, \cdot)}_\infty \leq (\overline{c} + \overline{h})/(1-\gamma)$, and $0 \leq h^{\pi}(s) \leq \overline{h}$,
and $(b)$ follows from identity \eqref{vi_pmd}, and the definition of $\nu^{\#}_\cH = \tsum_{s \in \cS^* / \cH} \nu^*(s)$.
Next, we also know that 
\begin{align*}
(T_3 )
 = \tsum_{s' \in \cS_\rho} \rho(s') \tsum_{ s \in \cS_\rho \cap \cH \setminus \cbr{s'} } \tfrac{\nu^*(s)}{\rho(s)} D^{\pi^*}_{\pi_k}(s) & \overset{(a)}{=} \tsum_{s' \in \cS_\rho} \rho(s') \tsum_{ s \in \cH \setminus \cbr{s'} } \tfrac{\nu^*(s)}{\rho(s)} D^{\pi^*}_{\pi_k}(s) \\
& \overset{(b)}{=} \tsum_{s \in \cH} \tsum_{s' \in \cS_\rho \setminus \cbr{s } }  \rho(s') \tfrac{\nu^*(s)}{\rho(s)} D^{\pi^*}_{\pi_k}(s) \\
& = \tsum_{s \in \cH} (1- \rho(s)) \tfrac{\nu^*(s)}{\rho(s)} D^{\pi^*}_{\pi_k}(s) \\
& \overset{(c)}{\leq}  (1 - \rho^\dagger_{\cH}) \tsum_{s \in \cH}  \tfrac{\nu^*(s)}{\rho(s)} D^{\pi^*}_{\pi_k}(s) ,
\end{align*}
where $(a)$ follows from $\cH \subseteq \cS_\rho$;
$(b)$ follows from changing the order of summation and using $\cH \subseteq \cS_\rho$;
$(c)$ follows from the definition of $\rho^\dagger_{\cH}$.
Finally, using the fact that $V^{\pi_{k+1}}(s) \leq V^{\pi_k}(s) $ for any $s  \in \cS$ based from \eqref{ineq:value_decrease}, we have 
\begin{align*}
(T_2) &  \geq \tfrac{1}{\rho^\dagger_{\cH}} \tsum_{s \in \cH}   \nu^*(s) \EE_{s_k} \sbr{V^{\pi_{k+1}}(s) - V^{\pi_{k}}(s)}
+ \tfrac{1}{\rho^\dagger_{\cH}} \tsum_{s \notin \cH}   \nu^*(s) \EE_{s_k} \sbr{V^{\pi_{k+1}}(s) - V^{\pi_{k}}(s)}  \\
& = \tfrac{1}{\rho^\dagger_{\cH}} \EE_{s_k} \sbr{ f(\pi_{k+1} ) - f(\pi_k)}.
\end{align*}

By using the above observations on $(T_1)$, $(T_2)$,  and $(T_3)$ with \eqref{adaptive_expectation_raw}, we obtain  
\begin{align*}
&  \eta_k \sbr{ (1-\gamma)\sbr{ f(\pi_k) - f(\pi^*)} - \tfrac{2 \nu^{\#}_{\cH}  ( \overline{c} + \overline{h})}{1-\gamma}} +  \tfrac{\eta_k}{\rho^\dagger_{\cH}} \EE_{s_k} \sbr{ f(\pi_{k+1} ) - f(\pi_k)}   \nonumber
\\  
&  \leq   
  \tsum_{s \in  \cH }  \tfrac{\nu^*(s)}{\rho(s)}    D^{\pi^*}_{\pi_k}(s)  - (1+ \eta_k \mu)    \tsum_{s\in \cH}  \tfrac{\nu^*(s)}{\rho(s)}    \EE_{s_k} \big[ D^{\pi^*}_{\pi_{k+1}}(s) \big]+ \eta_k \mu   (1 - \rho^\dagger_{\cH}) \tsum_{s \in \cH}  \tfrac{\nu^*(s)}{\rho(s)} D^{\pi^*}_{\pi_k}(s)   \nonumber.
\end{align*}
Dividing both sides of above relation by $\eta_k$ and  the desired claim follows after simple rearrangement.
\end{proof}

Lemma \ref{lemma_adaptive_convergence_general} can be treated as the a refined version of Lemma \ref{lemma:bpmd_generic_recursion} with $\rho_k = \rho$ in the following sense.
First, Lemma \ref{lemma_adaptive_convergence_general} does not require the sampling distribution $\rho$ to be exploratory. 
Indeed,  whenever $\rho$ is not exploratory and hence $\cS_\rho \subset \cS^*$, 
by setting $\cH = \cS_\rho$ in \eqref{eq_adaptive_convergence_general}, 
 the price BPMD pays can be upper bounded by $\nu^{\#}_{\cH} \rho^\dagger_\cH ( \overline{c} + \overline{h})/ (1-\gamma)$, which directly depends on the probability mass not explored by $\rho$, via $\nu^{\#}_{\cH} $.
 
Second,  \eqref{lemma_adaptive_convergence_general} in Lemma \ref{lemma_adaptive_convergence_general}  in fact induces {\it a family of} upper bounds, each associated with a particular choice of subset $\cH$.
When $\cS_\rho \supseteq \cS^*$ and $\cH = \cS^*$, it is immediate that Lemma \ref{lemma_adaptive_convergence_general} recovers Lemma \ref{lemma:bpmd_generic_recursion} by taking $\rho_k = \rho$ therein.

By specializing Lemma \ref{lemma_adaptive_convergence_general}, we can immediately obtain the following adaptive convergence bounds for BPMD with sampling distribution $\rho$.
The notion of adaptivity will be discussed in detail after the statement of the theorem. 
For brevity, we use SC in short for strongly-convex and NSC for non-strongly-convex.

\begin{theorem}[Adaptive Convergence of BPMD]\label{thrm_adaptive_convergences}
Let $\rho_k = \rho$ for all $k \geq 0$ in the BPMD method, and let $\cS_\rho$ be the support of $\rho$.
For any $\cH \subseteq \cS_\rho$, let $(\rho^\dagger_{\cH}, \nu^{\#}_{\cH})$ be defined as in Lemma \ref{lemma_adaptive_convergence_general},
and let $\phi_{\cH}(\pi^*, \pi) = \sum_{s \in \cH} \nu^*(s) D^{\pi^*}_{\pi}(s)$.
Then the convergence of BPMD satisfies the following bounds simultaneously for all $\cH \subseteq \cS_\rho$.

 \underline{I}. SC regularizers: 
Let  $\eta_t = \eta$ for all $t \geq 0$, with
$
1 + \eta \mu \geq \tfrac{1}{\gamma},
$
then  for all $k \geq 1$,
\begin{align*}
& \EE\sbr{f(\pi_{k}) - f(\pi^*) + \rho^\dagger_{\cH} (\tfrac{1}{\eta} + \mu) \phi_{\cH}(\pi^*, \pi_k) }   \\
\leq &  \rbr{1 - (1-\gamma) \rho^\dagger_{\cH}}^k \sbr{f(\pi_0) - f(\pi^*) 
+ \rbr{\tfrac{1}{\eta} + \mu (1- \rho^\dagger_{\cH}) }  \log \abs{\cA} }
+ 
 \tfrac{2 \nu^{\#}_{\cH}   ( \overline{c} + \overline{h})}{(1-\gamma)^2}
.
\end{align*}

 \underline{II}. NSC regularizers - linear convergence: 
Let stepsizes $\cbr{\eta_k}$ staisfy
$
\eta_k  \geq 
\eta_{k-1} \cdot  [ 1 - (1-\gamma) \rho^\dagger_{\cH}]^{-1},
$
then 
\begin{align*}
 \EE\sbr{f(\pi_{k}) - f(\pi^*) 
+ 
  \tfrac{\rho^\dagger_{\cH}}{\eta_{k-1}}   \phi_{\cH}(\pi^*, \pi_k) }
\leq    \rbr{1 - (1-\gamma) \rho^\dagger_{\cH}}^k \sbr{f(\pi_0) - f(\pi^*) 
+ \tfrac{\log \abs{\cA}}{\eta_0} }
+ 
 \tfrac{ 2\nu^{\#}_{\cH}   ( \overline{c} + \overline{h})}{(1-\gamma)^2}
, ~ \forall k \geq 1.
\end{align*}
In particular, the $\rho$-agnostic stepsize scheme $\eta_k  \geq 
\eta_{k-1} \cdot  \gamma^{-1} $ yields the above result.

 \underline{III}. NSC regularizers - sublinear convergence: 
 For any constant stepsize $\eta > 0$, 
\begin{align*}
 \EE \sbr{  f(\pi_k) - f(\pi^*)}
 \leq  \tfrac{ \log \abs{\cA} /\eta +  f(\pi_0) - f(\pi^*)}{k (1-\gamma) \rho^\dagger_{\cH} }
 + 
 \tfrac{2 \nu^{\#}_{\cH}   ( \overline{c} + \overline{h})}{(1-\gamma)^2}
, ~ \forall k \geq 1.
\end{align*}
\end{theorem}

\begin{proof}
The proof follows the same arguments as the proofs of Corollary \ref{cor_sc_linear_static}, \ref{cor_non_sc_sublinear}, and \ref{cor_non_sc_static_linear}, but now we will apply Lemma \ref{lemma_adaptive_convergence_general} instead of Lemma \ref{lemma:bpmd_generic_recursion}.
\end{proof}

It is worth pointing out that in Theorem \ref{thrm_adaptive_convergences},
 the convergence results we obtain yield a family of upper bounds for the optimality gap, each corresponding to a unique $\cH$.
 Note that the selection of $\cH$ is only for the purpose of analytically estimating the performance bound of BPMD, 
and  the actual implementation of BPMD does not necessarily require $\cH$.
This  immediately implies that the obtained convergence is implicitly adaptive, in the sense that one can properly choose an $\cH$ that minimizes the obtained upper bound, and this minimization process naturally adapts to the sampling distribution $\rho$ and the {\it unknown} $\nu^*$.

\subsection{Acceleration with Adaptivity}\label{subsec_acc_by_adaptivity}
We are now ready to discuss the power brought by the adaptivity, which can lead to acceleration with only approximate instance information, defined as follows.

\begin{definition}[$M$-surrogate]\label{def_m_good}
A distribution $\rho \in \Delta_{\cS}$ is called an $M$-surrogate of $\nu^*$ if $\norm{\nu^* / \rho}_\infty \leq M$ for some (potentially large) $M \geq 1$.
\end{definition}

Note that Definition \ref{def_m_good} requires $\cS_\rho \supseteq \cS^*$, i.e., $\cS_\rho$ being a conservative estimate of $\cS^*$.
In particular, $\cS_\rho = \cS$ trivially satisfies this condition. 
Moreover, $M = 1$ if and only if $\rho = \nu^*$, and hence $M$ can be understood as the proxy on the quality of $\rho$.
We will refer to $\rho_{\mathrm{unif}} = \mathrm{Unif}(\cS_\rho)$ as the uniform sampling.

\begin{theorem}\label{thrm_speedup_general}
Assume the same setup as Theorem \ref{thrm_adaptive_convergences}.
Suppose $\rho$ is an $M$-surrogate for $M \geq 1$.
Let $\rho_{\cH}^{\#}  =  \tsum_{s \in \cS^* / \cH} \rho(s) \leq \tsum_{s \in \cS / \cH} \rho(s) $.

 \underline{I}. SC regularizers: 
Let  $\eta_t = \eta$ for all $t \geq 0$, with
$
1 + \eta \mu \geq \tfrac{1}{\gamma},
$
then for all $k \geq 1$,
\begin{align*}
& \EE\sbr{f(\pi_{k}) - f(\pi^*) + \rho^\dagger_{\cH} (\tfrac{1}{\eta} + \mu) \phi_{\cH}(\pi^*, \pi_k) }   \\
\leq  & \rbr{1 - (1-\gamma) \rho^\dagger_{\cH}}^k \sbr{f(\pi_0) - f(\pi^*) 
+ \rbr{\tfrac{1}{\eta} + \mu}  \log \abs{\cA} }
+ 
 \tfrac{ 2 M \rho^{\#}_{\cH}   ( \overline{c} + \overline{h})}{(1-\gamma)^2}
.
\end{align*}

 \underline{II}. NSC regularizers - linear convergence: 
Let $\cbr{\eta_k}$ staisfy
$
\eta_k  \geq 
\eta_{k-1} \cdot  \gamma^{-1}.
$
Then for all $ k \geq 1$,
\begin{align*}
 \EE\sbr{f(\pi_{k}) - f(\pi^*) 
+ 
  \tfrac{\rho^\dagger_{\cH}}{\eta_{k-1}}   \phi_{\cH}(\pi^*, \pi_k) }
\leq    \rbr{1 - (1-\gamma) \rho^\dagger_{\cH}}^k \sbr{f(\pi_0) - f(\pi^*) 
+ \tfrac{\log \abs{\cA}}{\eta_0} }
+ 
 \tfrac{ 2 M \rho^{\#}_{\cH}   ( \overline{c} + \overline{h})}{(1-\gamma)^2}.
\end{align*}

 \underline{III}. NSC regularizers - sublinear convergence: 
 For any constant stepsize $\eta > 0$, 
\begin{align*}
 \EE \sbr{  f(\pi_k) - f(\pi^*)}
 \leq  \tfrac{ \log \abs{\cA} /\eta +  f(\pi_0) - f(\pi^*)}{k (1-\gamma) \rho^\dagger_{\cH} }
 + 
 \tfrac{ 2 M \rho^{\#}_{\cH}   ( \overline{c} + \overline{h})}{(1-\gamma)^2}
, ~ \forall k \geq 1.
\end{align*}
\end{theorem}

\begin{proof}
The results follow from Theorem \ref{thrm_adaptive_convergences}, noting that $\nu_{\cH}^{\#} \leq M \rho^{\#}_{\cH}$. 
\end{proof}

Given a choice of $\cH$, the first term in the convergence bounds of Theorem \ref{thrm_speedup_general} measures how fast  the initial optimality gap decays, and benefits if a larger $\rho_{\cH}^\dagger$ can be obtained, simply by using a smaller $\cH$. 
The second term measures the limiting optimality gap  for the given choice of $\cH$, and can be reduced with a larger $\cH$.
Notably, choosing $\cH = \cS^*$ immediately recovers the bounds in Corollary \ref{cor_sc_linear_static}, \ref{cor_non_sc_sublinear}, \ref{cor_non_sc_static_linear}.
It is then clear that there exists a trade-off between the first and second term via the choice of $\cH$, and optimally balancing this trade-off with a properly chosen $\cH$ can lead to improved convergence bounds.

The previous observation demonstrates the potential of initial acceleration over the uniform sampling by using $\rho$-sampling. 
To illustrate, given an iteration $k$, suppose one can find $\cH$ with $\rho_{\cH}^\dagger \gg \rho_{\mathrm{unif}}^\dagger$, while the first term in the bound dominates the second term, then the rate of BPMD with  $\rho$-sampling is $\cO((1 - (1-\gamma) \rho_{\cH}^\dagger)^k)$ (resp. $\cO(1/ (k \rho_{\cH}^\dagger))$) for the linearly  (resp. sublinear) converging variants, faster than that of BPMD with $\rho_{\mathrm{unif}}$.
An important note to make is that for any $\cH \subsetneq \cS^*$, the second term will eventually dominate the first term.
Hence for showing global optimality, one eventually needs $\cH \supseteq \cS^*$, and the convergence will be bottlenecked by $\rho^\dagger$ (cf. Corollary \ref{cor_sc_linear_static}, \ref{cor_non_sc_sublinear}, \ref{cor_non_sc_static_linear}), resulting into a global iteration complexity that can be worse than sampling with $\rho_{\mathrm{unif}}$.
This observation clearly aligns with the instance constructed in Proposition \ref{prop_optimal_visitation_worse_than_uniform}, which shows that for a close-to-optimal initial policy, $\rho_{\mathrm{unif}}$  outperforms sampling with exact $\nu^*$. 

We now show that 
with a hybrid sampling scheme, BPMD can attain the benefits offered by both $\rho$-sampling and uniform sampling. 
In particular, the hybrid sampling scheme first adopts $\rho$-sampling, and switches to the uniform sampling after some iterations. 
With a properly chosen switching point, the hybrid sampling scheme can maintain the initial acceleration of sampling with $\rho$, leading to a global iteration complexity better than the uniform sampling.
We focus on linear convergence with NSC regularizers for brevity.
We will also assume $\phi(\pi^*, \pi) \leq \Phi$ for any $\pi$, and through a slightly more delicate analysis a similar claim can be obtained for the general case.

\begin{theorem}[Global Acceleration with Hybrid Sampling Scheme]\label{thrm_hybrid_sampling}
Suppose $\rho$ is an $M$-surrogate of $\nu^*$.
Consider BPMD applied to NSC regularizers 
 with the hybrid sampling scheme: 
\begin{enumerate}
\item Stage I - sample with $\rho$: $s_k \sim \rho$ for $0 \leq k < k_\tau$.
\item Stage II - sample with $\rho_{\mathrm{unif}}$: $s_k \sim \rho_{\mathrm{unif}}$  for $k \geq k_\tau$.
\end{enumerate}
In addition, consider the following stepsizes:
\begin{align}
\label{stepsize_hybrid_sampling}
\eta_k = \eta_{k-1}  \rbr{1 - (1-\gamma) \rho_{\cH}^\dagger}^{-1}, ~ 0 \leq k \leq k_\tau;
~~~
\eta_k = \eta_{k-1}  \rbr{1 - (1-\gamma) \rho_{\mathrm{unif}}^\dagger}^{-1}, ~ k > k_\tau.
\end{align}
 Let  $\Delta_0 = f (\pi_0) - f(\pi^*) 
+2 \Phi/ \eta_0$.
For $\cH \subseteq \cS_\rho$,
suppose $\epsilon_\tau =   \tfrac{2 M \rho^{\#}_{\cH}   ( \overline{c} + \overline{h})}{(1-\gamma)^2}$ satisfies $\epsilon_\tau < \Delta_0$,
Define $k_\tau  = \tfrac{1}{(1-\gamma) \rho_{\cH}^\dagger} \log (\tfrac{ \Delta_0}{\epsilon_\tau})$, 
then  for any $\epsilon > 0$, 
BPMD with hybrid sampling finds an $\epsilon$-optimal policy in 
\begin{align*}
\mathrm{K}_{\mathrm{hybrid}}(\epsilon) = 
\begin{cases}
\tfrac{1}{(1-\gamma) \rho_{\cH}^\dagger} \log (\tfrac{2 \Delta_0}{\epsilon})  , ~ \epsilon \geq  2\epsilon_\tau, \\
k_\tau
+ \tfrac{1}{(1-\gamma) \rho_{\mathrm{unif}}^\dagger} \log (\tfrac{2 \epsilon_\tau}{\epsilon})
  , ~ \epsilon < 2 \epsilon_\tau.
\end{cases}
\end{align*}
iterations. 
In comparison, BPMD with uniform sampling needs 
$
\mathrm{K}_{\mathrm{unif}}(\epsilon) =  \tfrac{1}{(1-\gamma) \rho_{\mathrm{unif}}^\dagger} \log \rbr{ \tfrac{\Delta_0}{\epsilon}}
$
iterations.

Consequently,   for $\epsilon \geq 2 \epsilon_\tau$, we have  
\begin{align*}
\text{Stage I:} ~~~  \tfrac{\mathrm{K}_{\mathrm{unif}}(\epsilon)}{ \mathrm{K}_{\mathrm{hybrid}}(\epsilon)} = \tfrac{\rho_{\cH}^\dagger \log(\Delta_0/\epsilon) }{\rho_{\mathrm{unif}}^\dagger \log(2 \Delta_0/ \epsilon)}
 ,
\end{align*}
and for any $\epsilon < 2 \epsilon_\tau$, we have 
\begin{align*}
\text{Stage II:} ~~~
 \mathrm{K}_{\mathrm{unif}}(\epsilon) - \mathrm{K}_{\mathrm{hybrid}}(\epsilon) = 
\tfrac{1}{1-\gamma} \sbr{\tfrac{1}{\rho_{\mathrm{unif}}^\dagger} \log (\tfrac{\Delta_0}{2 \epsilon_\tau}) - 
\tfrac{1}{\rho_{\cH}^\dagger} \log (\tfrac{\Delta_0}{ \epsilon_\tau}) 
  }. 
\end{align*}
\end{theorem}

\begin{proof}
Since BPMD uses $\rho$ as the sampling distribution before iteration $k_{\tau}$, then from Theorem \ref{thrm_adaptive_convergences} - II, 
\begin{align} \label{eq_phase_1_convergence}
 \EE\sbr{f(\pi_{k_{\tau}}) - f(\pi^*) 
  }
\leq    \rbr{1 - (1-\gamma) \rho^\dagger_{\cH}}^{k_{\tau}} \sbr{f(\pi_0) - f(\pi^*) 
+ \tfrac{\Phi}{\eta_0} }
+ 
 \tfrac{ 2 M \rho^{\#}_{\cH}   ( \overline{c} + \overline{h})}{(1-\gamma)^2}. 
\end{align}

On the other hand, as we switch the sampling policy to $\rho_{\mathrm{unif}} = \mathrm{Unif}(\cS_\rho)$ at iteration $k_\tau$,
in view of \eqref{eq_nonsc_linear_static_use_for_unif},
 we obtain that for any $k \geq k_\tau$,
\begin{align}
&  \EE\sbr{f(\pi_{k}) - f(\pi^*) } \label{eq_phase_2_convergence} \\
\leq   & 
  \rbr{1 - \rho_{\mathrm{unif}}^{\dagger} }^{k - k_\tau}
 \EE\sbr{f(\pi_{k_{\tau}}) - f(\pi^*) 
+ 
  \tfrac{\Phi}{ \eta_{k_{\tau} }}   }  \nonumber \\
    \leq &
    \rbr{1 -\rho_{\mathrm{unif}}^{\dagger}  }^{k - k_\tau}
     \rbr{1 - (1-\gamma) \rho^\dagger_{\cH}}^{k_{\tau}} \sbr{ f(\pi_0) - f(\pi^*) 
+ \tfrac{2 \Phi}{ \eta_0} }
+   \rbr{1 - \rho_{\mathrm{unif}}^{\dagger} }^{k - k_\tau}   \tfrac{ 2 M \rho^{\#}_{\cH}   ( \overline{c} + \overline{h})}{ (1-\gamma)^2}, \nonumber
\end{align}
where the last inequality follows from \eqref{eq_phase_1_convergence} and  the choice of stepsizes $\cbr{\eta_k}$.
Given the definition of $\Delta_0$ and $\epsilon_\tau$, 
for $\epsilon \geq 2 \epsilon_\tau$, we obtain from \eqref{eq_phase_1_convergence} that 
BPMD attains an optimality gap smaller than $\epsilon$
in 
$
\mathrm{K}_{\mathrm{hybrid}}(\epsilon) = \tfrac{1}{(1-\gamma) \rho_{\cH}^\dagger} \log (\tfrac{2 \Delta_0}{\epsilon})
\leq k_\tau
$
iterations.
On the other hand, for any $\epsilon < 2 \epsilon_\tau$, in view of \eqref{eq_phase_2_convergence} and the definition of $k_\tau$, BPMD finds an  $\epsilon$-optimal policy in at most
$
\mathrm{K}_{\mathrm{hybrid}}(\epsilon) = k_\tau
+ \tfrac{1}{(1-\gamma) \rho_{\mathrm{unif}}^\dagger} \log (\tfrac{2 \epsilon_\tau}{\epsilon})
$ iterations.

If BPMD runs with $\rho_{\mathrm{unif}}$ at the beginning, the number of iterations to find an $\epsilon$-optimal policy, is given by the defined
$
\mathrm{K}_{\mathrm{unif}}(\epsilon), 
$
in view of Corollary \ref{cor_non_sc_static_linear}.
The rest of the claim follows directly from the definition of $\mathrm{K}_{\mathrm{hybrid}}(\epsilon)$ and $\mathrm{K}_{\mathrm{unif}}(\epsilon)$.
\end{proof}

Theorem \ref{thrm_hybrid_sampling} states that with a hybrid sampling scheme satisfying $M \rho_{\cH}^{\#} = \cO(\Delta_0)$ and $\rho_{\cH}^\dagger \gg \rho_{\mathrm{unif}}^\dagger$, 
 the acceleration over uniform sampling occurs by a multiplicative factor during stage I, and the accumulated acceleration is maintained at stage II when we switch to the uniform policy.
Qualitatively, this requires that the initial policy can not be close to optimal (cf. Proposition \ref{prop_optimal_visitation_worse_than_uniform}), and the estimation quality $M$ inversely depends on the tail mass $\rho_{\cH}^{\#}$. For cases where $\rho_{\cH}^{\#}$ is small, one can allow a large $M$ and retain acceleration.

{\bf Applications.} We now provide in the following corollary concrete applications of discussions above.

\begin{corollary}[Acceleration with Polynomial Tails]\label{cor_accelerate_with_power_law}
Suppose $\rho$ satisfies $\abs{\cS_\rho} =  \abs{\cS}$, and $\rho_i \propto i^{-q}$ for some fixed $q > 1$,
where $\rho_i$ denotes the $i$-th largest value of $\rho$.
Let $\cH_n \subseteq \cS_\rho$ denote the subset with the $n$-largest values of $\rho(\cdot)$. 
 Let $n = \ceil{ \abs{\cS}^{1/(2q)}}$, and choose $\cH = \cH_n$.
 
If the initial policy satisfies $\Delta_0 = \Omega(\abs{\cS}^{-(q-1)/4})$,
and $\rho$ is an $M$-surrogate with $M \leq \abs{\cS}^{(q-1)/4}$, then
\begin{align*}
& \text{Stage I:} ~~~
 \mathrm{K}_{\mathrm{unif}}(\epsilon)  / \mathrm{K}_{\mathrm{hybrid}}(\epsilon) = \Omega(\abs{\cS}^{1/2} / (q-1)), ~ & \forall \epsilon = \Omega(\abs{\cS}^{-(q-1)/4}), \\
& \text{Stage II:} ~~~
 \mathrm{K}_{\mathrm{unif}}(\epsilon)  - \mathrm{K}_{\mathrm{hybrid}}(\epsilon) = \Omega \rbr{
 \tfrac{\abs{\cS}}{1-\gamma} (1 - \tfrac{q-1}{\abs{\cS}^{1/2}}) \log ( \tfrac{(1-\gamma)^2 \Delta_0 \abs{\cS}^{(q-1)/4}}{\overline{c} + \overline{h}}) }
 ,
~  & \forall \epsilon = \cO(\abs{\cS}^{-(q-1)/4}).
\end{align*}
\end{corollary}

\begin{proof}
Given the choice of $n$, it is immediate that $\rho_{\cH}^{\#} = \Theta( \abs{\cS}^{-(q-1)/(2q)})$, and $\rho_{\cH}^\dagger = \Theta(\abs{\cS}^{-1/2}/(q-1))$. 
The claim then follows directly from Theorem \ref{thrm_hybrid_sampling} by noting that $\epsilon_\tau = \cO(\abs{\cS}^{-(q-1)/4})$ for any $M \leq \abs{\cS}^{(q-1)/4}$. 
\end{proof}

Corollary \ref{cor_accelerate_with_power_law} states that if the sampling distribution is an $M$-surrogate with a light polynomial tail, 
the improvement factor of the iteration complexity by using the hybrid sampling over the uniform sampling, is at least  $\Omega(\abs{\cS}^{1/2} )$,  for finding any $\epsilon$-optimal policy with $\epsilon =  \Omega(\abs{\cS}^{-(q-1)/4})$.
Interestingly, this also states that in stage I, the iteration complexity of finding an $\epsilon$-optimal policy has only $\cO(\sqrt{S})$ dependence on the state space.
Note that this even holds if the $M$-surrogate $\rho$ is of low quality, where $M$ can depend polynomially 
on the state space.
The benefit of using a hybrid policy clearly gets more evident with large state spaces.

\vspace{0.05in}
{\bf Connection to Coordinate Descent Method.}
To conclude our development in this section, we briefly discuss some similarities and differences of the hybrid-sampling BPMD and the random coordinate descent method (RCDM) \cite{nesterov2012efficiency}.
 For minimizing a smooth convex function over $\RR^d$ with coordinate-wise smoothness parameters $\cbr{L_i}_{i=1}^d$, 
RCDM has a computational complexity of order $\cO(\tsum_{i=1}^d L_i / d)$,  which can be much smaller than that of gradient descent method with $\cO(\max_{i} L_i)$, if the coordinate-wise smoothness is imbalanced. 
This feature is clearly shared by BPMD, with the role of smoothness replaced by distribution $\nu^*$.
On the other hand, it should be noted that both the analyses and the mechanism for acceleration seem different.
In particular, at least two different traits can be identified: (1)  The acceleration of RCDM is global, while BPMD requires properly switching back to uniform sampling, whose necessity is discussed in Proposition \ref{prop_optimal_visitation_worse_than_uniform};
(2) RCDM also requires  coordinate-wise stepsize $\eta_i \propto 1/L_i$, while BPMD does not require state-wise stepsize. 
In fact, it is clear that using an infinite stepsize in the example constructed in Proposition \ref{prop_optimal_visitation_worse_than_uniform} still makes the $\nu^*$-sampling worse than the uniform sampling.

\vspace{0.05in}
{\bf On the Role of Weighted Objective.}
Our previous discussions focus on the case of $\nu = \nu^*$ in the weighted objective \eqref{eq:mdp_single_obj_raw} to simplify the analysis.
Here, we briefly note for general weights $\nu$, all prior results should hold, with the role of $\nu^*$ in the definition of $\nu_{\cH}^{\#}$ replaced by $d_{\nu}^{\pi^*}$,
defined as $d_{\nu}^{\pi^*}(\cdot) = \sum_{s} \nu(s) d^{\pi^*}_{s}(\cdot)$.
The contraction factor will be replaced by $ \rbr{1 - (1-\gamma) \rho^\dagger_{\cH} / \norm{d_\nu^{\pi^*} / \nu^* }_\infty}$, and this occurrence of distribution mismatch coefficient also appears in batch PG methods \cite{xiao2022convergence, agarwal2020optimality, li2022homotopic}. 
\section{Stochastic Block Policy Mirror Descent}\label{sec:stochastic}

The BPMD method presented in Section \ref{sec:bpmd} requires the exact value of Q-function at each iteration. 
For typical reinforcement learning problems,  the environment parameters $\cP$ are unknown in advance of learning.
Instead, one needs to use samples (also known as trajectories) collected through historical interactions with the environment to facilitate policy improvement.  
In this section, we propose and analyze the stochastic counterpart of the BPMD method, which only requires a stochastic estimation, instead of the exact value of Q-function at each iteration.

\begin{algorithm}[htb!]
    \caption{The stochastic block policy mirror descent (SBPMD) method}
    \label{alg:sbcpmd}
    \begin{algorithmic}
    \STATE{\textbf{Input:} Initial policy $\pi_0$,  stepsizes $\{\eta_k\}$ and sampling distribution $\rho \in \Delta_{\cS}$.}
    \FOR{$k=0, 1, \ldots$}
	\STATE{ Sample $s_k \sim \rho$.}
	\STATE{Update policy:
	\vspace{-0.15in}
	\begin{align*}
	\pi_{k+1} (\cdot| s) = \begin{cases}
	\argmin_{p(\cdot|s_k)  \in \Delta_{\cA}} \eta_k\sbr{ \inner{Q^{\pi_k, \xi_k,s_k}(s_k, \cdot)}{ p(\cdot| s_k ) } + h^p(s_k) } + D^{p}_{\pi_k}(s_k), ~~~ & s = s_k, \\
	\pi_k(\cdot| s) , ~~~ & s \neq s_k.
	\end{cases}
	\end{align*}
	}
    \ENDFOR
    \end{algorithmic}
\end{algorithm}


The stochastic block policy mirror descent (SBPMD) method shares the similar update with the BPMD method, except that it uses a stochastic estimate $Q^{\pi_k, \xi_k, s_k}$ to replace the exact Q-function in the update,
where $\xi_k$ denotes the random vector used to construct the stochastic estimate.
The construction of  $Q^{\pi_k, \xi_k,s_k}$  is allowed to depend on the sampled state $s_k$.
We make the following assumption on the stochastic estimation. Detailed construction of the  estimation satisfying this assumption will be discussed in Section \ref{subsec_sample}.

\begin{assumption}\label{assump:noise}
We have access to a stochastic estimation $Q^{\pi_k, \xi_k,s_k} \in \RR^{\abs{\cS} \times \abs{\cA}}$ satisfying
\begin{align}
\EE_{\xi_k |s_k} \big[ Q^{\pi_k, \xi_k,s_k}(s_k, \cdot) \big]    = \overline{Q}^{\pi_k,s_k}( s_k,\cdot),   \nonumber
\\
\EE_{\xi_k | s_k} \big[\norm{ Q^{\pi_k, \xi_k,s_k}(s_k, \cdot)}_\infty^2 \big] \leq M_k^2, \label{assump:bounded_norm} \\
 \norm{\overline{Q}^{\pi_k,s_k}(s_k, \cdot) - Q^{\pi_k}(s_k, \cdot) }_\infty  \leq \varepsilon_k . \label{assump:bias}
\end{align}
\end{assumption}

To proceed, the following lemma generalizes Lemma \ref{lemma:three_point}, whose proof follows the same arguments.

\begin{lemma}\label{lemma:sbcpmd_three_point}
For any $p(\cdot|s_k) \in \Delta_{\abs{\cA}}$, the pair of policies $(\pi_k, \pi_{k+1})$ generated by SBPMD satisfy
\begin{align}\label{sbcpmd:three_point}
\eta_k \big[ \inner{Q^{\pi_k, \xi_k, s_k}(s_k, \cdot)}{ \pi_{k+1} (\cdot| s_k )- p (\cdot| s_k)} & + h^{\pi_{k+1}}(s_k)  - h^p(s_k) \big]  + D^{\pi_{k+1}}_{\pi_k} (s_k) 
  \\ 
 & \leq D^p_{\pi_k} (s_k) - (1+ \eta_k \mu) D^p_{\pi_{k+1}} (s_k) . \nonumber
\end{align}
\end{lemma}

We then proceed to establish the basic convergence properties of SBPMD.
It is worth mentioning  that we will require $h$ to have bounded subgradients.
This can be satisfied, for example, when $h \equiv 0$  or $h^{\pi}(s) = \norm{\pi(\cdot|s)}_p^p$ for $p \geq 1$.

\begin{lemma}\label{lemma_stoch_generic_recursion}
Let $\rho$ be the static sampling policy,
and let $\cS_\rho$ denote the support of $\rho$. Let $\cS^*$ denote the support of $\nu^*$.
Suppose $0 \leq c(s,a) \leq \overline{c}$ and $0 \leq h^{\pi}(s) \leq \overline{h}$, for any $(s,a)$ and $\pi \in \Pi$.
Furthermore, suppose $h^{\pi}$ has bounded subgradients, i.e., $\norm{\partial h^{\pi}(s)}_\infty \leq \ell_h$ holds for any $\pi \in \Pi$  for some $\ell_h \geq 0$. 

For any $\cH \subseteq \cS_\rho$, define $\nu^{\#}_\cH = \tsum_{s \in \cS^* / \cH} \nu^*(s)$, and $\rho^\dagger_{\cH} = \min_{s \in \cH \cap \cS^*} \rho(s)$.
Then the policy pair $(\pi_k, \pi_{k+1})$ generated by SBPMD satisfies 
\begin{align}
& (1-\gamma)\sbr{ f(\pi_k) - f(\pi^*)}  \label{stochastic_adaptive_convergence_recursion} \\
\leq &
 \sbr{\tfrac{1}{\eta_k} + \mu (1- \rho^\dagger_{\cH}) }    \tsum_{s \in \cH}  \tfrac{\nu^*(s)}{\rho(s)}  D^{\pi^*}_{\pi_k}(s)  -  \sbr{\tfrac{1}{\eta_k} + \mu }    \EE_{k} \sbr{   \tsum_{s \in \cH}  \tfrac{\nu^*(s)}{\rho(s)} D^{\pi^*}_{\pi_{k+1}}(s)}  \nonumber \\
& ~ +  \tfrac{2  \nu^{\#}_{\cH}  \rho^\dagger_{\cH} ( \overline{c} + \overline{h})}{1-\gamma}
+ \eta_k (M_k^2 + \ell_h^2)  + 2\varepsilon_k. \nonumber
\end{align}
\end{lemma}

\begin{proof}
By taking $p(\cdot|s_k) = \pi^*(\cdot|s_k)$ in \eqref{sbcpmd:three_point},   we have 
\begin{align*}
& \eta_k \big[ \inner{Q^{\pi_k, \xi_k, s_k}(s_k, \cdot)}{ \pi_{k+1}(\cdot| s_k )- \pi^* (\cdot| s_k)}
+  h^{\pi_{k+1}}(s_k)  - h^{\pi^*}(s_k) \big]  + D^{\pi_{k+1}}_{\pi_k}(s_k) \\
\leq 
& D^{\pi^*}_{\pi_k}(s_k)  - (1 + \eta_k \mu)D^{\pi^*}_{\pi_{k+1}}(s_k) .
\end{align*}
Letting $\delta_k= Q^{\pi_k, \xi_k, s_k} - Q^{\pi_k}$, we obtain from the above relation that 
\begin{align*}
& \eta_k \big[ \inner{Q^{\pi_k}(s_k, \cdot)}{ \pi_{k}(\cdot| s_k )- \pi^* (\cdot| s_k)}
+  h^{\pi_{k}}(s_k)  - h^{\pi^*}(s_k) \big]   \\
\leq 
& D^{\pi^*}_{\pi_k}(s_k)  - (1 + \eta_k \mu)D^{\pi^*}_{\pi_{k+1}}(s_k) 
-  D^{\pi_{k+1}}_{\pi_k}(s_k) \\
& ~ + \eta_k \sbr{
 \inner{Q^{\pi_k, \xi_k, s_k}(s_k, \cdot)}{ \pi_{k}(\cdot| s_k )- \pi_{k+1} (\cdot| s_k)}
+  h^{\pi_{k}}(s_k)  - h^{\pi_{k+1}}(s_k) 
} \\
&~ - \eta_k \inner{\delta_k(s_k, \cdot)}{\pi_{k}(\cdot| s_k )- \pi^* (\cdot| s_k)} \\
 \overset{(a)}{\leq} & 
D^{\pi^*}_{\pi_k}(s_k)  - (1 + \eta_k \mu)D^{\pi^*}_{\pi_{k+1}}(s_k) -  D^{\pi_{k+1}}_{\pi_k}(s_k)  +  \eta_k \inner{\delta_k(s_k, \cdot)}{\pi_{k}(\cdot| s_k )- \pi^* (\cdot| s_k)} \\
& ~ + \eta_k \rbr{ \norm{Q^{\pi_k, \xi_k, s_k}(s_k, \cdot)}_\infty + \ell_h} 
\norm{\pi_{k}(\cdot| s_k )- \pi_{k+1} (\cdot| s_k)}_1 \\ 
\overset{(b)}{\leq}& 
D^{\pi^*}_{\pi_k}(s_k)  - (1 + \eta_k \mu)D^{\pi^*}_{\pi_{k+1}}(s_k) 
 +  \eta_k \inner{\delta_k (s_k, \cdot)}{\pi_{k}(\cdot| s_k )- \pi^* (\cdot| s_k)}   \\
& ~ 
+ \eta_k^2 \rbr{\norm{Q^{\pi_k, \xi_k, s_k}(s_k, \cdot)}_\infty^2 + \ell_h^2},
\end{align*} 
where $(a)$ follows from $h^{\pi_{k}}(s_k)  - h^{\pi_{k+1}}(s_k) \leq \inner{\partial h^{\pi_k}(s_k)}{\pi_{k}(\cdot|s_k) - \pi_{k+1}(\cdot|s_k)} \leq \norm{\partial h^{\pi_k}(s_k)}_\infty \norm{\pi_{k}(\cdot| s_k )- \pi_{k+1} (\cdot| s_k)}_1$ and $\norm{\partial h^{\pi_k}(s_k)}_\infty \leq \ell_h$;
$(b)$ follows from $D^{\pi_{k+1}}_{\pi_k}(s_k)  \geq \tfrac{1}{2}  \norm{\pi_{k}(\cdot| s_k )- \pi_{k+1} (\cdot| s_k)}_1^2$ and the Fenchel-Young inequality.
 Multiplying both sides of inequality $(b)$ in the above relation with $\nu^*(s_k) / \rho(s_k)$ gives 
\begin{align*}
& \eta_k \tfrac{\nu^*(s_k)}{\rho(s_k)} \sbr{ \inner{Q^{\pi_k}(s_k, \cdot)}{ \pi_{k}(\cdot| s_k )- \pi^* (\cdot| s_k)}  +  h^{\pi_{k+1}}(s_k)  - h^{\pi^*}(s_k)}\\
\leq 
& 
 \tfrac{\nu^*(s_k)}{\rho(s_k)} D^{\pi^*}_{\pi_k}(s_k)  - (1 + \eta_k \mu)  \tfrac{\nu^*(s_k)}{\rho(s_k)} D^{\pi^*}_{\pi_{k+1}}(s_k)  \\
& ~ +  \tfrac{\nu^*(s_k)}{\rho(s_k)} \sbr{ \eta_k^2 \rbr{\norm{Q^{\pi_k, \xi_k, s_k}(s_k, \cdot)}_\infty^2 + \ell_h^2} 
 +  \eta_k \inner{\delta_k(s_k, \cdot)}{\pi_{k}(\cdot| s_k )- \pi^* (\cdot| s_k)} },
\end{align*}
By comparing the above relation and \eqref{adaptive_starting_eq}, it is clear that the only difference is the replacement of term $\eta_k \sbr{V^{\pi_{k+1}}(s_k) - V^{\pi_{k}}(s_k)} $ in \eqref{adaptive_starting_eq}  by  the trailing term above. 
Thus following similar arguments as in the proof of Lemma \ref{lemma_adaptive_convergence_general}, 
and letting $\EE_k[\cdot]$ denote taking expectation over $s_k$ and $\xi_k$ conditioned on $\pi_{k}$, 
we  obtain 
\begin{align*}
&  \eta_k \sbr{ (1-\gamma)\sbr{ f(\pi_k) - f(\pi^*)} - \tfrac{ 2 \nu^{\#}_{\cH}  ( \overline{c} + \overline{h})}{1-\gamma}}  \nonumber
\\  
  \leq   &
  \tsum_{s \in  \cH }  \tfrac{\nu^*(s)}{\rho(s)}    D^{\pi^*}_{\pi_k}(s)  - (1+ \eta_k \mu)    \tsum_{s\in \cH}  \tfrac{\nu^*(s)}{\rho(s)}    \EE_{k} \big[ D^{\pi^*}_{\pi_{k+1}}(s) \big]+ \eta_k \mu   (1 - \rho^\dagger_{\cH}) \tsum_{s \in \cH}  \tfrac{\nu^*(s)}{\rho(s)} D^{\pi^*}_{\pi_k}(s)   \nonumber \\
   &
  ~ + \EE_{k} \sbr{ \tfrac{\nu^*(s_k)}{\rho(s_k)} \sbr{ \eta_k^2 \rbr{\norm{Q^{\pi_k, \xi_k, s_k}(s_k, \cdot)}_\infty^2 + \ell_h^2} 
 + \eta_k \inner{\delta_k}{\pi_{k}(\cdot| s_k )- \pi^* (\cdot| s_k)} } } \\
  \leq   &
  \tsum_{s \in  \cH }  \tfrac{\nu^*(s)}{\rho(s)}    D^{\pi^*}_{\pi_k}(s)  - (1+ \eta_k \mu)    \tsum_{s\in \cH}  \tfrac{\nu^*(s)}{\rho(s)}    \EE_{k} \big[ D^{\pi^*}_{\pi_{k+1}}(s) \big]+ \eta_k \mu   (1 - \rho^\dagger_{\cH}) \tsum_{s \in \cH}  \tfrac{\nu^*(s)}{\rho(s)} D^{\pi^*}_{\pi_k}(s)   \nonumber \\
   &
  ~ + \eta_k^2 \tsum_{s \in \cH} \nu^*(s) \EE_{\xi_k|s_k=s} \rbr{ \norm{ Q^{\pi_k, \xi_k,s_k}(s_k, \cdot)}_\infty^2 + \ell_h}^2
  +2  \eta_k \tsum_{s\in \cH} \nu^*(s) \norm{\EE_{\xi_k|s_k = s} \delta_k(s_k, \cdot) }_\infty \\
  \leq &
  \tsum_{s \in  \cH }  \tfrac{\nu^*(s)}{\rho(s)}    D^{\pi^*}_{\pi_k}(s)  - (1+ \eta_k \mu)    \tsum_{s\in \cH}  \tfrac{\nu^*(s)}{\rho(s)}    \EE_{k} \big[ D^{\pi^*}_{\pi_{k+1}}(s) \big]+ \eta_k \mu   (1 - \rho^\dagger_{\cH}) \tsum_{s \in \cH}  \tfrac{\nu^*(s)}{\rho(s)} D^{\pi^*}_{\pi_k}(s)   \nonumber \\
 &~ + \eta_k^2 ( M_k^2 + \ell_h^2) + 2 \eta_k \varepsilon_k.
\end{align*}  
The desired claim follows after simple rearrangement. 
\end{proof}

\vspace{-0.05in}

\subsection{Convergence with Strongly-convex Regularizers}
For strongly-convex regularizers, we show that SBPMD converges at the rate of ${\cO}(\log k/k)$ until the bias level in the stochastic estimate is reached.

\begin{theorem}\label{thrm_stoch_sc}
Suppose $h$ satisfies $\mu > 0$, and $M_k \leq M$, $\varepsilon_k \leq \varepsilon$ for all $k \geq 0$.
For any $\cH \subseteq \cS_\rho$,  by choosing stepsizes $\eta_k = 1/ [\mu \rho_{\cH}^\dagger (k+1)]$, 
and letting $\cR \sim \mathrm{Unif}(\cbr{0, \cdots, k-1})$,
then
\begin{align*}
&  \EE \sbr{ f(\pi_{\cR}) - f(\pi^*)} 
 \leq 
 \tfrac{[1 + \mu^2 (1-\rho_{\cH}^\dagger) \rho_{\cH}^\dagger] \log \abs{\cA}}{\mu (1-\gamma) \rho_{\cH}^\dagger k}
+   \tfrac{ (M^2 + \ell_h^2) \log k}{ \mu (1-\gamma) \rho_{\cH}^\dagger k}
+ \tfrac{ 2 \nu^{\#}_{\cH}  \rho^\dagger_{\cH} ( \overline{c} + \overline{h})}{(1-\gamma)^2} 
 +  \tfrac{2  \varepsilon}{1-\gamma}.
\end{align*}
\end{theorem}

\begin{proof}
In view of \eqref{stochastic_adaptive_convergence_recursion}, let  
$\phi_{\cH} (\pi^*, \pi_k) = \tsum_{s \in \cH}  \nu^*(s)  D^{\pi^*}_{\pi_k}(s) / \rho(s)$, and multiplying both sides by some nonnegative $\lambda_k$, we obtain
\begin{align}
& (1-\gamma) \lambda_k \sbr{ f(\pi_k) - f(\pi^*)} \label{stoch_sc_telescope_recursion} \\
\leq &
\lambda_k \sbr{\tfrac{1}{\eta_k} + \mu (1- \rho^\dagger_{\cH}) }    \phi_{\cH}(\pi^*, \pi_k)  -   \lambda_k \sbr{\tfrac{1}{\eta_k} + \mu }    \EE_{k} \sbr{ \phi_{\cH} (\pi^*, \pi_{k+1}) }  \nonumber \\
& ~ +  \tfrac{2 \lambda_k  \nu^{\#}_{\cH}  \rho^\dagger_{\cH} ( \overline{c} + \overline{h})}{1-\gamma}
+ \eta_k \lambda_k (M^2 + \ell_h^2)  + 2 \lambda_k \varepsilon. \nonumber
\end{align}

 Suppose $\cbr{(\lambda_k, \eta_k)}$ satisfy 
\begin{align}\label{stoch_sc_telescope_stepsize}
\lambda_{k+1}  \sbr{\tfrac{1}{\eta_{k+1}} + \mu (1- \rho^\dagger_{\cH}) }
\leq \lambda_k \sbr{\tfrac{1}{\eta_k} + \mu } .
\end{align}
Then by taking total expectation, we can take telescoping sum of \eqref{stoch_sc_telescope_recursion} from $t= 0$ to $k-1$, and obtain 
\begin{align*}
& (1-\gamma) \tsum_{t=0}^{k-1} \lambda_t \EE \sbr{ f(\pi_t) - f(\pi^*)} \\
\leq & 
 \lambda_0 \sbr{\tfrac{1}{\eta_0} + \mu (1- \rho^\dagger_{\cH}) }    \phi_{\cH}(\pi^*, \pi_0) 
 + \tsum_{t=0}^{k-1} \tfrac{2 \lambda_k \nu^{\#}_{\cH}  \rho^\dagger_{\cH} ( \overline{c} + \overline{h})}{1-\gamma}
 +  \tsum_{t=0}^{k-1} \eta_t \lambda_t (M^2 + \ell_h^2) 
 + 2 \tsum_{t=0}^{k-1} \lambda_t \varepsilon.
\end{align*}
One can readily verify that $\eta_k = [\mu \rho_{\cH}^\dagger (k+1)]^{-1}$ and $\lambda_k = \lambda > 0$ satisfy condition \eqref{stoch_sc_telescope_stepsize}.
Given the definition of $\cR$, the above relation then implies 
\begin{align*}
& (1-\gamma) \EE \sbr{ f(\pi_{\cR}) - f(\pi^*)} \\ 
\leq & 
  \tfrac{ 1/[\mu \rho_{\cH}^\dagger] + \mu (1- \rho^\dagger_{\cH}) }{k}    \phi_{\cH}(\pi^*, \pi_0) 
 +   (M^2 + \ell_h^2) \tfrac{\log k}{k \mu \rho_{\cH}^\dagger}
 +  \tfrac{2 \nu^{\#}_{\cH}  \rho^\dagger_{\cH} ( \overline{c} + \overline{h})}{1-\gamma}   + 2  \varepsilon  \\
 \leq &
 \tfrac{[1 + \mu^2 (1-\rho_{\cH}^\dagger) \rho_{\cH}^\dagger] \log \abs{\cA}}{\mu \rho_{\cH}^\dagger k}
+   \tfrac{\log k (M^2 + \ell_h^2)}{k \mu \rho_{\cH}^\dagger}
+ \tfrac{2 \nu^{\#}_{\cH}  \rho^\dagger_{\cH} ( \overline{c} + \overline{h})}{1-\gamma} 
 + 2  \varepsilon.
\end{align*}
The desired claim follows after simple rearrangement.
\end{proof}

Similar to our discussions in Section \ref{subsec_acc_by_adaptivity}, the adaptive convergence bounds in Theorem \ref{thrm_stoch_sc} allow one to adopt a hybrid sampling scheme, and facilitate potential acceleration over the uniform sampling. 
Meanwhile, we can establish the convergence with any static and exploratory sampling distribution, in particular, the uniform sampling.

\begin{corollary}\label{cor_stoch_sc_uniform_sampling}
Under the same setup as in Theorem \ref{thrm_stoch_sc}.
 Let $\eta_k =1/ [\rho^\dagger \mu (k+1)]$ where $\rho$ is exploratory, 
\begin{align*}
&  \EE \sbr{ f(\pi_{\cR}) - f(\pi^*)} 
 \leq 
 \tfrac{[ 1/\rho^\dagger  + \mu^2] \log \abs{\cA}}{\mu (1-\gamma)  k}
+   \tfrac{  (M^2 + \ell_h^2) \log k}{ \mu \rho^\dagger (1-\gamma)  k}
 +  \tfrac{2  \varepsilon}{1-\gamma}.
\end{align*}
In particular, for $\rho = \mathrm{Unif}(\cS)$, 
let stepsize $\eta_k = \abs{\cS} / [ \mu (k+1)]$, then 
\begin{align*}
&  \EE \sbr{ f(\pi_{\cR}) - f(\pi^*)} 
 \leq 
 \tfrac{[\abs{\cS}  + \mu^2] \log \abs{\cA}}{\mu (1-\gamma)  k}
+   \tfrac{  \abs{\cS} (M^2 + \ell_h^2) \log k}{ \mu (1-\gamma)  k}
 +  \tfrac{2  \varepsilon}{1-\gamma}.
\end{align*}
\end{corollary}

\begin{proof}
The claim follows by taking $\cH = \cS^*$ in Theorem \ref{thrm_stoch_sc}, and using $\rho$ being exploratory and $\nu_{\cH}^{\#} = 0$. 
\end{proof}

\subsection{Convergence with Non-strongly-convex Regularizers}
For non-strongly-convex regularizers, we show that with constant stepsizes, SBPMD converges at the rate of $\cO (\ssqrt{1/k})$ until the bias level in the stochastic estimate is reached.

\begin{theorem}\label{thrm_stoch_non_sc}
Suppose $h$ satisfies $\mu = 0$, and $M_k \leq M$, $\varepsilon_k \leq \varepsilon$ for all $k \geq 0$.
For any $\cH \subseteq \cS_\rho$, by fixing the total number of iteration $k$, and choosing constant stepsize  $\eta = \ssqrt{  \log \abs{\cA}  / [ k (M^2 + \ell_h^2) \rho_{\cH}^\dagger  ] }$, then
\begin{align}\label{eq_stoch_non_sc}
\EE \sbr{ f(\pi_{\cR}) - f(\pi^*)} 
\leq 
\ssqrt{ \tfrac{(M^2 + \ell_h^2)  \log \abs{\cA} }{ k \rho_{\cH}^\dagger (1-\gamma)^2}}
+ 
 \tfrac{ 2 \nu^{\#}_{\cH}   ( \overline{c} + \overline{h})}{(1-\gamma)^2}
  +   \tfrac{2 \varepsilon}{1-\gamma}.
\end{align}
where $\cR \sim \mathrm{Unif}(\cbr{0, \cdots, k-1})$.
\end{theorem}

\begin{proof}
In view of \eqref{stochastic_adaptive_convergence_recursion}, 
by choosing constant stepsize $\eta$, 
we have 
\begin{align*}
 (1-\gamma)\sbr{ f(\pi_k) - f(\pi^*)}   
& \leq 
 \tfrac{1}{\eta}   \tsum_{s \in \cH}  \tfrac{\nu^*(s)}{\rho(s)}  D^{\pi^*}_{\pi_k}(s)  - \tfrac{1}{\eta}    \EE_{k} \sbr{   \tsum_{s \in \cH}  \tfrac{\nu^*(s)}{\rho(s)} D^{\pi^*}_{\pi_{k+1}}(s)}  \nonumber \\
& ~~~~~ +  \tfrac{ 2\nu^{\#}_{\cH}  \rho^\dagger_{\cH} ( \overline{c} + \overline{h})}{1-\gamma}
+ \eta (M^2 + \ell_h^2)  + 2 \varepsilon. \nonumber
\end{align*}
Summing above inequality from $t = 0$ to $k-1$, 
and taking total expectation, 
we obtain 
\begin{align*}
(1-\gamma) \tsum_{t=0}^{k-1}\EE \sbr{ f(\pi_t) - f(\pi^*)} 
& \leq \tfrac{1}{\eta}   \tsum_{s \in \cH}  \tfrac{\nu^*(s)}{\rho(s)}  D^{\pi^*}_{\pi_0}(s) 
 +  \tfrac{2 k \nu^{\#}_{\cH}  \rho^\dagger_{\cH} ( \overline{c} + \overline{h})}{1-\gamma}
+ \eta k (M^2 + \ell_h^2)  +  2 k\varepsilon .
\end{align*}
Given the definition of $\cR$, we obtain from the above relation that 
\begin{align*}
(1-\gamma) \EE \sbr{ f(\pi_{\cR}) - f(\pi^*)} 
\leq 
\tfrac{1}{\eta k \rho_{\cH}^\dagger}  \log \abs{\cA}
+ \eta  (M^2 + \ell_h^2)
+ 
 \tfrac{2 \nu^{\#}_{\cH}  \rho^\dagger_{\cH} ( \overline{c} + \overline{h})}{1-\gamma}
  +  2\varepsilon.
\end{align*}
The desired claim follows by taking $\eta = \ssqrt{ \log \abs{\cA}  / [ k (M^2 + \ell_h^2) \rho_{\cH}^\dagger  ]}$.
\end{proof}

By specializing Theorem \ref{thrm_stoch_non_sc} for $\cH = \cS^*$, we can now obtain the convergence of SBPMD with exploratory and static sampling.

\begin{corollary}\label{cor_stoch_nonsc_uniform_sampling}
Under the same setup as in Theorem \ref{thrm_stoch_non_sc}.
 Let $\eta = \ssqrt{  \log \abs{\cA}  / [ k (M^2 + \ell_h^2) \rho^\dagger ]}$ where $\rho$ is exploratory, then
\begin{align*}
\EE \sbr{ f(\pi_{\cR}) - f(\pi^*)} 
\leq 
\ssqrt{ \tfrac{(M^2 + \ell_h^2)   \log \abs{\cA}  }{ k \rho^\dagger (1-\gamma)^2}}
  +   \tfrac{2 \varepsilon}{1-\gamma}.
\end{align*}
In particular, for $\rho = \mathrm{Unif}(\cS)$, letting $\eta = \ssqrt{ \abs{\cS} \log \abs{\cA}  / [ k (M^2 + \ell_h^2)  ]}$ gives
\begin{align*}
\EE \sbr{ f(\pi_{\cR}) - f(\pi^*)} 
\leq 
\ssqrt{ \tfrac{(M^2 + \ell_h^2) \abs{\cS}  \log \abs{\cA} }{ k  (1-\gamma)^2}}
  +   \tfrac{2 \varepsilon}{1-\gamma}.
\end{align*}
\end{corollary}
\begin{proof}
The proof follows the same lines as Corollary \ref{cor_stoch_sc_uniform_sampling}.
\end{proof}

In view of Corollary \ref{cor_stoch_sc_uniform_sampling}, \ref{cor_stoch_nonsc_uniform_sampling}, 
the iteration complexities of SBPMD are bounded by $\cO(\abs{\cS} \log (1/\epsilon) / (\mu \epsilon))$ (resp. $\cO(\abs{\cS} / \epsilon^2)$) for strongly-convex (resp. non-strongly-convex) regularizers.
Similar to what we have seen in Section \ref{subsubsec_uniform_sampling_bpmd} for the deterministic setting, these iteration complexities naturally match that of stochastic batch PG methods (e.g., stochastic PMD methods \cite{lan2021policy}), up to a factor of $\abs{\cS}$.

\subsection{Sample Complexity of SBPMD}\label{subsec_sample}

Our previous discussions of SBPMD require Assumption \ref{assump:noise} to hold for the stochastic estimate of the Q-function. 
We now consider the a specific construction that satisfies this condition, and consequently determine the total sample complexity of the SBPMD method for both strongly-convex (SC)  and non-strongly-convex (NSC)  regularizers.

 {\it The method of independent trajectories} assumes  a generative model with the following capability.
At a given iteration $k$,
for any state-action pair $(s, a) \in \cS \times \cA$, we can generate a trajectory of length $T_k$, by following policy $\pi_k$ starting from $(s,a)$. The trajectory, denoted by $\zeta_k (s,a)$, takes the form of
$
\zeta_k(s,a) = \{ (s_0 = s, a_0 = a), (s_1, a_1), \ldots, (s_{T_k - 1}, a_{T_k - 1}) \},
$
and the trajectories are independent  for different $(s,a)$ pairs.

Let $\xi_k = \{ \zeta_k(s_k, a),  a \in \cA \}$ be the set of random variables for constructing the stochastic estimate. Then the stochastic estimate $Q^{\pi_k, \xi_k,s_k}$ is defined by 
\begin{align*}
Q^{\pi_k, \xi_k,s_k} (s, a ) = \begin{cases}
 \tsum_{t=0}^{T_k - 1} \gamma^t \sbr{c(s_t,  a_t) + h^{\pi_k} (s_t)  }, &   s= s_k,   a \in \cA ,\\
  0, & \forall s \neq s_k, a \in \cA.
\end{cases}
\end{align*}
One can readily verify that $Q^{\pi_k, \xi_k,s_k}$ satisfies Assumption \ref{assump:noise} with 
\begin{align}\label{method_it_noise_prop}
M_k  \leq  \tfrac{\overline{c} + \overline{h}}{1-\gamma}, ~~
\varepsilon_k \leq  \tfrac{(\overline{c} + \overline{h} )\gamma^{T_k} }{1-\gamma}.
\end{align}

For a clear comparison with the batch stochastic PG methods, we will focus on the uniform sampling scheme. 
Sample complexities for general static sampling schemes can be similarly derived.
We first establish the sample complexities for SC regularizers. 

\begin{theorem}[Sample Complexity, $\mu > 0$]
For any $\epsilon > 0$, 
set $T_k = T = \tfrac{1}{1-\gamma}  \log (\tfrac{6 (\overline{c} + \overline{h})}{\epsilon(1-\gamma)})$ in the method of independent trajectories.
 Then with $\eta_k = \abs{\cS} / [ \mu (k+1)]$,  
 SBPMD with uniform sampling finds a policy $\pi_{\cR}$ satisfying $ \EE \sbr{ f(\pi_{\cR}) - f(\pi^*)} 
 \leq \epsilon$,
 where $\cR \sim \mathrm{Unif}(\cbr{0, \cdots, k-1})$,
  in 
 \begin{align*}
 k =  \tfrac{3 (\abs{\cS} + \mu^2) \log \abs{\cA}}{\mu (1-\gamma) \epsilon} 
 + \tfrac{2 \abs{\cS} [ (\overline{c} + \overline{h})^2  + \ell_h^2]}{(1-\gamma)^3 \mu \epsilon}
 \log \rbr{\tfrac{2 \abs{\cS} [ (\overline{c} + \overline{h})^2  + \ell_h^2]}{(1-\gamma)^3 \mu \epsilon} }
 \end{align*}
 iterations. 
 In addition, the total number of samples required is bounded by 
 \begin{align*}
 \cO \rbr{
 \tfrac{\abs{\cS} \log \abs{\cA} [(\overline{c} + \overline{h})^2 + \ell_h^2]}{\mu (1-\gamma)^4 \epsilon} \log^2 (\tfrac{1}{\epsilon})
 }.
 \end{align*}
\end{theorem}

\begin{proof}
Given the choice of $T_k$, it is clear from \eqref{method_it_noise_prop} that 
$\varepsilon_k \leq \varepsilon, M_k \leq M$, with
 $ \tfrac{2  \varepsilon}{1-\gamma} \leq \tfrac{\epsilon}{3}$,
and  $M = \tfrac{\overline{c} + \overline{h}}{1-\gamma}$.
In addition, by choosing the specified $k$,
direct calculation yields  
\begin{align*}
 \tfrac{[\abs{\cS}  + \mu^2] \log \abs{\cA}}{\mu (1-\gamma)  k}
+   \tfrac{  \abs{\cS} (M^2 + \ell_h^2) \log k}{ \mu (1-\gamma)  k}
 \leq \tfrac{2\epsilon}{3}.
\end{align*}
Hence in view of Corollary \ref{cor_stoch_sc_uniform_sampling}, we have  $ \EE \sbr{ f(\pi_{\cR}) - f(\pi^*)} 
 \leq \epsilon$.
The sample complexity is bounded by $k T$.
\end{proof}

Similarly, sample complexities for NSC regularizers are established below.

\begin{theorem}[Sample Complexity, $\mu = 0$]\label{thrm_sample_sbpmd_nonsc}
For any $\epsilon > 0$, fix 
$
 k = \tfrac{4 [(\overline{c} + \overline{h})^2 + \ell_h^2] \abs{\cS} \log \abs{\cA}}{(1-\gamma)^4 \epsilon^2}, 
$
and set $T_s = T = \tfrac{1}{1-\gamma}  \log (\tfrac{4 (\overline{c} + \overline{h})}{\epsilon(1-\gamma)})$ in the method of independent trajectories for all $s \geq 0$.
 Then with $ \eta =  \ssqrt{\abs{\cS} \log \abs{\cA}  / [ k [ (\overline{c} + \overline{h})^2 / (1-\gamma)^2 + \ell_h^2]  ]}$,  
 SBPMD finds a policy $\pi_{\cR}$ satisfying $ \EE \sbr{ f(\pi_{\cR}) - f(\pi^*)} \leq \epsilon$ in $k$
 iterations. 
 In addition, the total number of samples required is bounded by 
 \begin{align*}
 \cO \rbr{
 \tfrac{ [(\overline{c} + \overline{h})^2 + \ell_h^2] \abs{\cS} \log \abs{\cA}}{(1-\gamma)^5 \epsilon^2} \log ( \tfrac{1}{\epsilon})
 }.
 \end{align*}
\end{theorem}

\begin{proof}
Given the choice of $T_k$, it is clear from \eqref{method_it_noise_prop} that 
$\varepsilon_k \leq \varepsilon, M_k \leq M$, with
 $ \tfrac{2  \varepsilon}{1-\gamma} \leq \tfrac{\epsilon}{2}$,
and  $M = \tfrac{\overline{c} + \overline{h}}{1-\gamma}$.
By choosing the specified $k$, direct calculation yields 
$
\ssqrt{ \tfrac{(M^2 + \ell_h^2) \abs{\cS}  \log \abs{\cA} }{ k  (1-\gamma)^2}} \leq \tfrac{\epsilon}{2}.
$
Hence in view of Corollary \ref{cor_stoch_nonsc_uniform_sampling}, we have  $ \EE \sbr{ f(\pi_{\cR}) - f(\pi^*)} 
 \leq \epsilon$.
The sample complexity is bounded by $k T$.
\end{proof}

We now summarize the obtained iteration and sample complexities of SBPMD with uniform sampling, and compare with that of the SPMD \cite{lan2021policy}, which attains the best sample complexities among stochastic batch PG methods.
The results are summarized in Table \ref{tab_iter_sample_complexity_comparison}.
The total sample complexities of SBPMD match that of SPMD for both SC and NSC regularizers up to a potentially logarithmic factor. 
Meanwhile, the samples required at each iteration for SBPMD are $\abs{\cS}$-times less than that of SPMD. 
Since sampling takes up majority of the computation for both methods,
the per-iteration computation for SBPMD is thus $\abs{\cS}$-times smaller than that of SPMD.
In light of this observation, it should be clear that SBPMD with the uniform sampling shares similar benefits over stochastic batch PG methods as we have highlighted in Remark \ref{remark_benefits_uniform}.

\begin{table}[tb!]
\centering
\begin{tabular}{ ||c|c|c|c|c|| } 
\hline
Regularizers & Method & Samples per Iter. & Iter. Complexity & Sample Complexity  \\
\hhline{||=|=|=|=|=||}
\multirow{2}{*}{$\mu > 0$} & SPMD & $ \tfrac{\abs{\cS} \abs{\cA}}{1-\gamma} \log(\tfrac{1}{\epsilon})$ & $\tfrac{ \log \abs{\cA}}{\mu (1-\gamma)^3 \epsilon} $ & 
$\tfrac{\abs{\cS} \abs{\cA} \log \abs{\cA}}{\mu (1-\gamma)^4 \epsilon} \log (\tfrac{1}{\epsilon})$  \\
\cline{2-5}
& SBPMD  & $ \tfrac{\abs{\cA}}{1-\gamma} \log(\tfrac{1}{\epsilon})$ & $\tfrac{\abs{\cS} \log \abs{\cA}}{\mu (1-\gamma)^3 \epsilon} \log (\tfrac{1}{\epsilon})$ & $\tfrac{\abs{\cS} \abs{\cA} \log \abs{\cA}}{\mu (1-\gamma)^4 \epsilon} \log^2 (\tfrac{1}{\epsilon})$ \\
\hhline{||=|=|=|=|=||}
\multirow{2}{*}{$\mu = 0$} & SPMD & $ \tfrac{\abs{\cS} \abs{\cA}}{1-\gamma} \log(\tfrac{1}{\epsilon})$ &  $\tfrac{ \log \abs{\cA}}{(1-\gamma)^4 \epsilon^2} $ &
$\tfrac{\abs{\cS} \abs{\cA} \log \abs{\cA}}{(1-\gamma)^5 \epsilon^2} \log(\tfrac{1}{\epsilon}) $  \\ 
\cline{2-5}
& SBPMD  & $ \tfrac{\abs{\cA}}{1-\gamma} \log(\tfrac{1}{\epsilon})$ & $\tfrac{\abs{\cS} \log \abs{\cA}}{(1-\gamma)^4 \epsilon^2} $ &
$\tfrac{\abs{\cS} \abs{\cA} \log \abs{\cA}}{(1-\gamma)^5 \epsilon^2} \log(\tfrac{1}{\epsilon}) $   \\ 
\hline
\end{tabular}
\caption{{Comparison of sample and iteration complexities between SBPMD with uniform sampling,  and the stochastic batch PG method SPMD. Each entry value omits an $\cO(\cdot)$ notation.
The sample complexities of SPMD are improved versions of those presented in \cite{lan2021policy},  as suggested by the author.}
}
\label{tab_iter_sample_complexity_comparison}
\end{table}

Before we end this section, we briefly remark that similar to the discussions in the deterministic setting (Section \ref{sec_effect_sampling}), SBPMD with a hybrid sampling scheme enjoys potential instance-dependent acceleration, leading to even better sample complexities in terms of dependence on the state space (cf., Corollary \ref{cor_accelerate_with_power_law}). 
We omit detailed technical discussions for the sake of presentation brevity.

\section{Numerical Study}\label{sec:exp}

We conduct numerical experiments to illustrate the efficiency of the proposed BPMD methods, with the benchmark RL environment GridWorld.
 The environment specifies a square-shaped grid with a given size,  and a learning agent.
Each gridpoint is associated with a cost, and there exists four types of gridpoints: {\it goal, trap, regular, and block}, 
 taking up $f_G, f_T, f_R, f_B$-fraction of the total points, respectively.
At each timestep, the agent has four actions available (moving left/right/up/down by one gridpoint).
Upon choosing an action, the agent then moves along the chosen direction with probability $p$, or a randomly chosen direction with probability $1-p$.
If the gridpoint to be moved to is either a block or a point outside the grid, then the agent remains at the current location. 
The agent then receives a reward at updated gridpoint and the process repeats itself at the next timestep.
The costs of goal, target, and regular gridpoints are denoted by $c_G$, $c_T$, and $c_R$ respectively, with $c_G < c_R < c_T$. 
When either at a goal or a trap point, the agent transits to a pre-specified regular point. 
 Given an integer $d$, we generate a GridWorld environment of size $d$, and different types of gridpoints are randomly distributed with their ratio specified by $(f_G, f_T, f_R, f_B)$.
For all realizations of GridWorld environments, we set discount factor $\gamma = 0.9$,  $p = 0.7$,
$(f_G, f_T, f_R, f_B) = (0.05, 0.05, 0.8, 0.1)$, and $(c_G, c_R, c_T)=(-0.1, 0.0, 10)$.
The construction of cost creates a non-trivial MDP by making separations of goal and regular states difficult even in the deterministic setting.

\subsection{Deterministic Setting}
We compare 
the following methods:
(1) BPMD with uniform sampling;
(2) BPMD with $\nu^*$-sampling;
(3) BPMD with hybrid sampling. 
(4) BPMD with a randomly constructed sampling. 
(5) Policy mirror descent (PMD) \cite{lan2021policy}.
(6) Natural policy gradient (NPG) \cite{kakade2001natural}.
Note that NPG shares the same update as PMD instantiated with KL-divergence. 
(7) Policy gradient (PG) \cite{agarwal2020optimality}.
Here PMD, NPG, and PG are batch PG methods.
Key hyperparameters are set as follows. 

{\it Sampling Distributions of BPMD:}
(1) Uniform sampling refers to sampling with $\mathrm{Unif}(\cS)$.
(2) $\nu^*$-sampling uses $\nu^*$ as the sampling distribution, computed by first running policy iteration and getting the optimal policy.
(3) Hybrid sampling first uses $\nu^*$-sampling and then switching to the uniform sampling.
(4) Random sampling refers to using a $\rho \in \Delta_{\cS}$ randomly constructed  as follows. We sample $\rho(s) \sim \mathrm{Unif}([0,1])$, and then divide entries by the sum of $\rho$.
We only consider the case of exact information to better illustrate theories.

{\it Stepsizes:}
(1) For BPMD with static samplings, including uniform, $\nu^*$, and random sampling, the stepsizes are set according to \eqref{cor_non_sc_static_linear} with equality.
(2) For BPMD with hybrid sampling, stepsizes are detailed later.
(3) PMD/NPG with exponential stepsizes: $\eta_0 = 1, \eta_k = \eta_{k-1} \gamma^{-1}$, which attains linear convergence \cite{xiao2022convergence}.
(3) PMD/NPG with a constant stepsize: $\eta =1$, which attains sublinear convergence \cite{agarwal2020optimality, lan2021policy}.
(4) PG uses a constant stepsize  given in Theorem 4.1,  \cite{agarwal2020optimality}.

For hybrid sampling, across all three environments, we set $k_\tau = \ceil{\alpha /\rho_{\cH}^\dagger}$ in light of its definition in Theorem \ref{thrm_hybrid_sampling}, where $\cH$ is set to be the subset of $ \nu^*$ with the top $2\%$ values.
We set $\alpha = 5$ for $\abs{\cS} \in \cbr{100, 225}$ and $\alpha= 15$ for $\abs{\cS} = 625$.
The stepsizes are set according to \eqref{stepsize_hybrid_sampling} with equality.

\begin{figure}[htb!]
\makebox[\linewidth][c]{%
     \centering
     \begin{subfigure}[b]{0.33\textwidth}
         \centering
         \includegraphics[width=\textwidth]{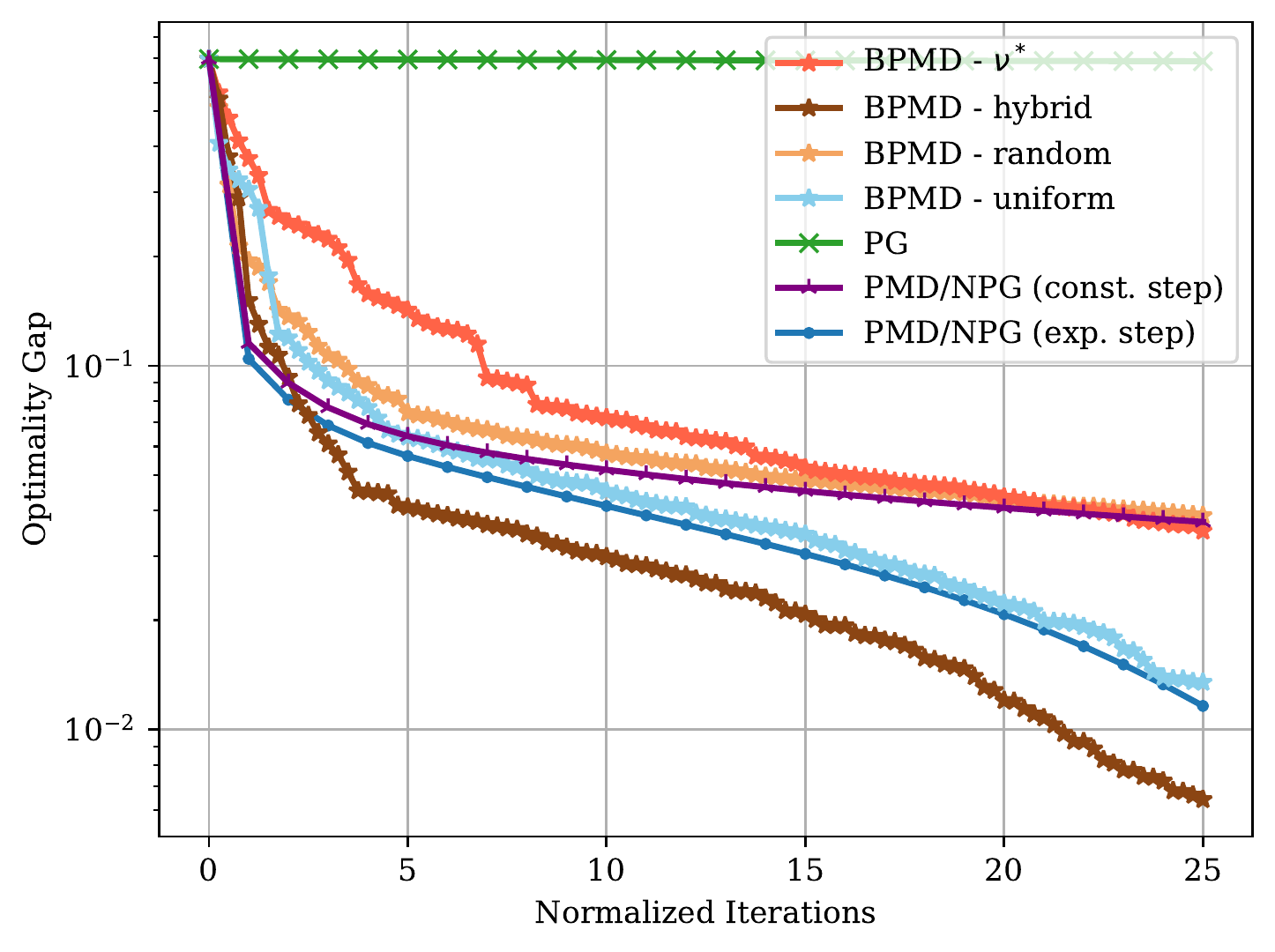}
             \vspace{-0.25in}
         \caption{$\abs{\cS} = 100$.}
     \end{subfigure}
     \begin{subfigure}[b]{0.33\textwidth}
         \centering
         \includegraphics[width=\textwidth]{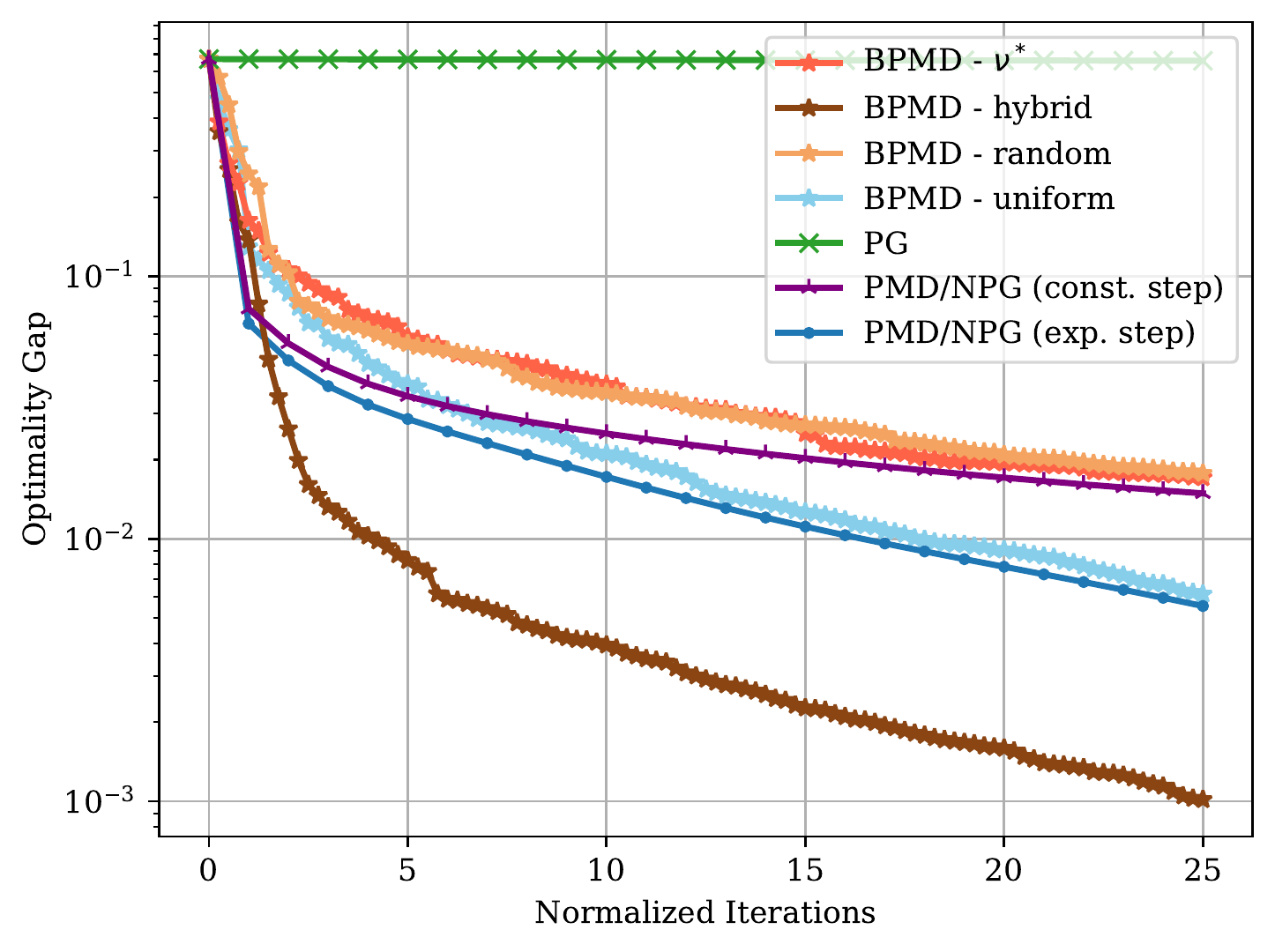}
             \vspace{-0.25in}
         \caption{$\abs{\cS} = 400$.}
     \end{subfigure}
        \begin{subfigure}[b]{0.33\textwidth}
         \centering
         \includegraphics[width=\textwidth]{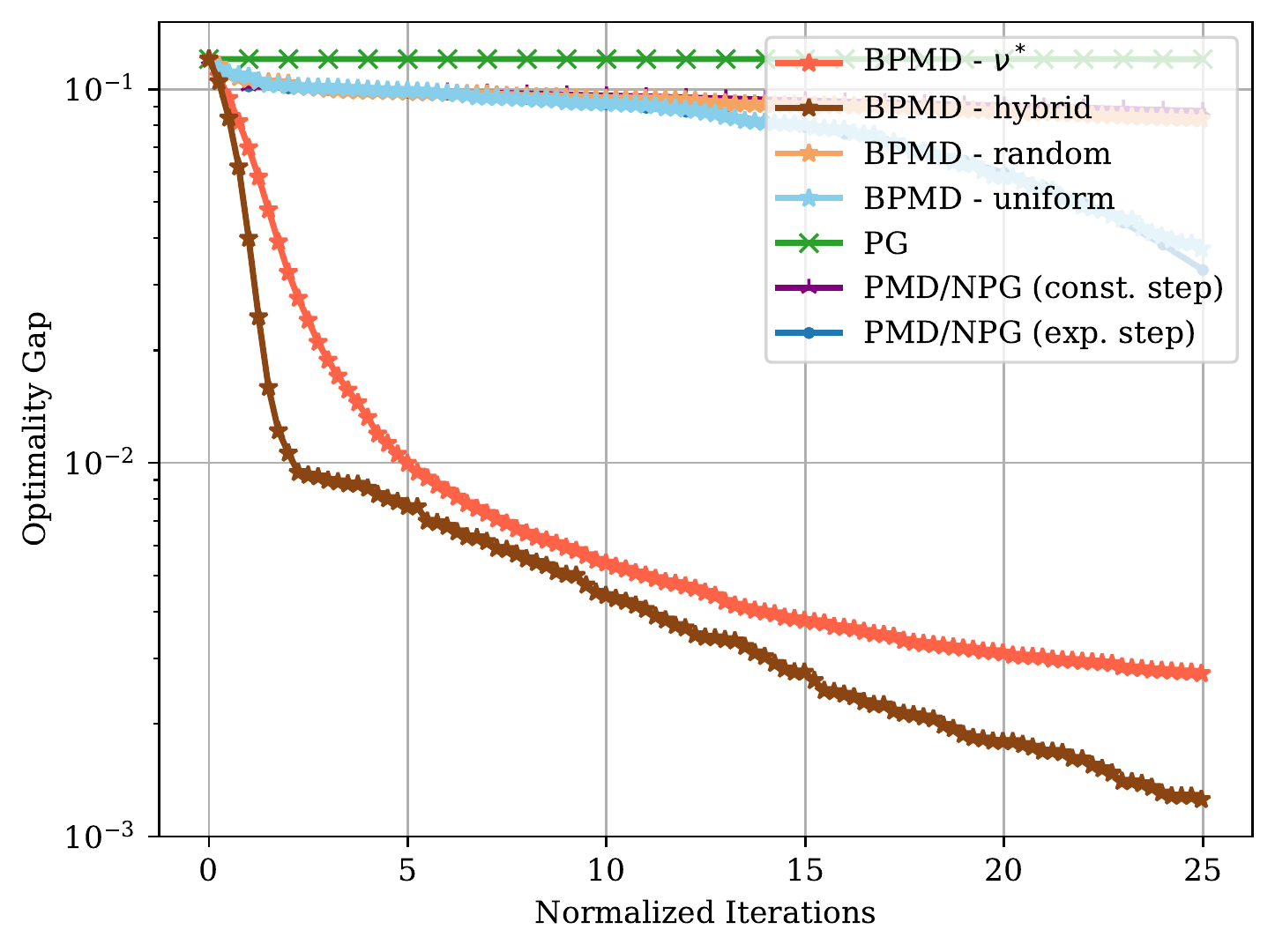}
             \vspace{-0.25in}
         \caption{$\abs{\cS} = 625$.}
     \end{subfigure}    }    
    \vspace{-0.4in}
  \caption{ 
  Comparison of BPMD variants with batch PG methods on randomly generated GridWorld environments, each with different number of states.
  }
  \label{fig:deterministic}
\end{figure}

To faciliate a fair comparison between BPMD and batch PG methods, we define the normalized iteration to be number of state-wise policy updates divided by the number of states.
Figure \ref{fig:deterministic} reports the results for 3 randomly generated GridWorld instances with different sizes.
The following observations can be made.
First, BPMD with uniform sampling is comparable to PMD/NPG with exponential stepsize. 
Second, for BPMD variants, (1) Hybrid sampling consistently outperforms uniform sampling and other BPMD variants;
(2) Random sampling underperforms uniform sampling;
(3) 
The initial accleration of using $\nu^*$-sampling gets more evident for larger state space,
and using $\nu^*$-sampling eventually plateaus despite of fast progress in the early stage. 
The benefit of adopting hybrid sampling is consistent with the discussions in Theorem \ref{thrm_hybrid_sampling}.
Moreover,  greedily exploiting $\nu^*$ eventually leads to bottlenecked convergence, consistent with Proposition \ref{prop_optimal_visitation_worse_than_uniform}.
 Finally, PG and PMD/NPG with constant stepsize clearly exhibit  sublinear convegence.

\begin{figure}[htb!]
\makebox[\linewidth][c]{%
     \begin{subfigure}[b]{0.33\textwidth}
         \centering
          \includegraphics[width=\textwidth]{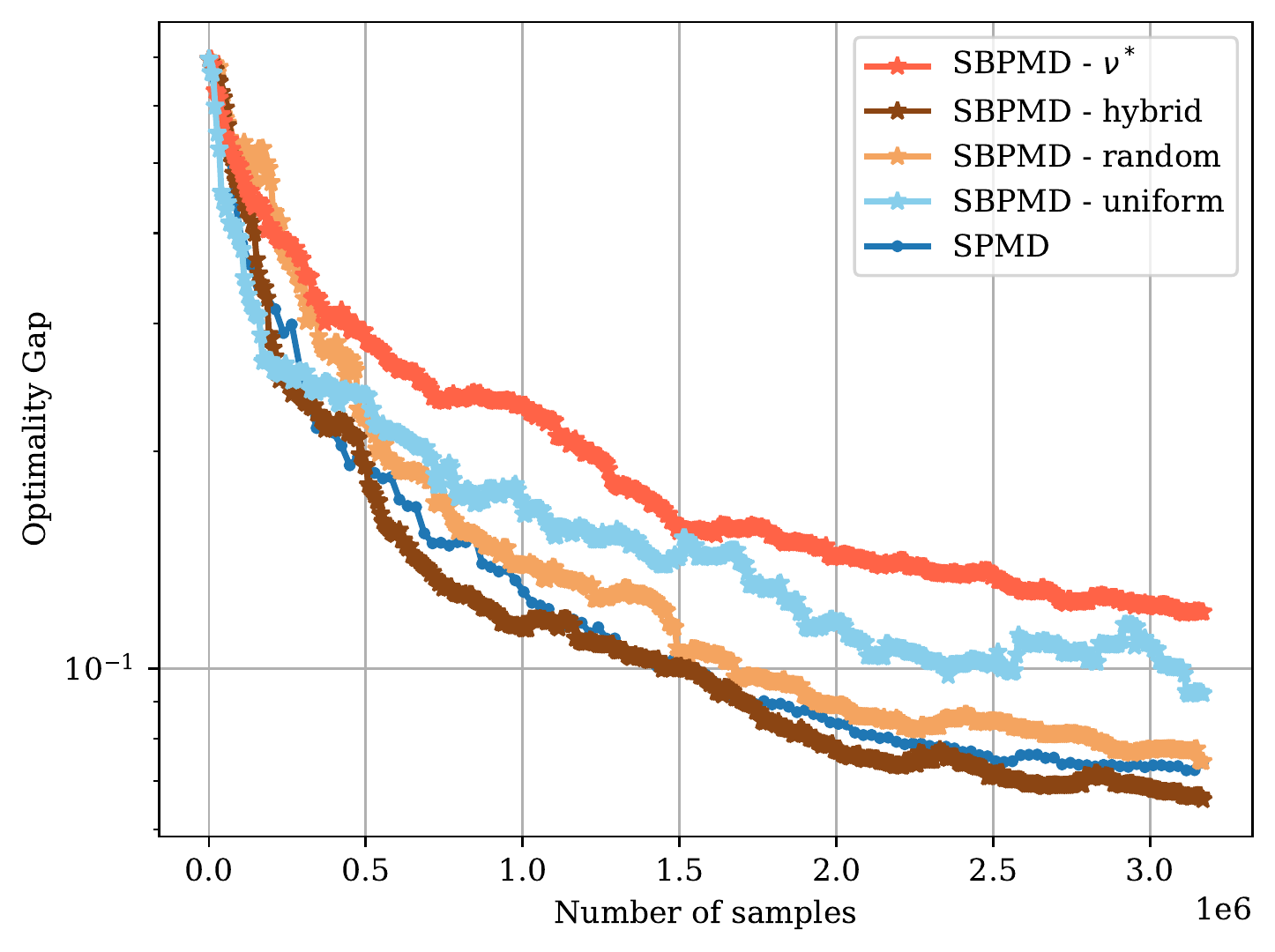}
          \vspace{-0.25in}
         \caption{$\abs{\cS} = 100$.}
     \end{subfigure}
        \begin{subfigure}[b]{0.33\textwidth}
         \centering
         \includegraphics[width=\textwidth]{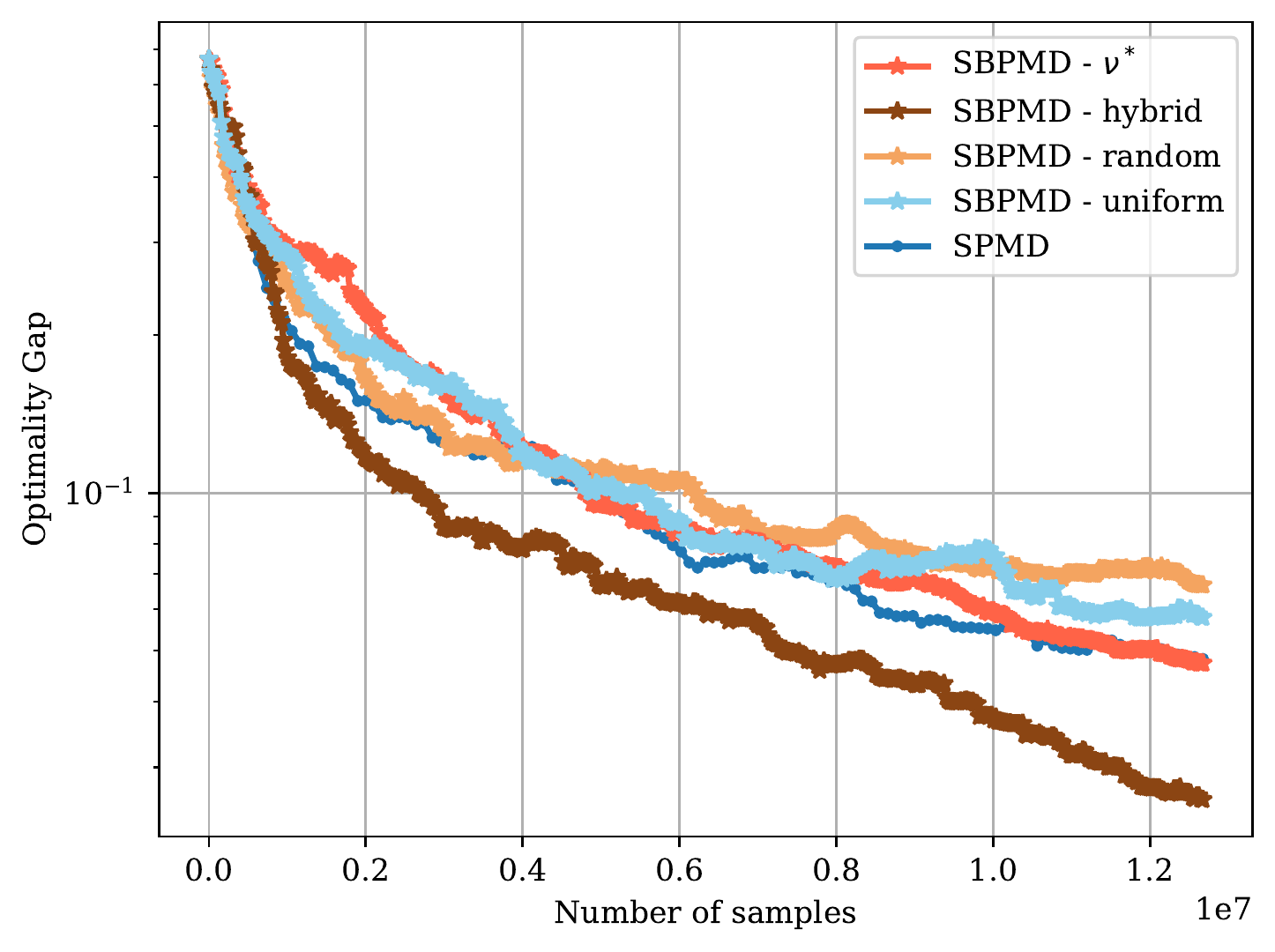}
          \vspace{-0.25in}
         \caption{$\abs{\cS} = 400$.}
     \end{subfigure}        
             \begin{subfigure}[b]{0.345\textwidth}
         \centering
         \includegraphics[width=\textwidth]{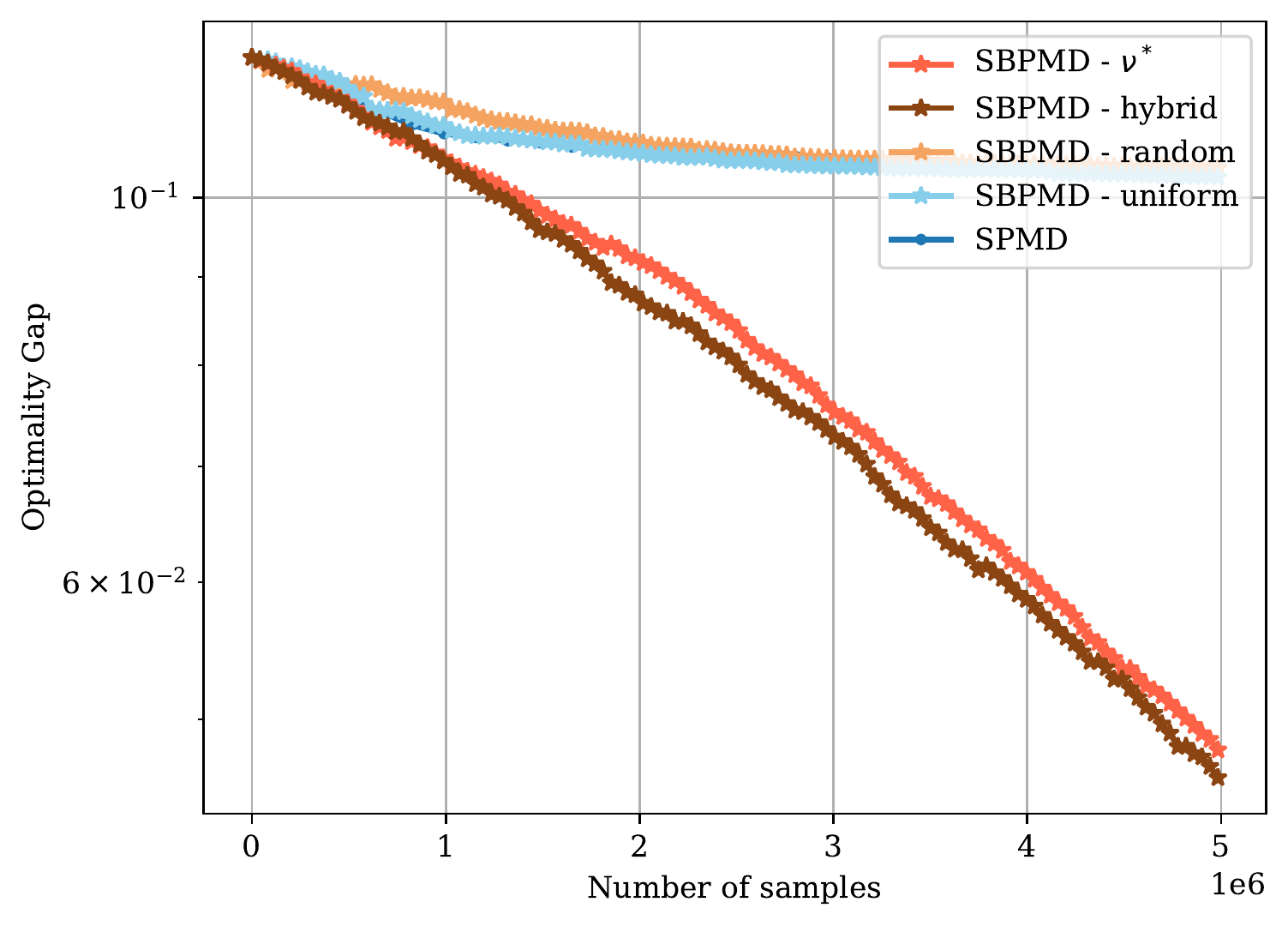}
          \vspace{-0.25in}
         \caption{$\abs{\cS} = 625$.}
     \end{subfigure}   
        }    
    \vspace{-0.4in}
  \caption{ 
 Comparison of BPMD variants with SPMD, which attains the best sample complexity among batch PG methods.
 Policy evaluation uses the method of independent trajectories. 
  }
  \label{fig:independent}
\end{figure}

\subsection{Stochastic Setting}

We compare the following methods: 
(1) SBPMD with uniform sampling; 
(2) SBPMD with $\nu^*$ sampling;
(3) SBPMD with hybrid sampling;
(4) SBPMD with a randomly constructed sampling.
(5) SPMD \cite{lan2021policy}, which attains the best sample complexity among batch stochastic PG methods.
Note that SPMD instantiated with KL-divergence shares the same update with the stochastic NPG method. 
The sampling distributions in BPMD are constructed in the  same way as the deterministic setting.

{\it Stepsizes:}
Given Corollary \ref{cor_stoch_nonsc_uniform_sampling}, one can readily check that for any target optimality gap $\epsilon > 0$, the choice of stepsize for SBPMD with an static sampling (e.g., uniform, $\nu^*$, and random sampling) is given by $\eta = \beta \epsilon (1-\gamma)^3 / \overline{c}^2$, where $\beta$ denotes an absolute constant.
We set the same stepsize for SPMD.

{\it Trajectories:}
For all variants of SBPMD and SPMD, the trajectory length is set  as in Theorem \ref{thrm_sample_sbpmd_nonsc}.

{\it Target Precision:} Noting that stepsize and trajectory require a target optimality gap $\epsilon$, we set $\epsilon = 0.05$.

It is worth pointing out that the sample complexities of SPMD presented in Table \ref{tab_iter_sample_complexity_comparison} are indeed  achieved by the aforementioned stepsize and trajectory setup, with a simplified version of Lemma \ref{lemma_stoch_generic_recursion} for SPMD.
For simplicity, we set $\beta = \overline{c}^2 / (1-\gamma)^3$ so that $\eta = \epsilon$.
The switching point $k_\tau$ for hybrid sampling takes the form of $k_\tau =  \ceil{\alpha/(\rho_{\cH}^\dagger )}$, in light of the first term in \eqref{eq_stoch_non_sc}, where $\cH$ is the same as the deterministic setting.
We set $\alpha = 20$ for $\abs{\cS} \in \cbr{100, 400}$, and $\alpha = 300$ for $\abs{\cS} = 625$.

We report the progresses of optimality gap versus the number of samples in Figure  \ref{fig:independent}.
The comparative performances of stochastic methods seem consistent with their deterministic counterparts in the low-precision region ($\sim 10^{-1}$, cf. Figure \ref{fig:deterministic}).
In particular:
(1) SBPMD with uniform sampling is comparable to SPMD. 
(2) The early acceleration of $\nu^*$-sampling becomes more evident for large state space.
(3) The benefit of using a hybrid sampling is also clear for large state space, where we can use a larger switching point.

\bibliographystyle{plain}

\vspace{-0.1in}
\bibliography{bcpmd}

\appendix

\end{document}